\documentclass[twoside]{article}

\usepackage[preprint]{aistats2027}

\usepackage[round]{natbib}

\usepackage{mathtools}

\usepackage{amsmath,amsfonts,bm}

\def\1{\bm{1}}

\DeclareMathAlphabet{\mathsfit}{\encodingdefault}{\sfdefault}{m}{sl}
\SetMathAlphabet{\mathsfit}{bold}{\encodingdefault}{\sfdefault}{bx}{n}

\def\gE{{\mathcal{E}}}

\def\gL{{\mathcal{L}}}

\newcommand{\R}{\mathbb{R}}

\DeclarePairedDelimiterX{\infdivx}[2]{(}{)}{%
  #1\;\delimsize\|\;#2%
}

\DeclarePairedDelimiterX{\inp}[2]{\langle}{\rangle}{#1, #2}

\newcounter{protocol}

\usepackage{amssymb,amsthm}
\usepackage{graphicx}
\usepackage{xcolor}
\usepackage{booktabs}
\usepackage{enumitem}
\usepackage[hidelinks]{hyperref}
\usepackage{aliascnt}
\usepackage[capitalise,nameinlink]{cleveref}

\newtheorem{theorem}{Theorem}[section]
\newaliascnt{lemma}{theorem}
\newtheorem{lemma}[lemma]{Lemma}
\aliascntresetthe{lemma}
\newaliascnt{proposition}{theorem}
\newtheorem{proposition}[proposition]{Proposition}
\aliascntresetthe{proposition}
\newaliascnt{corollary}{theorem}
\newtheorem{corollary}[corollary]{Corollary}
\aliascntresetthe{corollary}
\theoremstyle{definition}
\newaliascnt{definition}{theorem}
\newtheorem{definition}[definition]{Definition}
\aliascntresetthe{definition}
\newaliascnt{remark}{theorem}
\newtheorem{remark}[remark]{Remark}
\aliascntresetthe{remark}
\crefname{lemma}{Lemma}{Lemmas}
\crefname{proposition}{Proposition}{Propositions}
\crefname{corollary}{Corollary}{Corollaries}
\crefname{definition}{Definition}{Definitions}
\crefname{remark}{Remark}{Remarks}

\newcommand{\cL}{\mathcal{L}}
\newcommand{\Ltil}{\widetilde{\mathcal{L}}}
\newcommand{\cC}{\mathcal{C}}
\newcommand{\cD}{\mathcal{D}}
\newcommand{\cS}{\mathcal{S}}
\newcommand{\xe}{x_{\mathrm{e}}}
\newcommand{\xm}{x_{-}}
\newcommand{\xp}{x_{+}}

\newcommand{\he}{h_{\mathrm{e}}}
\newcommand{\hmid}{h_{-}}
\newcommand{\hp}{h_{+}}
\newcommand{\thh}{\hat\vartheta}
\newcommand{\thb}{\bar\vartheta}
\newcommand{\ths}{\bar\vartheta^{\star}}
\newcommand{\relu}[1]{[#1]_{+}}
\newcommand{\ip}[2]{\langle #1,#2\rangle}
\newcommand{\Gs}{G^{\star}}
\newcommand{\cCu}{\cC_a^{+}}
\newcommand{\thu}{\hat\theta_{\mathrm u}}
\newcommand{\thbu}{\bar\theta}
\DeclarePairedDelimiter{\norm}{\lVert}{\rVert}

\begin{document}

\runningtitle{A Minimal ReLU Classifier at and beyond the Edge of Stability}
\runningauthor{Quan M. Nguyen, Robin Yadav, Param Mody, Mark Schmidt}

\twocolumn[

\aistatstitle{A Minimal ReLU Classifier on Non-Linearly Separable Data: Convergence and Implicit Bias at and beyond the Edge of Stability}

\aistatsauthor{
Quan Nguyen\textsuperscript{1*$\dagger$} \\
{\normalfont\small\href{mailto:manhquan233@gmail.com}{\texttt{manhquan233@gmail.com}}} \And
Robin Yadav\textsuperscript{1*} \\
{\normalfont\small\href{mailto:yadavr@cs.ubc.ca}{\texttt{yadavr@cs.ubc.ca}}} \And
Param Mody\textsuperscript{1} \\
{\normalfont\small\href{mailto:pm13@student.ubc.ca}{\texttt{pm13@student.ubc.ca}}} \And
Mark Schmidt\textsuperscript{1,2} \\
{\normalfont\small\href{mailto:schmidtm@cs.ubc.ca}{\texttt{schmidtm@cs.ubc.ca}}}
}
\vspace{4pt}
\aistatsaddress{
\textsuperscript{1}University of British Columbia \\
\textsuperscript{2}Canada CIFAR AI Chair (Amii)
} ]

\begingroup
\renewcommand{\thefootnote}{\ensuremath{\fnsymbol{footnote}}}
\footnotetext[1]{Equal contribution.}
\footnotetext[2]{Corresponding author: Quan Nguyen.}
\endgroup

\begin{abstract}
Gradient descent (GD) with large learning rates can successfully optimize neural networks even though classical smoothness-based learning rate conditions are violated. Yet theory for this regime largely covers linear models or linearly separable data. We analyze a minimal setting outside both regimes. Specifically, a two-neuron ReLU network trained with logistic loss on 1D non-linearly separable data. The objective is nonconvex and nonsmooth, and GD exhibits edge-of-stability (EoS) behavior. 
We identify an active cone in parameter space that contains every perfect classifier, where the problem becomes piecewise logistic regression with a varying effective dataset. Within this cone, we prove a finite-time loss bound for constant learning rates and show that GD converges in direction to an explicit max-margin vector, while the orthogonal component converges to the unique minimizer of a non-strongly convex potential. We further establish catapult results where exponentially small loss is reached after a single loss spike. Furthermore, we propose an adaptive schedule that lets learning rates grow to infinity without leaving the active cone. We prove that this schedule achieves a loss rate of $O(1/(T\log{T})$, which is faster than the standard $O(1/T)$ rate. Our theoretical results extend to local convergence on more general network and are validated on MNIST dataset.
Overall, we show that this setup provides a tractable neural-network model for studying large and adaptive step sizes beyond linear predictors and separable data.
\end{abstract}

\section{INTRODUCTION}\label{sec:intro}
\looseness=-1 The learning rates $(\eta_t)_t$ of gradient descent (GD), $\theta_{t+1}=\theta_t-\eta_t\nabla\cL(\theta_t)$ for training practical deep learning models frequently violate the classical condition $\eta<2/\beta$ for a $\beta$-smooth objective.
Full-batch GD on neural networks typically operates at the \emph{edge of stability} (EoS), where the sharpness hovers around $2/\eta$ and the loss decreases non-monotonically \citep{cohen2021gradient}. Larger step sizes can trigger a loss spike followed by a fast decrease \citep{lewkowycz2020large}, known as \emph{catapult}.

\begin{figure*}[t]
\centering
\includegraphics[width=\textwidth]{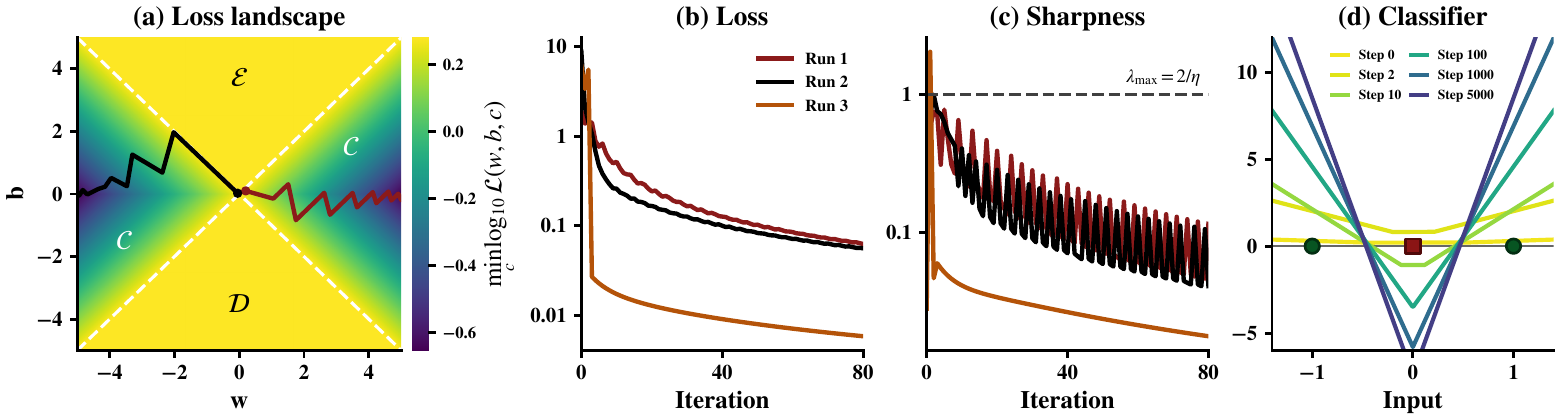}
\vspace{-6mm}
\caption{\textbf{Training a simplified ReLU network with GD on a three point non-linearly separable dataset.} \textbf{(a)} shows the loss landscape with respect to the hidden unit weight and bias $w$ and $b$. The active cone $\cC$ contains all perfect classifiers, the dead region $\cD$ has zero gradient, while the loss on $\gE$ is bounded below by a constant. Two GD trajectories with $\eta=1$ are shown. \textbf{(b)} Loss and \textbf{(c)} normalized sharpness $\eta\lambda_{\max}(\nabla^2\cL(\theta_t))/2$ (above $1$ means above the EoS threshold) for three runs; the loss is non-monotone and the sharpness exceeds $2/\eta$. \textbf{(d)} is the evolution of the learned classifier on the data points for the initialization $(0.2,0.1,0)$ with $\eta=1$, which approaches $(2|x|-1)\log(\eta t)+(1-|x|)\log3$ (\Cref{thm:implicit}). }
\label{fig:overview}
\end{figure*}

For logistic regression, large-step GD is generally well understood. On linearly separable data it converges in direction to the max-margin solution for every constant step size \citep{WuNeurIPS2023ImplicitBiasGDLogRegressionsEoS}, and large or adaptive step sizes accelerate it \citep{WuCOLT24,ZhangICML2025gradient,vaswani2025armijo}, whereas on non-separable data it may cycle \citep{meng2024gradient}.
For neural networks, implicit-bias theory concerns gradient flow or step sizes small enough for monotone descent \citep{lyu2020gradient,ji2020directional,chizat2020implicit}, and large-step analyses of two-layer networks from initialization assume linearly separable data \citep{WuCOLT24,cai2024large}.

General analyses of EoS describe GD only near a manifold of minimizers or a reference trajectory \citep{arora2022understanding,damian2023selfstabilization}.
Other works on EoS study objectives with a minimizer at finite distance, for instance functions that grow subquadratically around a local minimizer \citep{Ma2022quadraticapproximationmultiscalestructure}, or rely on conditions on all stationary points of the loss and their Hessians \citep{Ahn2022Understanding}. These conditions are not applicable to the logistic loss of a network that fits its data, which has no minimizer and tends to zero only as $\|\theta\|\to\infty$.
This leaves open the following question:

\begin{center}
\emph{Is there a neural-network classification problem, beyond linear predictors and linearly separable data, for which the entire trajectory of GD with large and adaptive step sizes can be characterized exactly?}
\end{center}

We answer this question affirmatively with a minimal example.  Three points $x\in\{-1,0,1\}$ with labels $(+1,-1,+1)$, which are not linearly separable, and a two-neuron ReLU network defined as
\begin{equation}\label{eq:model}
f_{w,b,g}(x)=\relu{wx+b}+\relu{-wx+b}+g
\end{equation}
trained with the logistic loss.
A similar model $f_{a_-,a_+,b}=a_-\sum_i \relu{-x[i]+b} + a_+\sum_i \relu{+x[i]+b}$ was studied in~\citet{ahn2023learning}, which focuses on the role of large learning rates in the \emph{initial phase} of training, where the bias $b$ decreases rapidly. 
Instead, we focus on the \emph{full trajectory} of all three parameters $w, b$ and $g$ in a simpler model.
Despite its simplicity, the problem~\eqref{eq:model} has a nonconvex, nonsmooth objective with an absorbing ``dead'' region, and GD displays EoS behavior (\Cref{fig:overview}).
Since $f_{w,b,g}$ is $1$-homogeneous in $(w,b,g)$, gradient flow on it is covered by \citet{lyu2020gradient} once all three points are classified correctly. However, GD with step sizes at or above the EoS threshold, growing step sizes, and the phase before the data are fit remain open, and they are the subject of this paper.
Our contributions are as follows.

\begin{itemize}[leftmargin=*,itemsep=1pt,topsep=2pt]
\item \textbf{Active cone.} We identify a cone $\cC=\{|w|>|b|\}$ that contains every perfect classifier and on which training is a logistic regression whose effective dataset switches with the sign of the hidden bias $b$. In the feature basis, all coordinates of the reflected iterate are non-decreasing along GD in $\cC$; consequently, GD started in $\cC$ either stays in $\cC$ forever or falls into the dead region, where the loss remains at least $\log\frac{27}{4}$ (\Cref{lem:dichotomy}).
\item \textbf{Constant step sizes and full trajectories.} We give two cone-invariance conditions that admit step sizes above the EoS threshold. For an explicit set of initializations outside $\cC$, we follow GD from the initial linear regime through finite-time entry into $\cC$ and subsequent convergence (\Cref{lem:cone-entry}). For constant step sizes, we give a finite-time loss bound and an exact description of the implicit bias: the reflected iterates $\vartheta_t$ diverge along the max-margin vector $\thh=(2,0,-1)$ of \emph{all} candidate features, and the orthogonal component converges to the unique minimizer of a convex, nonsmooth, and non-strongly convex potential.
\item \textbf{Beyond EoS.} \textit{Small losses and catapult.} For a range of initializations inside the active cone,
we prove that sufficiently large constant step sizes
cause a loss spike of order $\eta$, followed by
exponentially small loss after two to five updates.
The iterates remain in the active cone through these
updates (\Cref{thm:catapult}).
\textit{Adaptive step sizes: }We then propose the schedule $\eta_t=\kappa|w_t|$, which never leaves $\cC$ while $\eta_t\to\infty$ for $\kappa\le1/2$. For $0<\kappa\le1/4$, we prove an implicit bias result, indicating thatevery initialization in $\cC$ converges to the same max-margin direction as constant-step GD, with an additional $\log\log t$ term in the parameter growth. Consequently, this implies $\cL(\theta_t)\ = \frac{1}{\kappa t\log t}$ (\Cref{thm:adaptive}). 
\item \looseness=-1 \textbf{Breaking the symmetry.} We extend our analysis to a more realistic model in which the input weights are no longer $w$ and $-w$, but are de-coupled quantities $w_1$ and $w_2$. This unsymmetric setting is more challenging since the active cone is no longer invariant, even under gradient flow. We give an explicit region around the max-margin ray that GD with any step size $\eta\le1$ never leaves, together with a finite-time loss bound (\Cref{thm:unsymmetric,cor:unsymmetric}).
\end{itemize}

\section{Related work}

\textbf{EoS and catapults.} Prior analyses of EoS study progressive sharpening \citep{wang2022analyzing}, self-stabilization \citep{damian2023selfstabilization}, sharpness reduction \citep{arora2022understanding,kreisler2023gradient}, effective stochasticity \citep{kong2020stochasticity}, and minimal models \citep{zhu2023understanding,chen2023beyond,ahn2023learning}. 

\textbf{Logistic regression.} GD on separable data converges towards the max-margin direction \citep{soudry2018implicit,ji2019implicit}, even with arbitrary constant steps \citep{WuNeurIPS2023ImplicitBiasGDLogRegressionsEoS}. Our analysis extends the latter's orthogonal-component approach to switching features and a nonsmooth potential. Large and adaptive steps accelerate convergence \citep{WuCOLT24,ZhangICML2025gradient,crawshaw2026tight}; \citet{WuNeurIPS2025large} study regularization. For nonseparable inputs, linear models retain a bounded component \citep{ji2019implicit} or can cycle at large steps \citep{meng2024gradient}.

\textbf{Feature learning and ReLU dynamics.} On XOR data, \citet{glasgow2024sgd} proves near-optimal sample complexity for minibatch ReLU SGD, while \citet{frei2023random} show feature amplification and generalization with noisy labels and fixed output weights. \citet{taheri2025sharper} prove large-step XOR learning with quadratic activations and a linear loss. For square-loss regression, \citet{tinaz2026both} track joint training of both layers on a planted linear target, and \citet{lai2026relu} characterize near-minimum-norm implicit bias on high-dimensional random data. Our focus is on logistic loss in a two-layer network under constant and increasing steps.

\textbf{Neural implicit bias.} For homogeneous networks, \citet{lyu2020gradient} give KKT characterizations conditional on sufficiently small loss. Their ReLU results concern gradient flow, and their GD results require smoothness. \citet{ji2020directional} establish directional convergence under gradient flow. \citet{cai2024large} prove large-step convergence for smooth, nearly homogeneous predictors. Guaranteeing stable-phase entry additionally uses input linear separability and activation derivatives bounded away from zero. Our results handle ReLU activation changes and nonseparable inputs directly.

\section{PROBLEM SETUP}\label{sec:setup}

In this section, we introduce a minimal non-linearly separable neural network classification objective and characterize its loss landscape. First, we introduce our notation. 

\paragraph{Notation.}
We let $\ip{u}{v}$ and $\|u\|$ denote the Euclidean inner product and norm, and $\relu{z}=\max\{z,0\}$ denote the $\mathrm{ReLU}$ activation with standard convention $\mathrm{ReLU}'(0)=0$. Let $\ell(z)=\log(1+e^{-z})$ be the logistic loss with sigmoid $\sigma(z)=1/(1+e^{-z})$.

\paragraph{Setup.}
We consider the dataset $\cS=\{(x_i,y_i)\}_{i=1}^3=\{(-1,+1),(0,-1),(+1,+1)\}$ of three points with features $x_i \in \R$ and labels $y_i \in \{-1, 1\}$. Clearly, this dataset is not linearly separable in the input features since the two endpoints have a different label than the middle point. We train the  neural network in \eqref{eq:model} consisting of two hidden ReLU neurons with shared input weights $\pm w$ and shared bias $b$, unit output weights, and an output bias $g$. The objective is
\[
\cL(w,b,g)=\sum_{(x,y)\in\cS}\ell\big(y\,f_{w,b,g}(x)\big).
\]
We write $\theta=(w,b,g)$ for the original parameters and $\cL(\theta):=\cL(w,b,g)$. Since $f_{w,b,g}=f_{-w,b,g}$, we use the reflected vector $\vartheta=(|w|,b,g)$ for the feature analysis. We denote $(w_t, b_t, g_t)$ as the iterates produced by GD $\theta_{t} = \theta_{t-1} - \eta_{t-1}\nabla_{\theta}\cL(\theta_{t-1})$ with step sizes $(\eta_t)_t$.

We first identify a region of the loss landscape that we call the active cone. 

\begin{definition}[Active cone, dead region and linear region]\label{def:cone}
The set $\cC=\{(w,b,g):|w|>|b|\}$ is called the \emph{active cone}. 
The \emph{dead region} is $\cD=\{(w,b,g):b\le-|w|\}$.
The \emph{linear region} is $\gE = \{(w,b,g): b > |w|\}$.
\end{definition}

We show that the active cone is key to successful optimization. Outside it, at least one example has nonpositive margin, so it is incorrectly classified, and the loss is bounded below by a constant. Thus, any GD trajectory whose loss converges to zero must eventually remain in the active cone. Moreover, once a trajectory enters the dead region it cannot escape. The following lemma formalizes these observations.

\begin{lemma}[Landscape outside $\cC$]\label{lem:outside}
If $\theta\notin\cC$, then $\cL(\theta)\ge\log\frac{27}{4}$ and there exists an example with $y_i f_\theta(x_i) \leq 0$. Moreover, if $\theta\in\gE\cup\cD$, then $\nabla_w \cL = 0$; on $\cD$, also $\nabla_b\cL=0$.
\end{lemma}


Our second key observation provided in \Cref{lem:piecewise} is that, within the active cone, the loss  transforms into a logistic-regression objective with an effective feature set that depends on the sign of $b$. Each endpoint example activates exactly one hidden neuron, whereas the middle example activates both neurons when $b>0$ and neither when $b\le0$. In the reflected coordinates $\vartheta$, these activation patterns produce three signed feature vectors,
\begin{equation*}\label{eq:features}
\xe=(1,1,1),\quad \xm=(0,0,-1),\quad \xp=(0,-2,-1). 
\end{equation*}

\begin{lemma}[Piecewise logistic regression]\label{lem:piecewise}
For $\theta\in\cC$, define the \emph{endpoint margin} $u:=|w|+b+g=\ip{\vartheta}{\xe}$ and the \emph{middle margin} $v:=-2\relu{b}-g$. Equivalently,
\begin{align*}
v&=\begin{cases*}
\ip{\vartheta}{\xm}, & if $b\le0$,\\
\ip{\vartheta}{\xp}, & if $b>0$,
\end{cases*}
\end{align*}
Then $\cL(\theta)=2\ell(u)+\ell(v)$.
\end{lemma}
\looseness=-1 Therefore, within the active cone the loss is convex in the reflected coordinates, but non-smooth at $b=0$. The effective dataset is $\{\xe,\xe,\xm\}$ when $b_t\le0$ and $\{\xe,\xe,\xp\}$ when $b_t>0$.  In \Cref{lem:maxmargin}, we show that there exists a common separating max-margin vector in the feature space.

\begin{lemma}[Max-margin solution]\label{lem:maxmargin}
The problem $\min_{\vartheta\in\R^3}\frac12\|\vartheta\|^2$ s.t.\ $\ip{\vartheta}{x}\ge1$ for $x\in\{\xe,\xm,\xp\}$ has the unique solution $\thh :=(2,0,-1)=2\xe+2\xm+\xp$, and all three constraints are tight.
\end{lemma}
\begin{remark}[Degenerate active dataset]
    The vector $\thh$ in \Cref{lem:maxmargin} is the max-margin vector of \emph{all} candidate features, although only two of them are active at any time. 
    Since no subset of size two of $\{\xe, \xm, \xp\}$ can span $\R^3$, the GD updates in the reflected coordinates come from a \emph{degenerate} dataset in every round. This presents a technical challenge in our implicit bias proofs that does not exist in~\cite{WuNeurIPS2023ImplicitBiasGDLogRegressionsEoS}.
\end{remark}
Furthermore, we show in \Cref{lem:dichotomy} that a trajectory can only leave $\cC$ through the boundary $b=-|w|$ ending up in the dead region $\cD$. Once GD is in $\cC$, it either remains there forever or enters $\cD$ and becomes trapped. Hence, successful optimization needs the iterates to remain in the active cone. 
\begin{remark} The network with a fixed output layer uses nonlinear ReLU activations to map a nonseparable dataset into a higher-dimensional feature space where the points can become linearly separable. A single one-dimensional input point can have multiple high-dimensional feature representations under different activation patterns. Furthermore,  the \emph{set of transformed feature vectors need not be fixed} throughout the optimization process as the activation patterns constantly change, e.g. $b_t$ fluctuates around $0$ as shown in \Cref{fig:implicit}(a).
Handling these \emph{non-stationary feature representations} is a key component of our analysis.
\end{remark}



\section{CONSTANT STEP SIZES}\label{sec:constant}

In this section, we study the convergence, EoS behaviour and implicit bias of GD with large constant step sizes. 
\Cref{sec:convergence} proves an average-iterate upper bound $\tilde{O}(1/T)$ on $\gL$.
The implicit bias with constant $\eta_t = \eta$ is presented in~\Cref{sec:implicit}.

\subsection{Cone Invariance and Loss Convergence}\label{sec:convergence}
In this subsection, we give sufficient conditions for GD to remain in the active cone and achieve an $\mathcal{O}((\log T)^2/T)$ average-loss bound with large constant step sizes. Finally, we show that certain initialization outside the active cone can enter it in finite time. 

We begin with a lemma providing sufficient conditions to guarantee that the iterates of GD (possibly with time-varying learning rates $\eta_t$) remain in the active cone. The proof is in~\Cref{app:coneinvariance-constant-step-size}.
\begin{lemma}[Cone invariance]\label{lem:invariance}
Let $\theta_t\in\cC$ and $\eta_t>0$. Then $\theta_{t+1}\in\cC$ under either of the following conditions:
\begin{enumerate}[label=\textup{(I\arabic*)},leftmargin=*,itemsep=0pt,topsep=2pt]
\item\label{I1} $0 < \eta_t\le |w_t|/2$.
\item\label{I2} $0 < \eta_t\le1$ and $2g_t<|w_t|+b_t$.
\end{enumerate}
\end{lemma}

For a fixed learning rate and an initialization $\theta_0\in\cC$, where either $0<\eta<|w_0|/2$ or $0<\eta\le1$ and $2g_0<|w_0|+b_0$, this lemma ensures that all iterates remain in $\cC$ (\Cref{cor:fixed-cone} in \Cref{app:constant}). Neither condition depends on the Hessian nor requires the step size to remain below the EoS threshold. In fact, \Cref{rem:eos-threshold} shows that these conditions allow for step sizes above the EoS threshold even at initialization.


\Cref{thm:finite} below guarantee an $O((\log{T})^2/T)$ average-iterate loss convergence rate for constant learning rates. As stated in \Cref{rem:eos-threshold}, these learning rates can be above the EoS threshold.

\begin{theorem}[Finite-time bound]\label{thm:finite}
Let $T\ge2$, $\eta_t = \eta$ be constant, and suppose $\theta_t\in\cC$ for all $t<T$. Then
\[
\begin{aligned}
\frac1T\sum_{t<T}\cL(\theta_t)=O\!\left(\frac{(\log T+\eta+\|\theta_0\|)^2}{\eta T}\right),
\end{aligned}
\]
\end{theorem}

\looseness=-1 Furthermore, \Cref{thm:sharpening} establishes progressive sharpening followed by EoS threshold crossing, consistent with the behavior observed in Figure~\ref{fig:overview}. The result holds for suitable initializations and step sizes satisfying either \ref{I1} or \ref{I2}, which also guarantee convergence.

Next, we also analyze certain initializations and trajectories outside of the active cone. From \Cref{lem:outside}, we know that initializations in $\cD$ remain trapped with loss at least $\log\frac{27}{4}$. Thus, it remains to consider initializations in the linear region $\gE$. However, we show in \Cref{cor:cone-entry-obstruction} that certain initializations in $\gE$ are stationary as well and so GD cannot enter the active cone. The following lemma provides sufficient conditions on initializations in the linear region for GD to enter the active cone in finite time and remain there.

\begin{lemma}[Entry into the active cone]\label{lem:cone-entry}
Suppose $b_0>|w_0|>0$ and $g_0 - b_0/2 + 5|w_0|/2>\log2$. If either $0<\eta\le |w_0|/2$, or $0<\eta\le1$ and $2g_0+b_0\le3|w_0|$, then there exists a finite time $\tau$ such that the GD iterates satisfy $b_t\ge|w_t|$ for $0\le t<\tau$ and $\theta_t\in\cC$ for all $t\ge\tau$.
\end{lemma}
Upon entry, these conditions ensure that \ref{I1} or \ref{I2} hold, so all subsequent iterates remain in $\cC$. Thus, GD starts in the linear region, changes activation patterns upon entering $\cC$, and then converges with the loss bound of \Cref{thm:finite}. Therefore, \Cref{lem:cone-entry}, \Cref{lem:outside} together with \Cref{thm:finite} give a characterization of the loss landscape and trajectories of GD with large learning rates.



\subsection{Implicit Bias}\label{sec:implicit}
Now, we show that when GD remains in the active cone, the iterates converge in direction to the max-margin vector, while their orthogonal component converges to a unique limit. We also derive exact asymptotics for the parameters, loss, and learned classifier.

We adopt the same coordinate transformation in~\cite{WuNeurIPS2023ImplicitBiasGDLogRegressionsEoS} and decompose $\vartheta$ in the orthonormal basis $\{\thh/\sqrt5,\,(1,0,2)/\sqrt5,\,(0,1,0)\}$ as follows:
\[
p=\frac{2|w|-g}{\sqrt5},\qquad \thb=(q,b),\quad q=\frac{|w|+2g}{\sqrt5}.
\]
Every feature has component $\gamma=1/\sqrt5$ along $\thh/\|\thh\|$, so the margins split as $\ip{\vartheta}{x_s}=\gamma p+\ip{\thb}{h_s}$ with orthogonal parts
\[
\he=\big(\tfrac{3}{\sqrt5},1\big),\quad \hmid=\big(-\tfrac{2}{\sqrt5},0\big),\quad \hp=\big(-\tfrac{2}{\sqrt5},-2\big).
\]
Since $2\he+2\hmid+\hp=0$, no $\thb\neq0$ has positive margin on all $h_s$: for every $\thb$, $\min_s\ip{\thb}{h_s}\le-m\|\thb\|$ with $m=2/\sqrt{30}$. Crucially, the feature attaining this minimum is always \emph{active}: if $b>0$ then $\ip{\thb}{\hp}<\ip{\thb}{\hmid}$ and $\xp$ is active, and vice versa. 
When $p\to\infty$ with $\thb$ bounded, $\cL(\theta)=e^{-\gamma p}\,G(\thb)\,(1+o(1))$ where the potential function $G$ is
\begin{equation}\label{eq:G}
G(q,b)=2e^{-\frac{3q}{\sqrt5}-b}+e^{\frac{2q}{\sqrt5}+2\relu{b}}.
\end{equation}
\begin{lemma}[Unique minimizer of $G$]\label{lem:G}
$G$ is convex and coercive, nondifferentiable on $\{b=0\}$, and not strongly convex. It has the unique minimizer $\ths=(q^\star,b^\star)=(\log3/\sqrt5,0)$, with $\Gs=G(\ths)=5\cdot3^{-3/5}$.
\end{lemma}

The minimizer lies exactly on the kink of $G$. In contrast to the fixed-dataset potential of \citet{WuNeurIPS2023ImplicitBiasGDLogRegressionsEoS}, $G$ is neither smooth nor strongly convex, so convergence of $G(\thb_t)$ alone does not give convergence of $\thb_t$. The uniqueness of the minimizer does (\Cref{lem:subgrad}; see also \citealp{PlanidenAndWang2015SIAM-StrongMinimizers}).

\begin{theorem}[Implicit bias and exact asymptotics]\label{thm:implicit}
Let $\eta>0$ be constant, and suppose $\theta_t\in\cC$ for all $t$ (e.g., under \ref{I1} or \ref{I2}).
Then, as $t\to\infty$:
\begin{enumerate}[label=\textup{(\alph*)},leftmargin=*,itemsep=0pt,topsep=2pt]
\item $(|w_t|,b_t,g_t)=\thh\log t+\big(\log\tfrac{\eta^2}{3},\,0,\,\log\tfrac{3}{\eta}\big)+o(1)$;
in particular, $(|w_t|,b_t,g_t)/\|\theta_t\|\to(2,0,-1)/\sqrt5$.
\item $\cL(\theta_t)=\dfrac{5}{\eta t}\,\big(1+o(1)\big)$.
\item $f_{\theta_t}(x)=(2|x|-1)\log(\eta t)+(1-|x|)\log3+o(1)$ uniformly over $x\in[-1,1]$.
In particular, the endpoint margins are $\log(\eta t)+o(1)$ and the middle margin is $\log(\eta t/3)+o(1)$.
\end{enumerate}
\end{theorem}

Part (c) describes the learned classifier: both ReLU kinks converge to the middle point ($b_t/|w_t|\to0$), and the decision boundary converges to $|x|=\frac12$, halfway between the classes. The middle margin lags the endpoint margins by exactly $\log3$, so asymptotically the middle point carries three fifths of the loss ($\sigma(-v_t)/(2\sigma(-u_t))\to3/2$), although neither the step size nor the initialization enters this split.



\subsection{Regularization and momentum}
The active-cone analysis also extends to quadratic
regularization and heavy-ball momentum.
Write $\theta_t=(w_t,b_t,g_t)$, where
$b_t$ is the shared hidden bias and $g_t$ is the output
bias. For $\lambda\ge0$, consider
$\cL_\lambda=\cL+\lambda\|\theta\|^2/2$.
If $\theta_0\in\cC$ and $2g_0\le |w_0|+b_0$, then GD
stays in the cone for
$0<\eta_t\le1/(1+\lambda)$.
With heavy-ball momentum $\beta\in[0,1)$ and zero
initial velocity, cone preservation holds for
$0<\eta_t\le(1-\sqrt{\beta})^2/(1+\lambda)$.
The dependence on $\lambda$ accounts for the contraction
introduced by regularization: an example shows that
a step size preserving the cone in the unregularized
problem can cause an exit after regularization is added.
The statements and proofs are in
Appendix~\ref{app:regularization}.

\section{BEYOND EDGE OF STABILITY}\label{sec:beyond}

\looseness=-1
In this section, we study large learning rates beyond EoS, showing that large constant rates can cause a loss spike followed by exponentially small loss, while increasing adaptive rates achieve $\mathcal{O}(\log T / T)$ convergence with analogous implicit bias results.

\begin{figure}[t]
\centering
\includegraphics[width=0.8\columnwidth]{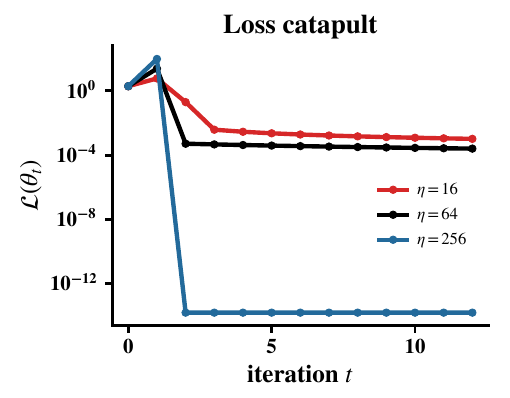}
\caption{\textbf{Two-step catapult.} GD from $\theta_0=(0.2,0.1,-0.1)$ with constant step sizes $\eta=16,64,256$.}
\label{fig:catapult}
\end{figure}

\subsection{Loss Spikes and Catapults}
\label{sec:init-phase}

The step-size conditions in \ref{I1} and \ref{I2} are
sufficient but not necessary to reach small loss.
For a range of initializations, arbitrarily large
constant step sizes cause a loss spike followed by
exponentially small loss after two updates.

\begin{theorem}[Two-step catapult]
\label{thm:catapult}
Fix $\theta_0$ independently of $\eta$, with
$0<b_0<|w_0|$. Suppose the initial margins satisfy $\sigma(-u_0)<\sigma(-v_0)<2\sigma(-u_0)$. Let $\nu=2\sigma(-u_0)-\sigma(-v_0)$ and suppose
$\nu>1/3$.
For every sufficiently large constant step size $\eta$,
the iterates $\theta_0,\theta_1,\theta_2$ remain in $\cC$.
Moreover, as $\eta\to\infty$,
\begin{align*}
\frac{\cL(\theta_1)}{\eta}\longrightarrow\nu>0, \quad 
-\frac{\log\cL(\theta_2)}{\eta}
\longrightarrow3\nu-1>0.
\end{align*}
\end{theorem}

Since the initialization is fixed, its loss does not
depend on $\eta$. Thus the first update increases
the loss to order $\eta$, while the loss after
two updates is exponentially small in $\eta$. For example, $\theta_0=(0.2,0.1,-0.1)$ satisfies the
two-step conditions; \Cref{fig:catapult} illustrates
the loss spike and recovery. 
We also obtain exponentially
small loss within five updates on an expanded initialization
set (\Cref{prop:finite-step-recovery}).


\begin{figure}[t]
\centering
\includegraphics[width=\columnwidth]{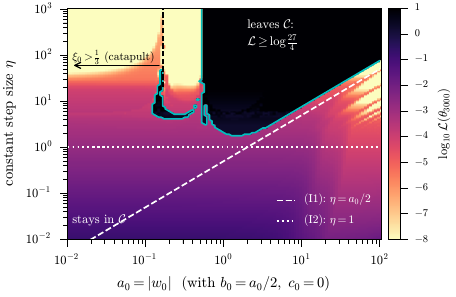}
\caption{\textbf{Phase diagram} of constant-step GD ($b_0=|w_0|/2$, $g_0=0$, $T=3000$). Color: final loss; cyan: boundary of the set of runs that ever leave $\cC$. Runs above the boundary end in the dead region with $\cL=\log\frac{27}{4}$. White lines: sufficient conditions \ref{I1} and \ref{I2}. Arrow: catapult regime of \Cref{thm:catapult}.}
\label{fig:phase}
\end{figure}

\subsection{Adaptive Learning Rate}
In \Cref{sec:convergence,sec:implicit}, we established convergence and implicit bias for constant-step GD in the active cone. One sufficient condition for cone invariance is $\eta\le |w_t|/2$ (condition \ref{I1}). By \Cref{lem:gd-update-insideCone}, $|w_t|$ is non-decreasing, and \Cref{thm:implicit} gives its exact asymptotic growth. This growing upper bound naturally suggests using the adaptive schedule $\eta_t=\kappa |w_t|$, $0<\kappa\le1/2$, which preserves the cone. \Cref{lem:adaptive-divergence} shows that $|w_t|\to\infty$ under this adaptive schedule, and hence $\eta_t\to\infty$.

For $\kappa\le1/4$, the same schedule admits a finite-time convergence bound from every initialization in the active cone.

\begin{theorem}[Convergence with increasing step sizes]\label{thm:increasing}
Let $\theta_0\in\cC$, $0<\kappa\le1/4$, and $\eta_t=\kappa|w_t|$. There is $C>0$, depending only on $\kappa$ and $\theta_0$, such that, for every integer $T\ge1$, $|w_T|\le C\log(T+1)$
\[
\frac1T\sum_{t<T}\cL(\theta_t)
\le\frac{C\log(T+1)}{T}.
\]
\end{theorem}
\begin{proof}[Sketch]
\looseness=-1 This bound improves on the fixed-step bound in \Cref{thm:finite} by a factor of $\log T$ for fixed $\eta$. The proof in \Cref{app:increasing} uses a different analysis from the shifted-comparator type argument of \citet{WuCOLT24} used to prove \Cref{thm:finite}. That argument controls cumulative loss by shifting a fixed comparator to absorb the quadratic update term and telescoping squared distances. For our growing steps, we first use exact parameter updates to bound cumulative loss by the growth of $|w_t|$. We then use a growth potential whose one-step increase is bounded by an exponentially decaying function of $|w_t|$ to obtain $|w_T|=O(\log(T+1))$ and the loss bound.
\end{proof}


\begin{theorem}[Implicit bias and exact asymptotics with increasing steps]\label{thm:adaptive}
Let $\theta_0\in\cC$, $0<\kappa\le1/4$, and $\eta_t=\kappa|w_t|$. Set $R_t=\log(2\kappa t\log t)$ for $t>1$. Then,
\begin{enumerate}[label=\textup{(\alph*)},leftmargin=*,itemsep=0pt,topsep=2pt]
\item 
$
(|w_t|,b_t,g_t)=(2R_t-\log3,\,0,\,-R_t+\log3)+o(1).
$
In particular, $(|w_t|,b_t,g_t)/\|\theta_t\|\to(2,0,-1)/\sqrt5$.
\item Loss converges as $\cL(\theta_t) = \frac{5}{2\kappa t\log t} (1+ o(1))$.
\item Uniformly over $x\in[-1,1]$,
\[
\begin{aligned}
f_{\theta_t}(x)={}&(2|x|-1)R_t
+(1-|x|)\log3+o(1).
\end{aligned}
\]
\end{enumerate}
\end{theorem}

The reflected max-margin direction and the transverse limit are the same as in \Cref{thm:implicit}; the growing step size adds a $\log\log t$ term along $\thh$. In particular, the decision boundary again converges to $|x|=1/2$, and the middle example contributes three fifths of the asymptotic loss.

The schedule is horizon-free and gains a factor $\log t$ over every constant step size (for which $t\,\cL(\theta_t)\to5/\eta$). Within the range covered by \Cref{thm:adaptive}, $\kappa=1/4$ gives the smallest asymptotic loss constant. Cone invariance holds up to $\kappa=1/2$ by \Cref{lem:invariance}; extending the asymptotic theorem to $1/4<\kappa\le1/2$ remains open. For linear logistic regression, larger gains are possible: horizon-dependent steps $\eta=\Theta(T)$ yield $\widetilde O(1/T^2)$ \citep{WuCOLT24} and loss-adaptive steps yield arbitrarily fast rates \citep{ZhangICML2025gradient}. In our model such steps are much larger than $|w_t|$ and are not covered by \Cref{lem:invariance}. Still, the loss-adaptive choice behaves remarkably well.


\begin{figure*}[!t]
\centering
\makebox[\textwidth][c]{%
\begin{minipage}[t]{0.5\textwidth}
\centering
\textbf{Trainable output layer}\\[2pt]
\includegraphics[width=0.5\linewidth]{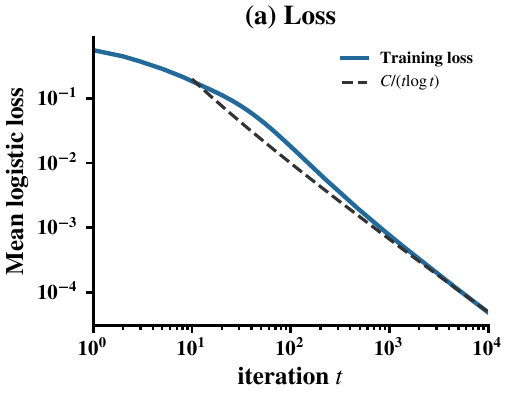}%
\includegraphics[width=0.5\linewidth]{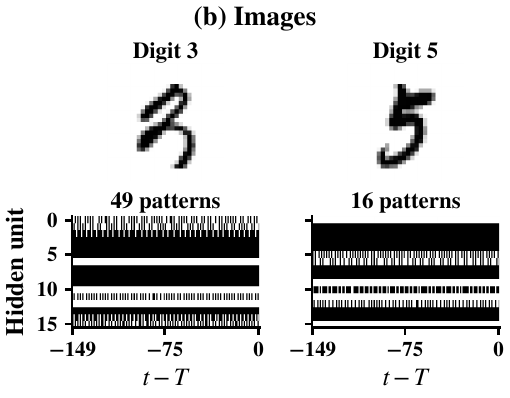}
\end{minipage}%
\begin{minipage}[t]{0.5\textwidth}
\centering
\textbf{Fixed output layer}\\[2pt]
\includegraphics[width=0.5\linewidth]{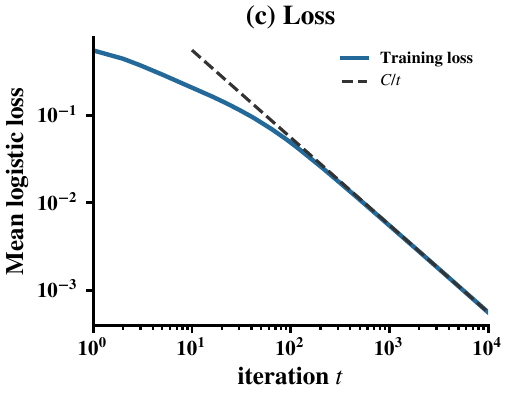}%
\includegraphics[width=0.5\linewidth]{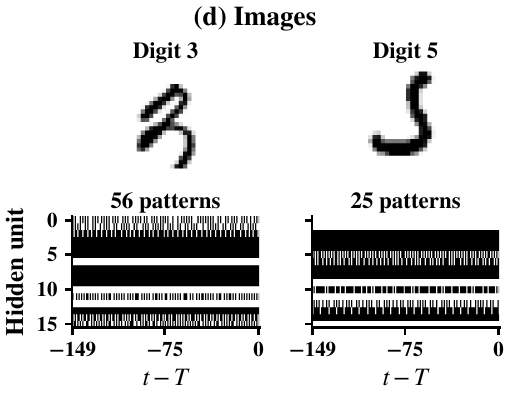}
\end{minipage}%
}
\caption{\textbf{Late-stage ReLU neural network training on binary MNIST exhibits behaviour similar to the simplified model on the three-point dataset.} We train a two-layer neural network with $16$ hidden neurons. \textbf{(a)} and \textbf{(b)} use a trainable output layer while \textbf{(c)} and \textbf{(d)} and use a fixed output layer. The loss \textbf{(a)} is asymptotically $\mathcal{O}(1/(t\ln{t}))$ while the loss in \textbf{(c)} is asymptotically $\mathcal{O}(1/t)$.
\textbf{(b)} and \textbf{(d)} show the activation patterns for certain examples towards the end of training. Black means the hidden unit is active i.e. the ReLU is not zero. Even in the late training stage, the activation patterns continue to switch which mirrors the mechanism in our simplified model. Each example moves between different feature vectors in the transformed space, giving rise to piecewise logistic regression dynamics.}
\label{fig:mnist}
\end{figure*}

\section{BREAKING THE SYMMETRY}\label{sec:unsymmetric}

So far, our analysis uses symmetric weights $\pm w$ for the hidden layer. As shown in Figure 1, an absolute value classifier separates the three points. Thus, using symmetric weights incorporates the structure of the classification task into the model. 
Towards making the learning problem more realistic, we now allow independent weights $w_1,w_2$ for the two hidden neurons. 
The model becomes,
\[
f_{w_1,w_2,b,g}^a(x)=\relu{w_1x+b}+\relu{w_2x+b}+g.
\]
The objective is $\cL^a:=\sum_{(x,y)\in\cS}\ell(y f_{w_1,w_2,b,g}^a(x))$. 
\begin{remark}
    De-coupling $w_1$ and $w_2$ is only one of many possible first steps towards making the model~\eqref{eq:model} more realistic. For instance, using smooth activation functions such as GELU leads to an optimization problem on time-varying data lying on a continuous subset of $\R^3$, which may fundamentally require new analyses.
\end{remark}
Without symmetry, successful optimization from same-sign initial weights $w_1$ and $w_2$ requires at least one weight to change sign. We describe the analogous active cones and dead region for $f_{w1,w2,b,g}^a$. 

\begin{definition}[Active cone and dead region for unsymmetric weights]\label{def:unsymmetric-cone}
For $\theta=(w_1,w_2,b,g)\in\R^4$, define the \emph{active region}
\[
\cC_a=\{\theta:w_1w_2<0,\ \min\{|w_1|,|w_2|\}>|b|\}
\]
and the \emph{dead region}
\[
\cD_a=\{\theta:b\le-\max\{|w_1|,|w_2|\}\}.
\]
The active region is the union of two cones $\cC_a^{\pm}=\{\theta:\pm w_1>|b|,\ \mp w_2>|b|\}$.
\end{definition}
Within either orientation of $\cC_a$, all signed features share a common separating direction. Moreover, every parameter vector that strictly classifies all three points must lie in $\cC_a$ (\Cref{lem:unsymmetric-ntk}).

The analysis of $f_{w1,w2,b,g}^a$ is more challenging because, unlike in the symmetric model, a step-size condition alone does not guarantee that the GD iterates remain in the active cone. Indeed, \Cref{lem:unsymmetric-gf-exit} in \Cref{app:unsymmetric-gf-exit} shows that even gradient flow can leave the active cone from certain initializations.

\begin{remark}\label{rem:unsymmetric-activation-regions}
A second difficulty is that removing the symmetry constraint introduces additional activation regions whose feature datasets in $\R^4$ need not be linearly separable. Thus, any analysis of the entire GD trajectory from general initializations must account for possible transitions into regions with nonseparable feature datasets. GD can even exit the active cone $\cC_a$ and become trapped with positive limiting loss in a region other than the dead region $\cD_a$.
\end{remark}


\paragraph{Loss convergence.} Inside $\cCu$, training is again a piecewise logistic regression in $\R^4$: in the original coordinates $\theta=(w_1,w_2,b,g)$ the features are $x_1=(1,0,1,1)$, $x_2=(0,-1,1,1)$, $\xm$ and $\xp$ (extended by a zero first coordinate), and the active middle feature again switches with the sign of $b$ (\Cref{lem:unsymmetric-gd}). The max-margin vector of the four features is $\thu=(2,-2,0,-1)$, representing the same classifier as the symmetric max-margin vector $\thh$, now with $\|\thu\|=3$.
The following theorem shows a \emph{local} convergence result for $\gL^a$ that is conditioned on $\theta_0$ being sufficiently aligned with the max-margin vector $\hat{\theta}_u$.

\begin{theorem}[Local Convergence]
There exist universal constants $M$ and $P$ such that, starting from any $\theta_0$ satisfying $\frac{\inp{\theta_0}{\hat{\theta}_u}}{\norm{\hat{\theta}_u}} \geq P$ and $\norm{\theta_0 - \frac{(\theta_0^\top\hat{\theta}_u)\hat{\theta}_u}{\norm{\hat{\theta}_u}^2}} \leq M$, for any $\eta \in (0,1]$ and $T \geq 2$,
\[
\frac1T\sum_{t<T}\cL^a(\theta_t)\le\frac{3}{T+1}+\frac{\big(\|\theta_0\|+3\,(\log(T+1)+3\eta)\big)^2}{2\eta T}.
\]
    \label{thm:local-convergence-unsymmetric}
\end{theorem}

\section{EMPIRICAL VALIDATION}\label{sec:experiments}
\paragraph{Set up.}
We perform our empirical validation with a two-layer ReLU network on a binary MNIST classification task (digits 3 and 5) using GD with a fixed learning rate. 
The results are presented in Figure~\ref{fig:mnist}.
For a given digit image, each hidden unit is active if its input to the ReLU is positive and inactive otherwise. We refer to the collection of these on/off states as the image's \emph{activation pattern}.
To make this explicit, write the two-layer ReLU network as
\[
\begin{aligned}
f(x)&=\sum_{j=1}^{m}a_j\max\{w_j^\top x+b_j,0\}\\
&=\sum_{j=1}^{m}a_j\mathbf{1}\{w_j^\top x+b_j>0\}
(w_j^\top x+b_j).
\end{aligned}
\]
\paragraph{Active Cones for MNIST.}
With fixed output weights $a_j$ and a fixed activation pattern, this expression is linear in the hidden weights $w_j$ and biases $b_j$: unit $j$ contributes coefficients $(a_jx,a_j)$ when active and zero otherwise. Concatenating these coefficients gives the image's effective feature vector. Each image has at most $2^m$ such vectors, and training selects among them by changing which units are active. Fixing the activation pattern of every training image defines a polyhedral cone in the hidden parameters on which $\cL$ is a logistic-regression objective. We call such a cone \emph{active} if it contains a perfect classifier; checking whether one exists amounts to a linear feasibility problem in these parameters.
\paragraph{Loss Convergence.} We find that the loss converges at a rate consistent with our theoretical predictions, and that these activation patterns continue to switch even in the late stages of training. This switching behavior is analogous to the mechanism we analyzed in our simplified model, where changes in hidden-unit activation patterns map the same input to different effective feature vectors. With fixed output weights, changes in hidden-unit activation patterns map the same image to different effective feature vectors corresponding to the $b_t<0$ and $b_t>0$ switching behaviour in our model. Across 30 MNIST runs, 23 fit the data and thereafter remained in active cones, while the other seven ended with all hidden units inactive, supporting the active-cone/dead-region dichotomy at the end of training despite transient visits to non-active cones. Additional experimental details are provided in \Cref{app:experiments}.

\section{CONCLUSION}\label{sec:conclusion}
We studied a simplified classification problem on non-linearly separable one-dimensional data, trained by a two-neuron ReLU network using large learning rates.
Despite its simplicity, the training dynamics exhibits the qualitative phenomena that are observed in training deep learning models: the loss is nonconvex, nonsmooth and non-monotone throughout the trajectory, while EoS and catapult are directly resulted from large learning rates.  
We characterized a particular region called active cone in the paramater space that contains all of the perfect classifiers, and proved finite-time loss convergence rates and implicit biases within this region. 
Our theoretical analyses are based on the key insight that the optimization process in this ReLU model can be modelled as a piecewise logistic regression problem on time-varying high-dimensional data, thereby leveraging and extending the techniques developed for standard logistic regression in prior works.

\paragraph{Limitations and future work.}
Our theory is limited to one-dimensional data and a two-neuron network with fixed output weights. In $d\ge2$ dimensions, hidden-weight directions can rotate during training, changing activation boundaries and requiring a geometric characterization of active cones to extend our convergence and max-margin arguments.  Training the output layer, increasing width, and adding depth introduce further interactions beyond the current analysis. Although our MNIST experiments with $16$ hidden units and a trainable output layer exhibit similar switching and loss decay, extending the theory to these settings remains open.

\subsection*{AI use statement}
In this work, we used generative AI tools (Claude, Anthropic) for drafting and editing parts of the text, for deriving and checking parts of the proofs (in particular the boundedness and trapping lemmas and the exact asymptotic constants), and for writing the code that produces the experiments and figures. We have not used generative AI tools for research ideation, identifying the simplified setup and the theoretical framework. We have reviewed all AI-assisted work, verified all proofs, and re-ran all experiments. We take responsibility for the final content of this work, including text, claims, or artifacts produced with the aid of generative AI.


\bibliography{bibliography}

@inproceedings{Wang2022TwoLayerReLUHiddenConvex,
	title={The Hidden Convex Optimization Landscape of Regularized Two-Layer {ReLU} Networks: an Exact Characterization of Optimal Solutions},
	author={Yifei Wang and Jonathan Lacotte and Mert Pilanci},
	booktitle={International Conference on Learning Representations},
	year={2022},
	url={https://openreview.net/forum?id=Z7Lk2cQEG8a}
}

@InProceedings{Mishkin2022FastOptTwoLayerReLU,
  title = 	 {Fast Convex Optimization for Two-Layer {R}e{LU} Networks: Equivalent Model Classes and Cone Decompositions},
  author =       {Mishkin, Aaron and Sahiner, Arda and Pilanci, Mert},
  booktitle = 	 {Proceedings of the 39th International Conference on Machine Learning},
  year = 	 {2022}
}

@InProceedings{WuCOLT24,
  title = 	 {Large Stepsize Gradient Descent for Logistic Loss: Non-Monotonicity of the Loss Improves Optimization Efficiency},
  author =       {Wu, Jingfeng and Bartlett, Peter L. and Telgarsky, Matus and Yu, Bin},
  booktitle = 	 {Proceedings of Thirty Seventh Conference on Learning Theory},
  pages = 	 {5019--5073},
  year = 	 {2024},
  editor = 	 {Agrawal, Shipra and Roth, Aaron},
  volume = 	 {247},
  series = 	 {Proceedings of Machine Learning Research},
  month = 	 {30 Jun--03 Jul},
  publisher =    {PMLR},
  url = 	 {https://proceedings.mlr.press/v247/wu24b.html}
}

@inproceedings{ZhangICML2025gradient,
title={Gradient Descent Converges Arbitrarily Fast for Logistic Regression via Large and Adaptive Stepsizes},
author={Ruiqi Zhang and Jingfeng Wu and Peter Bartlett},
booktitle={Forty-second International Conference on Machine Learning},
year={2025},
url={https://openreview.net/forum?id=cufJbug7du}
}

@inproceedings{WuNeurIPS2025large,
title={Large Stepsizes Accelerate Gradient Descent for Regularized Logistic Regression},
author={Jingfeng Wu and Pierre Marion and Peter Bartlett},
booktitle={The Thirty-ninth Annual Conference on Neural Information Processing Systems},
year={2025},
url={https://openreview.net/forum?id=w22e5MrS4X}
}

@inproceedings{WuNeurIPS2023ImplicitBiasGDLogRegressionsEoS,
author = {Wu, Jingfeng and Braverman, Vladimir and Lee, Jason D.},
title = {Implicit bias of gradient descent for logistic regression at the edge of stability},
year = {2023},
publisher = {Curran Associates Inc.},
address = {Red Hook, NY, USA},
booktitle = {Proceedings of the 37th International Conference on Neural Information Processing Systems},
articleno = {3247},
numpages = {28},
location = {New Orleans, LA, USA},
series = {NIPS '23}
}

@article{PlanidenAndWang2015SIAM-StrongMinimizers,
author = {Planiden, C. and Wang, X.},
title = {Strongly Convex Functions, Moreau Envelopes, and the Generic Nature of Convex Functions with Strong Minimizers},
journal = {SIAM Journal on Optimization},
volume = {26},
number = {2},
pages = {1341-1364},
year = {2016},
doi = {10.1137/15M1035550}
}

@inproceedings{cohen2021gradient,
  title     = {Gradient Descent on Neural Networks Typically Occurs at the Edge of Stability},
  author    = {Cohen, Jeremy M. and Kaur, Simran and Li, Yuanzhi and Kolter, J. Zico and Talwalkar, Ameet},
  booktitle = {International Conference on Learning Representations},
  year      = {2021}
}

@article{lewkowycz2020large,
  title   = {The Large Learning Rate Phase of Deep Learning: The Catapult Mechanism},
  author  = {Lewkowycz, Aitor and Bahri, Yasaman and Dyer, Ethan and Sohl-Dickstein, Jascha and Gur-Ari, Guy},
  journal = {arXiv preprint arXiv:2003.02218},
  year    = {2020}
}

@article{soudry2018implicit,
  title   = {The Implicit Bias of Gradient Descent on Separable Data},
  author  = {Soudry, Daniel and Hoffer, Elad and Nacson, Mor Shpigel and Gunasekar, Suriya and Srebro, Nathan},
  journal = {Journal of Machine Learning Research},
  volume  = {19},
  number  = {70},
  pages   = {1--57},
  year    = {2018}
}

@inproceedings{ji2019implicit,
  title     = {The Implicit Bias of Gradient Descent on Nonseparable Data},
  author    = {Ji, Ziwei and Telgarsky, Matus},
  booktitle = {Proceedings of the Thirty-Second Conference on Learning Theory},
  series    = {Proceedings of Machine Learning Research},
  volume    = {99},
  pages     = {1772--1798},
  year      = {2019}
}

@inproceedings{ji2020directional,
  title     = {Directional Convergence and Alignment in Deep Learning},
  author    = {Ji, Ziwei and Telgarsky, Matus},
  booktitle = {Advances in Neural Information Processing Systems},
  volume    = {33},
  year      = {2020}
}

@inproceedings{lyu2020gradient,
  title     = {Gradient Descent Maximizes the Margin of Homogeneous Neural Networks},
  author    = {Lyu, Kaifeng and Li, Jian},
  booktitle = {International Conference on Learning Representations},
  year      = {2020}
}

@inproceedings{chizat2020implicit,
  title     = {Implicit Bias of Gradient Descent for Wide Two-layer Neural Networks Trained with the Logistic Loss},
  author    = {Chizat, L{\'e}na{\"i}c and Bach, Francis},
  booktitle = {Proceedings of the Thirty-Third Conference on Learning Theory},
  series    = {Proceedings of Machine Learning Research},
  volume    = {125},
  pages     = {1305--1338},
  year      = {2020}
}

@inproceedings{crawshaw2026tight,
  title     = {Tight Bounds for Logistic Regression with Large Stepsize Gradient Descent in Low Dimension},
  author    = {Crawshaw, Michael and Liu, Mingrui},
  booktitle = {Proceedings of the Thirty-Ninth Conference on Learning Theory},
  series    = {Proceedings of Machine Learning Research},
  volume    = {336},
  pages     = {1575--1610},
  year      = {2026}
}

@article{meng2024gradient,
  title   = {Gradient Descent on Logistic Regression with Non-Separable Data and Large Step Sizes},
  author  = {Meng, Si Yi and Orvieto, Antonio and Cao, Daniel Yiming and De Sa, Christopher},
  journal = {arXiv preprint arXiv:2406.05033},
  year    = {2024}
}

@inproceedings{safran2022effective,
  title     = {On the Effective Number of Linear Regions in Shallow Univariate {ReLU} Networks: Convergence Guarantees and Implicit Bias},
  author    = {Safran, Itay and Vardi, Gal and Lee, Jason D.},
  booktitle = {Advances in Neural Information Processing Systems},
  volume    = {35},
  year      = {2022}
}

@article{lu2020dying,
  title   = {Dying {ReLU} and Initialization: Theory and Numerical Examples},
  author  = {Lu, Lu and Shin, Yeonjong and Su, Yanhui and Karniadakis, George Em},
  journal = {Communications in Computational Physics},
  volume  = {28},
  number  = {5},
  pages   = {1671--1706},
  year    = {2020}
}

@inproceedings{cai2024large,
  title     = {Large Stepsize Gradient Descent for Non-Homogeneous Two-Layer Networks: Margin Improvement and Fast Optimization},
  author    = {Cai, Yuhang and Wu, Jingfeng and Mei, Song and Lindsey, Michael and Bartlett, Peter L.},
  booktitle = {Advances in Neural Information Processing Systems},
  volume    = {37},
  year      = {2024}
}

@inproceedings{zhu2023understanding,
  title     = {Understanding Edge-of-Stability Training Dynamics with a Minimalist Example},
  author    = {Zhu, Xingyu and Wang, Zixuan and Wang, Xiang and Zhou, Mo and Ge, Rong},
  booktitle = {International Conference on Learning Representations},
  year      = {2023}
}

@inproceedings{chen2023beyond,
  title     = {Beyond the Edge of Stability via Two-step Gradient Updates},
  author    = {Chen, Lei and Bruna, Joan},
  booktitle = {Proceedings of the 40th International Conference on Machine Learning},
  series    = {Proceedings of Machine Learning Research},
  volume    = {202},
  year      = {2023}
}

@inproceedings{ahn2023learning,
  title     = {Learning Threshold Neurons via Edge of Stability},
  author    = {Ahn, Kwangjun and Bubeck, S{\'e}bastien and Chewi, Sinho and Lee, Yin Tat and Suarez, Felipe and Zhang, Yi},
  booktitle = {Advances in Neural Information Processing Systems},
  volume    = {36},
  year      = {2023}
}

@inproceedings{kreisler2023gradient,
  title     = {Gradient Descent Monotonically Decreases the Sharpness of Gradient Flow Solutions in Scalar Networks and Beyond},
  author    = {Kreisler, Itai and Nacson, Mor Shpigel and Soudry, Daniel and Carmon, Yair},
  booktitle = {Proceedings of the 40th International Conference on Machine Learning},
  series    = {Proceedings of Machine Learning Research},
  volume    = {202},
  year      = {2023}
}

@inproceedings{wang2022analyzing,
  title     = {Analyzing Sharpness along {GD} Trajectory: Progressive Sharpening and Edge of Stability},
  author    = {Wang, Zixuan and Li, Zhouzi and Li, Jian},
  booktitle = {Advances in Neural Information Processing Systems},
  volume    = {35},
  year      = {2022}
}

@inproceedings{damian2023selfstabilization,
  title     = {Self-Stabilization: The Implicit Bias of Gradient Descent at the Edge of Stability},
  author    = {Damian, Alex and Nichani, Eshaan and Lee, Jason D.},
  booktitle = {International Conference on Learning Representations},
  year      = {2023}
}

@inproceedings{arora2022understanding,
  title     = {Understanding Gradient Descent on the Edge of Stability in Deep Learning},
  author    = {Arora, Sanjeev and Li, Zhiyuan and Panigrahi, Abhishek},
  booktitle = {Proceedings of the 39th International Conference on Machine Learning},
  series    = {Proceedings of Machine Learning Research},
  volume    = {162},
  pages     = {948--1024},
  year      = {2022}
}

@inproceedings{kong2020stochasticity,
  title     = {Stochasticity of Deterministic Gradient Descent: Large Learning Rate for Multiscale Objective Function},
  author    = {Kong, Lingkai and Tao, Molei},
  booktitle = {Advances in Neural Information Processing Systems},
  volume    = {33},
  year      = {2020}
}

@inproceedings{nacson2019lexicographic,
  title     = {Lexicographic and Depth-Sensitive Margins in Homogeneous and Non-Homogeneous Deep Models},
  author    = {Nacson, Mor Shpigel and Gunasekar, Suriya and Lee, Jason D. and Srebro, Nathan and Soudry, Daniel},
  booktitle = {Proceedings of the 36th International Conference on Machine Learning},
  series    = {Proceedings of Machine Learning Research},
  volume    = {97},
  pages     = {4683--4692},
  year      = {2019}
}

@article{vardi2023implicit,
  title   = {On the Implicit Bias in Deep-Learning Algorithms},
  author  = {Vardi, Gal},
  journal = {Communications of the ACM},
  volume  = {66},
  number  = {6},
  pages   = {86--93},
  year    = {2023}
}

@inproceedings{brutzkus2018sgd,
  title     = {{SGD} Learns Over-parameterized Networks that Provably Generalize on Linearly Separable Data},
  author    = {Brutzkus, Alon and Globerson, Amir and Malach, Eran and Shalev-Shwartz, Shai},
  booktitle = {International Conference on Learning Representations},
  year      = {2018}
}

@inproceedings{pilanci2020neural,
  title     = {Neural Networks are Convex Regularizers: Exact Polynomial-time Convex Optimization Formulations for Two-layer Networks},
  author    = {Pilanci, Mert and Ergen, Tolga},
  booktitle = {Proceedings of the 37th International Conference on Machine Learning},
  series    = {Proceedings of Machine Learning Research},
  volume    = {119},
  pages     = {7695--7705},
  year      = {2020}
}

@inproceedings{glasgow2024sgd,
  title     = {{SGD} Finds Then Tunes Features in Two-Layer Neural Networks with Near-Optimal Sample Complexity: A Case Study in the {XOR} Problem},
  author    = {Glasgow, Margalit},
  booktitle = {International Conference on Learning Representations},
  year      = {2024},
  url       = {https://proceedings.iclr.cc/paper_files/paper/2024/file/e6d37cc5723e810b793c834bcb6647cf-Paper-Conference.pdf}
}

@inproceedings{taheri2025sharper,
  title     = {Sharper Guarantees for Learning Neural Network Classifiers with Gradient Methods},
  author    = {Taheri, Hossein and Thrampoulidis, Christos and Mazumdar, Arya},
  booktitle = {International Conference on Learning Representations},
  year      = {2025},
  url       = {https://openreview.net/forum?id=h7GAgbLSmC}
}

@article{frei2023random,
  title   = {Random Feature Amplification: Feature Learning and Generalization in Neural Networks},
  author  = {Frei, Spencer and Chatterji, Niladri S. and Bartlett, Peter L.},
  journal = {Journal of Machine Learning Research},
  volume  = {24},
  number  = {303},
  pages   = {1--49},
  year    = {2023},
  url     = {https://jmlr.org/papers/v24/22-1132.html}
}

@inproceedings{tinaz2026both,
  title     = {When Both Layers Learn: Training Dynamics of Representing Linear Models via {ReLU} Networks},
  author    = {Tinaz, Berk and Xie, Changzhi and Soltanolkotabi, Mahdi},
  booktitle = {Proceedings of the Thirty-Ninth Conference on Learning Theory},
  series    = {Proceedings of Machine Learning Research},
  volume    = {336},
  pages     = {6310--6371},
  year      = {2026},
  url       = {https://proceedings.mlr.press/v336/tinaz26a.html}
}

@article{lai2026relu,
  title   = {How Does the {ReLU} Activation Affect the Implicit Bias of Gradient Descent on High-dimensional Neural Network Regression?},
  author  = {Lai, Kuo-Wei and Wang, Guanghui and Tao, Molei and Muthukumar, Vidya},
  journal = {arXiv preprint arXiv:2603.04895},
  year    = {2026},
  url     = {https://arxiv.org/abs/2603.04895}
}

@inproceedings{vardi2021implicit,
  title     = {Implicit Regularization in {ReLU} Networks with the Square Loss},
  author    = {Vardi, Gal and Shamir, Ohad},
  booktitle = {Proceedings of the Thirty-Fourth Conference on Learning Theory},
  series    = {Proceedings of Machine Learning Research},
  volume    = {134},
  pages     = {4224--4258},
  year      = {2021},
  url       = {https://proceedings.mlr.press/v134/vardi21b.html}
}

@inproceedings{boursier2022gradient,
  title     = {Gradient Flow Dynamics of Shallow {ReLU} Networks for Square Loss and Orthogonal Inputs},
  author    = {Boursier, Etienne and Pillaud-Vivien, Loucas and Flammarion, Nicolas},
  booktitle = {Advances in Neural Information Processing Systems},
  volume    = {35},
  year      = {2022},
  url       = {https://papers.nips.cc/paper/2022/hash/7eeb9af3eb1f48e29c05e8dd3342b286-Abstract-Conference.html}
}

@InProceedings{Ahn2022Understanding,
  title = 	 {Understanding the unstable convergence of gradient descent},
  author =       {Ahn, Kwangjun and Zhang, Jingzhao and Sra, Suvrit},
  booktitle = 	 {Proceedings of the 39th International Conference on Machine Learning},
  pages = 	 {247--257},
  year = 	 {2022},
  editor = 	 {Chaudhuri, Kamalika and Jegelka, Stefanie and Song, Le and Szepesvari, Csaba and Niu, Gang and Sabato, Sivan},
  volume = 	 {162},
  series = 	 {Proceedings of Machine Learning Research},
  month = 	 {17--23 Jul},
  publisher =    {PMLR},
  url = 	 {https://proceedings.mlr.press/v162/ahn22a.html}
}

@misc{Ma2022quadraticapproximationmultiscalestructure,
      title={Beyond the Quadratic Approximation: the Multiscale Structure of Neural Network Loss Landscapes}, 
      author={Chao Ma and Daniel Kunin and Lei Wu and Lexing Ying},
      year={2022},
      eprint={2204.11326},
      archivePrefix={arXiv},
      primaryClass={cs.LG},
      url={https://arxiv.org/abs/2204.11326}, 
}

@inproceedings{
vaswani2025armijo,
title={Armijo Line-search Can Make (Stochastic) Gradient Descent Provably Faster},
author={Sharan Vaswani and Reza Babanezhad Harikandeh},
booktitle={Forty-second International Conference on Machine Learning},
year={2025},
url={https://openreview.net/forum?id=LKQIS65fgd}
}

\clearpage
\section*{Checklist}

\begin{enumerate}

  \item For all models and algorithms presented, check if you include:
  \begin{enumerate}
    \item A clear description of the mathematical setting, assumptions, algorithm, and/or model. [Yes] See \Cref{sec:setup}.
    \item An analysis of the properties and complexity (time, space, sample size) of any algorithm. [Yes] We analyze convergence rates of GD (\Cref{thm:finite,thm:implicit,thm:adaptive}); each iteration costs $O(1)$ operations.
    \item (Optional) Anonymized source code, with specification of all dependencies, including external libraries. [Yes] Supplementary material (Python, NumPy, SciPy, Matplotlib, mpmath).
  \end{enumerate}

  \item For any theoretical claim, check if you include:
  \begin{enumerate}
    \item Statements of the full set of assumptions of all theoretical results. [Yes]
    \item Complete proofs of all theoretical results. [Yes] See \Cref{app:setup,app:constant,app:implicit,app:adaptive,app:unsymmetric}.
    \item Clear explanations of any assumptions. [Yes] The initialization and step-size conditions, including those of \Cref{thm:adaptive}, are discussed in \Cref{sec:beyond,sec:conclusion}.
  \end{enumerate}

  \item For all figures and tables that present empirical results, check if you include:
  \begin{enumerate}
    \item The code, data, and instructions needed to reproduce the main experimental results (either in the supplemental material or as a URL). [Yes] Supplementary material.
    \item All the training details (e.g., data splits, hyperparameters, how they were chosen). [Yes] \Cref{sec:experiments} and \Cref{app:experiments}.
    \item A clear definition of the specific measure or statistics and error bars (e.g., with respect to the random seed after running experiments multiple times). [Yes] All runs are deterministic; for random initializations we report counts, medians and maxima.
    \item A description of the computing infrastructure used. (e.g., type of GPUs, internal cluster, or cloud provider). [Yes] A single laptop-class CPU (\Cref{sec:experiments}).
  \end{enumerate}

  \item If you are using existing assets (e.g., code, data, models) or curating/releasing new assets, check if you include:
  \begin{enumerate}
    \item Citations of the creator If your work uses existing assets. [Not Applicable]
    \item The license information of the assets, if applicable. [Not Applicable]
    \item New assets either in the supplemental material or as a URL, if applicable. [Not Applicable]
    \item Information about consent from data providers/curators. [Not Applicable]
    \item Discussion of sensible content if applicable, e.g., personally identifiable information or offensive content. [Not Applicable]
  \end{enumerate}

  \item If you used crowdsourcing or conducted research with human subjects, check if you include:
  \begin{enumerate}
    \item The full text of instructions given to participants and screenshots. [Not Applicable]
    \item Descriptions of potential participant risks, with links to Institutional Review Board (IRB) approvals if applicable. [Not Applicable]
    \item The estimated hourly wage paid to participants and the total amount spent on participant compensation. [Not Applicable]
  \end{enumerate}

\end{enumerate}

\clearpage
\appendix
\thispagestyle{empty}

\onecolumn
\aistatstitle{A Minimal Two-Layer ReLU Classifier on Non-Linearly\\ Separable Data: Supplementary Materials}


\providecommand{\proofstep}[1]{\par\smallskip\noindent\textit{#1}\enspace}

\paragraph{Conventions.} The original symmetric parameters are $\theta=(w,b,g)$. In the proofs we abbreviate $W:=|w|$ and work with the reflected vector $\vartheta=(W,b,g)$; $\thh$ and $\thb$ denote its max-margin direction and transverse coordinates (\Cref{app:implicit}). We use the convex extension
\[
\Ltil(W,b,g):=2\ell(W+b+g)+\log\bigl(1+e^{2\relu{b}+g}\bigr),
\qquad (W,b,g)\in\R^3,
\]
which satisfies $\|\vartheta\|=\|\theta\|$ and $\Ltil(\vartheta)=\cL(\theta)$ on $\cC$.
While a trajectory remains in $\cC$, the sign of $w_t$ is fixed (\Cref{lem:gd-update-insideCone}), so the reflection $\theta\mapsto\vartheta$ is a fixed orthogonal change of coordinates along the trajectory. Consequently, GD updates and sharpness eigenvalues can be computed in the reflected coordinates; at a ReLU kink we use the branch selected by the convention $\mathrm{ReLU}'(0)=0$. The unsymmetric section (\Cref{app:unsymmetric}) uses the original four-dimensional parameters and states its basis explicitly. \Cref{tab:notation} collects the notation of the symmetric model.

\begin{table}[h]
\centering
\caption{Notation. Here $s\in\{\mathrm e,-,+\}$ indexes the three signed features.}
\label{tab:notation}
\renewcommand{\arraystretch}{1.25}
\begin{tabular}{lll}
\toprule
Symbol & Meaning & Value \\
\midrule
$\xi_t$ & index of the active middle feature & $-$ if $b_t\le0$, $+$ if $b_t>0$ \\
$n_{\mathrm e},\,n_{\xi}$ & multiplicities of the active features & $2,\ 1$ \\
$\gamma$ & max-margin of the signed features & $1/\sqrt5$ \\
$h_s$ & component of $x_s$ orthogonal to $\thh$ &
  $h_{\mathrm e}=\big(\tfrac{3}{\sqrt5},1\big)$, $h_-=\big(-\tfrac{2}{\sqrt5},0\big)$, $h_+=\big(-\tfrac{2}{\sqrt5},-2\big)$ \\
$H$ & largest norm of an orthogonal component & $\max_s\|h_s\|=\sqrt{24/5}$ \\
$m$ & margin offset (\Cref{lem:active-nonsep}) & $2/\sqrt{30}$ \\
$u_t$ & endpoint margin & $\ip{\vartheta_t}{\xe}=W_t+b_t+g_t$ \\
$v_t$ & middle-point margin & $\ip{\vartheta_t}{x_{\xi_t}}=-2\relu{b_t}-g_t$ \\
$B_t,\,A_t$ & sigmoid weights of an endpoint and the middle point & $\sigma(-u_t),\ \sigma(-v_t)$ \\
$z_{s,t}$ & orthogonal margin of feature $s$ & $\ip{\thb_t}{h_s}$ \\
\bottomrule
\end{tabular}
\end{table}

\paragraph{Frequently used facts.}
The three signed features $\xe=(1,1,1)$, $\xm=(0,0,-1)$ and $\xp=(0,-2,-1)$ all have margin one under $\thh=(2,0,-1)$, and their Gram entries are
\begin{equation}\label{eq:feature-gram}
\ip{\thh}{\xe}=\ip{\thh}{\xm}=\ip{\thh}{\xp}=1,\qquad
\|\xe\|^2=3,\quad \|\xm\|^2=1,\quad \|\xp\|^2=5,\quad
\ip{\xe}{\xm}=-1,\quad \ip{\xe}{\xp}=-3 .
\end{equation}
For the logistic loss we use $\ell'(z)=-\sigma(-z)$, $\ell''(z)=\sigma(z)\sigma(-z)$, and the elementary bounds
\begin{equation}\label{eq:logistic-facts}
\ell(z)\le e^{-z},\qquad \sigma(-z)\le\ell(z),\qquad \sigma(-z)\le e^{-z}\qquad\text{for all }z\in\R,
\qquad\qquad
\sigma(-z)\ge\tfrac12e^{-z}\qquad\text{for }z\ge0 .
\end{equation}

\section{Detailed comparison with related work}\label{appendix:detailed-Related-Works}

\paragraph{Homogeneous-network implicit bias.}
The predictor in \eqref{eq:model} is positively $1$-homogeneous in its trained parameters: $f_{\lambda\theta}(x)=\lambda f_\theta(x)$ for $\lambda>0$, even though it includes trainable biases. Consequently, the gradient-flow results of \citet{lyu2020gradient} apply once their loss-threshold condition holds. For our summed logistic loss, $\cL(\theta)<\log2$ is a sufficient threshold, ensuring that every training margin is positive. Their result identifies limit directions, after rescaling, with KKT points of
\[
\min_\theta\frac12\|\theta\|^2
\quad\text{subject to}\quad
y_i f_\theta(x_i)\ge1\quad\text{for all }i.
\]
This is a conditional asymptotic statement: it does not establish that a trajectory reaches the required loss level. Their discrete-GD analysis additionally assumes a twice continuously differentiable predictor away from the origin and step sizes satisfying a loss-dependent upper bound, so it does not directly cover our ReLU updates.

In the coordinates $(W,b,g)$, our limit $\thh=(2,0,-1)$ satisfies the corresponding nonsmooth KKT conditions explicitly. Both endpoint margin gradients are $\xe$, and the middle margin has Clarke subdifferential $\mathrm{conv}\{\xm,\xp\}$ at $b=0$. All three example margins equal one at $\thh$, and
\[
\thh=2\xe+3\left(\tfrac23\xm+\tfrac13\xp\right).
\]
Thus each endpoint can have multiplier one and the middle example multiplier three. Our contribution goes beyond identifying this KKT point: \Cref{lem:invariance,lem:cone-entry} give conditions on initialization and step size that ensure the trajectory reaches and remains in the active cone, and \Cref{thm:implicit} proves convergence to this particular direction, the limiting orthogonal component, and the exact loss constant $t\cL(\theta_t)\to5/\eta$. These conclusions also cover admissible constant steps above the EoS threshold. We do not claim convergence from every non-dead initialization; \Cref{cor:cone-entry-obstruction} gives explicit obstructions.

\paragraph{Large-step convergence for nearly homogeneous networks.}
\citet{cai2024large} separate two issues. Their Theorem~2.2 treats smooth, Lipschitz, nearly $1$-homogeneous predictors: below a step-size-dependent loss threshold, GD has monotone loss decay, a $\Theta(1/t)$ rate, and an improving modified margin. This conditional stable-phase theorem does not require linear separability of the inputs. Their Theorems~3.2 and~4.1 guarantee finite-time transition for two-layer networks with fixed output weights by additionally assuming input linear separability and $\phi'(z)\ge\alpha>0$; the combined analysis also requires smoothness. ReLU fails these activation assumptions, and our three-point dataset fails input linear separability.

For our model, cone geometry provides a different route to convergence. On $\cC$, the updates are subgradient steps on a convex extension of the loss, with the active feature switching between $\xm$ and $\xp$ according to the sign of $b$. \Cref{thm:finite} controls the best-iterate loss without assuming initial correct classification or monotone loss decay, provided the iterates remain in $\cC$. The invariant-cone conditions make that requirement verifiable at initialization. \Cref{thm:implicit} then gives exact asymptotics for every constant-step trajectory that stays in the cone. For the increasing schedule $\eta_t=\kappa|w_t|$, \Cref{thm:adaptive} proves the same limiting direction and $t\log t\,\cL(\theta_t)\to5/(2\kappa)$ for $0<\kappa\le1/4$. These are results for a specific minimal ReLU model; the cited theorems cover broader network classes under their respective assumptions.

\paragraph{ReLU regression and further connections.}
The discussion of \citet{lai2026relu} places explicit ReLU implicit bias between worst-case obstructions and structured-data guarantees. \citet{vardi2021implicit} rule out a universal nontrivial parameter-only regularizer for square loss even with a single ReLU neuron. Under orthogonal inputs and vanishing initialization, \citet{boursier2022gradient} characterize gradient flow and its minimum-norm bias. \citet{lai2026relu} instead track discrete GD and activation stabilization on high-dimensional random data, obtaining an implicit bias close to minimum-norm interpolation. Our setting concerns logistic classification with diverging parameters and permits changing activation patterns. Other connections include margin maximization in non-homogeneous and infinitely wide networks \citep{nacson2019lexicographic,chizat2020implicit,vardi2023implicit}, shallow ReLU networks \citep{safran2022effective,brutzkus2018sgd}, dead neurons \citep{lu2020dying}, and convex reformulations over activation patterns \citep{pilanci2020neural,Mishkin2022FastOptTwoLayerReLU,Wang2022TwoLayerReLUHiddenConvex}.

\section{PROOFS FOR THE PROBLEM SETUP}\label{app:setup}

We first record how GD acts inside the active cone; most later proofs build on this lemma.

\begin{lemma}[GD in the active cone]\label{lem:gd-update-insideCone}
Let $\theta_t=(w_t,b_t,g_t)\in\cC$ and $\vartheta_t=(W_t,b_t,g_t)$, and define the residuals
\[
B_t:=\sigma\bigl(-(W_t+b_t+g_t)\bigr),\qquad A_t:=\sigma\bigl(2\relu{b_t}+g_t\bigr).
\]
Then one GD step with step size $\eta_t$ gives
\[
W_{t+1}=W_t+2\eta_t B_t,\qquad
b_{t+1}=b_t+2\eta_t\bigl(B_t-A_t\mathbf 1\{b_t>0\}\bigr),\qquad
g_{t+1}=g_t+\eta_t(2B_t-A_t),
\]
and $\operatorname{sign}(w_{t+1})=\operatorname{sign}(w_t)$.
\end{lemma}

\begin{proof}
We write the loss on $\cC$ in terms of the two margins and then differentiate.

\proofstep{Step 1 (the loss on $\cC$).}
Since $\theta_t\in\cC$, we have $W_t>|b_t|$. At an endpoint $x=\pm1$, the two preactivations $\pm w_tx+b_t$ are $W_t+b_t>0$ and $b_t-W_t<0$, so exactly one neuron is active. At the middle point $x=0$, both preactivations equal $b_t$. Hence
\[
f_{\theta_t}(\pm1)=W_t+b_t+g_t,\qquad f_{\theta_t}(0)=2\relu{b_t}+g_t .
\]
Since the labels are $(+1,-1,+1)$, the endpoint and middle margins are
\[
u_t=W_t+b_t+g_t,\qquad v_t=-2\relu{b_t}-g_t,
\qquad\text{so}\qquad
\cL(\theta_t)=2\ell(u_t)+\ell(v_t),\quad B_t=\sigma(-u_t),\quad A_t=\sigma(-v_t).
\]

\proofstep{Step 2 (partial derivatives).}
The endpoint margin $u=W+b+g$ has derivative $1$ in each of $W$, $b$, $g$. With $\mathrm{ReLU}'(0)=0$, the middle margin $v=-2\relu{b}-g$ has $\partial_bv=-2\cdot\mathbf 1\{b>0\}$ and $\partial_gv=-1$. Using $\ell'(z)=-\sigma(-z)$,
\begin{align*}
\partial_W\Ltil(\vartheta_t)&=2\ell'(u_t)=-2B_t,\\
\partial_b\cL(\theta_t)&=2\ell'(u_t)-2\ell'(v_t)\,\mathbf 1\{b_t>0\}=-2B_t+2A_t\,\mathbf 1\{b_t>0\},\\
\partial_g\cL(\theta_t)&=2\ell'(u_t)-\ell'(v_t)=-2B_t+A_t.
\end{align*}
The updates of $b$ and $g$ follow from $\theta_{t+1}=\theta_t-\eta_t\nabla\cL(\theta_t)$.

\proofstep{Step 3 (the weight $w$).}
Since $W=|w|>0$ on $\cC$, the chain rule gives $\partial_w\cL=\operatorname{sign}(w)\,\partial_W\Ltil=-2B_t\operatorname{sign}(w_t)$. Thus
\[
w_{t+1}=w_t+2\eta_tB_t\operatorname{sign}(w_t),
\]
so $w$ moves away from $0$: its sign is unchanged and $W_{t+1}=W_t+2\eta_tB_t$.
\end{proof}

In vector form, with $\xi_t\in\{-,+\}$ the index of the active middle feature (\Cref{tab:notation}), \Cref{lem:gd-update-insideCone} reads
\begin{equation}\label{eq:gd-vector}
\vartheta_{t+1}=\vartheta_t+\eta_t\bigl(2B_t\,\xe+A_t\,x_{\xi_t}\bigr),
\end{equation}
as one checks coordinatewise: $2B_t\xe+A_t\xm=(2B_t,\,2B_t,\,2B_t-A_t)$ and $2B_t\xe+A_t\xp=(2B_t,\,2B_t-2A_t,\,2B_t-A_t)$. In words, a GD step in $\cC$ is a logistic-regression step on the effective dataset $\{\xe,\xe,x_{\xi_t}\}$.

\begin{proof}[Proof of \Cref{lem:outside}]
The idea is that outside $\cC$ both neurons have the same activation on all three inputs, so the network outputs the same value at every training point.

\proofstep{Step 1 (constant output on the data).}
If $\theta\notin\cC$, then $|b|\ge W$, that is, $b\ge W$ or $b\le-W$. For every $x\in\{-1,0,1\}$ we have $|wx|\le W$, so both preactivations $\pm wx+b$ are nonnegative if $b\ge W$ and nonpositive if $b\le-W$. Hence, for all three inputs,
\[
\relu{wx+b}+\relu{-wx+b}=
\begin{cases}
(wx+b)+(-wx+b)=2b, & \text{if } b\ge W,\\
0, & \text{if } b\le -W,
\end{cases}
\]
so $f_\theta(x)=z$ on the data, with $z=2b+g$ in the first case and $z=g$ in the second.

\proofstep{Step 2 (loss lower bound and a misclassified point).}
The labels are $(+1,-1,+1)$, so the three margins are $z,-z,z$, and at least one of them is nonpositive. The loss is
\[
\cL(\theta)=\varphi(z):=2\ell(z)+\ell(-z),\qquad \varphi'(z)=-2\sigma(-z)+\sigma(z)=3\sigma(z)-2 .
\]
Since $\varphi'$ is increasing and vanishes where $\sigma(z)=\frac23$, i.e., at $z=\log2$, the convex function $\varphi$ is minimized there, and
\[
\cL(\theta)\ \ge\ \varphi(\log2)=2\log\tfrac32+\log3=\log\tfrac{27}{4}.
\]

\proofstep{Step 3 (vanishing gradients).}
Now let $\theta\in\gE\cup\cD$, i.e., $b>W$ or $b\le-W$. If $b>W$, all preactivations on the data are strictly positive; if $b\le-W$, they are all nonpositive. In both cases, with $\mathrm{ReLU}'(0)=0$, the $w$-derivative of the hidden layer at every input is
\[
x\,\mathbf 1\{wx+b>0\}-x\,\mathbf 1\{-wx+b>0\}=0,
\]
so $\nabla_w\cL=0$. If moreover $\theta\in\cD$ ($b\le-W$), all indicators vanish, so also $\nabla_b\cL=0$. GD then changes only $g$, which keeps $\theta$ in $\cD$; thus $\cD$ is invariant.

\proofstep{The face $b=|w|>0$.} The complement of $\cC$ is $\gE\cup\cD\cup\{b=|w|>0\}$, and on the remaining face the gradient claim fails. There, at each endpoint one preactivation equals $2W>0$ and the other is exactly $0$, so the convention $\mathrm{ReLU}'(0)=0$ switches off exactly one neuron, as in $\cC$. Consequently $\partial_w\cL=-2\sigma(-2W-g)\operatorname{sign}(w)\neq0$, and GD takes the active-cone step of \Cref{lem:gd-update-insideCone}; this is the step \eqref{eq:cone-entry-boundary} by which GD can enter $\cC$ from $\gE$ in \Cref{lem:cone-entry}. The loss bound and the misclassification claim still hold on this face, by Steps~1 and~2.

\end{proof}

\begin{proof}[Proof of \Cref{lem:piecewise}]
The formula for the loss was derived in Step~1 of the proof of \Cref{lem:gd-update-insideCone}: for $\theta\in\cC$,
\begin{equation}\label{eq:Lcone}
\cL(\theta)=2\ell(W+b+g)+\log\bigl(1+e^{2\relu{b}+g}\bigr)=2\ell(u)+\ell(v),
\end{equation}
with $u=\ip{\vartheta}{\xe}$ and $v=-2\relu b-g$. If $b\le0$, then $v=-g=\ip{\vartheta}{\xm}$; if $b>0$, then $v=-2b-g=\ip{\vartheta}{\xp}$. This proves the lemma.

We also record two properties of the extension $\Ltil$, i.e., the right-hand side of \eqref{eq:Lcone} viewed as a function of $\vartheta=(W,b,g)\in\R^3$; both are used repeatedly below.

\proofstep{(a) $\Ltil$ is convex.} The first term is the convex function $\ell$ composed with a linear function. The second is the convex nondecreasing function $s\mapsto\log(1+e^s)$ composed with the convex function $2\relu{b}+g$.

\proofstep{(b) The GD direction is a subgradient of $\Ltil$.} Fix $\vartheta=(W,b,g)$, let $\tau:=\mathbf 1\{b>0\}$, and consider the smooth convex function
\[
h(\vartheta'):=2\ell(W'+b'+g')+\log\bigl(1+e^{2\tau b'+g'}\bigr),\qquad \vartheta'=(W',b',g').
\]
Since $\relu{b'}\ge\tau b'$ for all $b'$, with equality at $b'=b$, we have $h\le\Ltil$ everywhere and $h(\vartheta)=\Ltil(\vartheta)$. Therefore, for every $\vartheta'$,
\[
\Ltil(\vartheta')\ \ge\ h(\vartheta')\ \ge\ h(\vartheta)+\ip{\nabla h(\vartheta)}{\vartheta'-\vartheta}\ =\ \Ltil(\vartheta)+\ip{\nabla h(\vartheta)}{\vartheta'-\vartheta},
\]
that is, $\nabla h(\vartheta)\in\partial\Ltil(\vartheta)$. Replacing $\relu{b}$ by $\tau b$ is exactly the convention $\mathrm{ReLU}'(0)=0$, so $\nabla h(\vartheta)=-2B\xe-Ax_{\xi}$ is the direction used by GD in \eqref{eq:gd-vector}.
\end{proof}

By \Cref{lem:gd-update-insideCone}, $W_t-b_t$ is non-decreasing while $\theta_t\in\cC$. Hence a trajectory in $\cC$ can leave $\cC$ only through the lower face $b=-|w|$, which borders the absorbing dead region $\cD$. \Cref{lem:dichotomy} gives the exact exit condition: GD either remains in $\cC$ forever, or enters $\cD$, becomes trapped, and has loss bounded away from zero. Hence, successful optimization reduces to showing that GD remains in the active cone.

\begin{lemma}[Dichotomy]\label{lem:dichotomy}
Let $\theta_t\in\cC$ and $\eta_t>0$. Then $\theta_{t+1}\notin\cC$ if and only if
\begin{equation}\label{eq:exit}
b_t>0\quad\text{and}\quad 2\eta_t\bigl(\sigma(-v_t)-2\sigma(-u_t)\bigr)\ge W_t+b_t,
\end{equation}
and in that case $\theta_{t+1}\in\cD$. Consequently, for any step sizes and any $\theta_0\in\cC$, either $\theta_t\in\cC$ for all $t$, or there is $t^\star$ with $\theta_t\in\cD$ and $\cL(\theta_t)\ge\log\frac{27}{4}$ for all $t\ge t^\star$.
\end{lemma}

\begin{proof}[Proof of \Cref{lem:dichotomy}]
Recall that $\theta\in\cC$ if and only if $W-b>0$ and $W+b>0$. We track these two quantities over one step. By \Cref{lem:gd-update-insideCone}, $W_{t+1}=|w_{t+1}|$ and
\begin{align*}
W_{t+1}-b_{t+1}&=W_t-b_t+2\eta_tA_t\,\mathbf 1\{b_t>0\}\ >\ 0,\\
W_{t+1}+b_{t+1}&=W_t+b_t+4\eta_tB_t-2\eta_tA_t\,\mathbf 1\{b_t>0\}.
\end{align*}
The first quantity stays positive, so $\theta_{t+1}\notin\cC$ if and only if $W_{t+1}+b_{t+1}\le0$.
\begin{itemize}
\item If $b_t\le0$, then $W_{t+1}+b_{t+1}=W_t+b_t+4\eta_tB_t>0$, so $\theta_{t+1}\in\cC$.
\item If $b_t>0$, then $W_{t+1}+b_{t+1}\le0$ if and only if $2\eta_t(A_t-2B_t)\ge W_t+b_t$. This is \eqref{eq:exit}, since $A_t=\sigma(-v_t)$ and $B_t=\sigma(-u_t)$.
\end{itemize}
In the exit case, $b_{t+1}\le-W_{t+1}=-|w_{t+1}|$, i.e., $\theta_{t+1}\in\cD$. By \Cref{lem:outside}, $\cD$ is invariant under GD and the loss on $\cD$ is at least $\log\frac{27}{4}$, which proves the dichotomy.
\end{proof}

\section{PROOFS FOR CONSTANT STEP SIZES}\label{app:constant}

\subsection{Cone Invariance and the Finite-Time Bound}
\label{app:coneinvariance-constant-step-size}
\begin{proof}[Proof of \Cref{lem:invariance}]
By \Cref{lem:dichotomy}, $\theta_{t+1}\in\cC$ automatically when $b_t\le0$. So assume $b_t>0$; by the same lemma, it suffices to show
\begin{equation}\label{eq:invariance-goal}
2\eta_t(A_t-2B_t)<W_t+b_t .
\end{equation}

\proofstep{Condition \ref{I1}.} Since $A_t<1$, $B_t>0$ and $b_t>0$,
\[
2\eta_t(A_t-2B_t)<2\eta_t\le W_t<W_t+b_t .
\]

\proofstep{Condition \ref{I2}.} Set $s:=\frac{W_t+b_t}{2}>0$, so that the assumption reads $g_t<s$. We bound both margins in terms of $s$:
\[
u_t=W_t+b_t+g_t=2s+g_t<3s,\qquad
v_t=-2b_t-g_t>-(W_t+b_t)-s=-3s,
\]
where the bound on $v_t$ uses $2b_t<W_t+b_t$ (as $b_t<W_t$). Since $\sigma$ is increasing, $A_t=\sigma(-v_t)<\sigma(3s)$ and $B_t=\sigma(-u_t)>\sigma(-3s)=1-\sigma(3s)$. Therefore
\[
A_t-2B_t\ <\ 3\sigma(3s)-2\ \le\ s,
\]
where the last step is the inequality
\begin{equation}\label{eq:sigma-linear}
3\sigma(y)-2\le \frac{y}{3}\qquad\text{for all }y\ge0
\end{equation}
(proved below) applied with $y=3s$. Since $0<\eta_t\le1$, we conclude that $2\eta_t(A_t-2B_t)<2\eta_ts\le2s=W_t+b_t$, which is \eqref{eq:invariance-goal}.

\proofstep{Proof of \eqref{eq:sigma-linear}.} Let $\phi(y):=\frac y3-3\sigma(y)+2$. Then $\phi(0)=\frac12$ and
\[
\phi'(y)=\frac13-3\sigma(y)\sigma(-y).
\]
On $y\ge0$, the product $\sigma(y)\sigma(-y)$ decreases from $\frac14$ to $0$, so $\phi'$ increases from $-\frac5{12}$ to $\frac13$ and has a single zero $y^\star$. It is determined by $\sigma(y^\star)\sigma(-y^\star)=\frac19$, i.e.,
\[
\sigma(y^\star)=\frac12+\frac{\sqrt5}{6},\qquad y^\star=4\log\frac{1+\sqrt5}{2}.
\]
Thus $\phi$ decreases on $[0,y^\star]$ and increases afterwards, so
\[
\min_{y\ge0}\phi(y)=\phi(y^\star)=\frac43\log\frac{1+\sqrt5}{2}+\frac12-\frac{\sqrt5}{2}\approx0.0236>0 .
\]
In particular, \eqref{eq:sigma-linear} holds with strict inequality.
\end{proof}

\begin{corollary}[Cone invariance with a fixed learning rate]\label{cor:fixed-cone}
Let $\theta_0\in\cC$ and let GD use a fixed learning rate $\eta>0$. If either $\eta\le W_0/2=|w_0|/2$, or $\eta\le1$ and $2g_0<W_0+b_0$, then $\theta_t\in\cC$ for every $t\ge0$.
\end{corollary}

\begin{proof}
By \Cref{lem:gd-update-insideCone}, while the iterates lie in $\cC$,
\[
W_{t+1}-W_t=2\eta B_t\ge0,\qquad
\Bigl(\frac{W_{t+1}+b_{t+1}}{2}-g_{t+1}\Bigr)-\Bigl(\frac{W_t+b_t}{2}-g_t\Bigr)=\eta A_t\,\mathbf 1\{b_t\le0\}\ge0 .
\]
Suppose $\theta_0,\dots,\theta_t\in\cC$. Under the first condition, monotonicity of $W$ gives $\eta\le W_0/2\le W_t/2$, which is \ref{I1} at time $t$. Under the second condition, monotonicity of $\frac{W+b}{2}-g$ gives $2g_t<W_t+b_t$, which together with $\eta\le1$ is \ref{I2} at time $t$. In either case \Cref{lem:invariance} gives $\theta_{t+1}\in\cC$, and induction on $t$ proves the claim.
\end{proof}

\begin{remark}[EoS threshold]\label{rem:eos-threshold}
Let $\eta_t = \eta$.
If $b_0=g_0=0$, $W_0>16$, and $\eta\in[8,W_0/2]$, then \ref{I1} holds and $\lambda_{\max}(\nabla^2\cL(\theta_0))>2/\eta$. If $g_0=0, 0<b_0<W_0\le\tfrac14, \eta\in[\tfrac9{10},1]$, then \ref{I2} holds and $\lambda_{\max}(\nabla^2\cL(\theta_0))>2/\eta$. In both cases, $\theta_t\in\cC$ for every $t\ge0$.
\end{remark}

\begin{proof}[Proof of \Cref{rem:eos-threshold}]
In both cases we exhibit a unit test vector whose Rayleigh quotient exceeds $2/\eta$, which lower-bounds the sharpness. Since \ref{I1} or \ref{I2} holds at time $0$, \Cref{cor:fixed-cone} gives $\theta_t\in\cC$ for every $t\ge0$.

\proofstep{Case \ref{I1}.} Since $\eta\le W_0/2$, condition \ref{I1} holds; the interval $[8,W_0/2]$ is nonempty because $W_0>16$. At $b_0=0$ the classical Hessian does not exist, but the convention $\mathrm{ReLU}'(0)=0$ selects the branch $b\le0$, on which $\cL=2\ell(\ip{\vartheta}{\xe})+\ell(\ip{\vartheta}{\xm})$. At $\vartheta_0=(W_0,0,0)$ the margins are $u_0=W_0$ and $v_0=0$, so the Hessian of this branch in the reflected coordinates is
\[
H_0=2\ell''(W_0)\,\xe\xe^\top+\ell''(0)\,\xm\xm^\top=2\ell''(W_0)\,\xe\xe^\top+\tfrac14\,\xm\xm^\top .
\]
Taking the unit vector $\xm$ and using $\ip{\xe}{\xm}=-1$,
\[
\lambda_{\max}(H_0)\ \ge\ \xm^\top H_0\,\xm=2\ell''(W_0)+\tfrac14\ >\ \tfrac14 .
\]
Hence $\eta\lambda_{\max}(H_0)>2$ for every $\eta\ge8$.

\proofstep{Case \ref{I2}.} Here $2g_0=0<W_0+b_0$ and $\eta\le1$, so \ref{I2} holds. Since $W_0>b_0>0$, the loss is smooth near $\theta_0$ (branch $b>0$), and the margins are
\[
u_0=W_0+b_0\in\bigl(0,\tfrac12\bigr),\qquad v_0=-2b_0\in\bigl(-\tfrac12,0\bigr).
\]
Write $\ell''(z)=\frac14\bigl(1-\tanh^2(z/2)\bigr)$. For $|z|<\frac12$, the bound $|\tanh(z/2)|\le|z|/2<\frac14$ gives
\begin{equation}\label{eq:ellpp-lower}
\ell''(z)>\frac14\Bigl(1-\frac1{16}\Bigr)=\frac{15}{64}\qquad\text{for }|z|<\frac12 .
\end{equation}
The Hessian in the reflected coordinates is $H_0=2\ell''(u_0)\,\xe\xe^\top+\ell''(v_0)\,\xp\xp^\top$. For the unit vector $r=(1,2,1)/\sqrt6$ we have $\ip{\xe}{r}=4/\sqrt6$ and $\ip{\xp}{r}=-5/\sqrt6$, so
\[
\lambda_{\max}(H_0)\ \ge\ r^\top H_0\,r=\frac{32\,\ell''(u_0)+25\,\ell''(v_0)}{6}\ >\ \frac{57}{6}\cdot\frac{15}{64}=\frac{285}{128}.
\]
Since $\eta\ge\frac9{10}$, we obtain $\eta\lambda_{\max}(H_0)>\frac{9}{10}\cdot\frac{285}{128}=\frac{513}{256}>2$.
\end{proof}

\begin{proof}[Proof of \Cref{lem:cone-entry}]
Inside the linear region $\gE$, $W_t=W_0$ and $g_t-b_t/2 = g_0 - \frac{b_0}{2}$ is conserved. 
The initialization condition keeps the common output strictly above $\log2$, forcing $b_t$ to decrease until it reaches the cone boundary. Each step-size condition prevents overshooting into $\cD$ and ensures invariance after entry. For example, $(W_0,b_0,g_0)=(1,2,0)$ satisfies the hypotheses for either condition.

We first write GD in the linear region, then show that $b_t$ reaches the cone boundary in finite time, and finally show, separately for each step-size condition, that the iterate enters $\cC$ without overshooting into $\cD$ and satisfies an invariance condition there.

\proofstep{Step 1 (GD in the linear region).}
If $b>W=|w|$, both neurons are active on all three inputs, so $f_\theta(x)=2b+g=:z$ on the data and $\cL=2\ell(z)+\ell(-z)$. The two neurons' contributions to $\partial_w\cL$ cancel, while $\partial_b\cL=2h(z)$ and $\partial_g\cL=h(z)$, where
\[
h(z):=3\sigma(z)-2 .
\]
Hence, as long as $\theta_t\in\gE$, GD reads
\begin{equation}\label{eq:outside-gd}
W_{t+1}=W_t,\qquad b_{t+1}=b_t-2\eta h(z_t),\qquad g_{t+1}=g_t-\eta h(z_t).
\end{equation}
Two consequences: $g_t-\frac{b_t}{2}$ is conserved, and the common margin evolves as $z_{t+1}=z_t-5\eta h(z_t)$. Note that $h$ is increasing, $h<1$, and $h(z)>0$ if and only if $z>\log2$.

\proofstep{Step 2 ($b_t$ reaches the boundary in finite time).}
Let $T$ be the first index with $b_T\le W_0$ ($T=\infty$ if there is none). For $t<T$ we have $\theta_t\in\gE$ and $W_t=W_0$. Using the conserved quantity $g_t-\frac{b_t}2=g_0-\frac{b_0}2$,
\[
z_t=2b_t+g_t=\Bigl(g_0-\frac{b_0}{2}\Bigr)+\frac52b_t=z_\partial+\frac52(b_t-W_0),
\qquad\text{where}\quad z_\partial:=g_0-\frac{b_0}{2}+\frac{5W_0}{2}>\log2 .
\]
For $t<T$ we have $b_t>W_0$, hence $z_t>z_\partial$ and
\[
b_t-b_{t+1}=2\eta h(z_t)\ \ge\ 2\eta h(z_\partial)\ >\ 0 .
\]
If $T$ were infinite, summing would give $b_t\le b_0-2\eta t\,h(z_\partial)\to-\infty$, contradicting $b_t>W_0$. Thus $T<\infty$ and $W_T=W_0$. It remains to show that $\theta_T$ or $\theta_{T+1}$ lies in $\cC$, and that an invariance condition holds there.

If $b_T=W_T$, each endpoint has one preactivation equal to $0$, so the convention $\mathrm{ReLU}'(0)=0$ selects the active-cone update of \Cref{lem:gd-update-insideCone}. With $A_T=\sigma(2W_T+g_T)$ and $B_T=\sigma(-2W_T-g_T)$, this update is
\begin{equation}\label{eq:cone-entry-boundary}
W_{T+1}=W_T+2\eta B_T,\qquad b_{T+1}=W_T+2\eta(B_T-A_T),\qquad g_{T+1}=g_T+\eta(2B_T-A_T),
\end{equation}
and in particular $W_{T+1}-b_{T+1}=2\eta A_T>0$.

\proofstep{Step 3a (the case $\eta\le W_0/2$).}
Since $h<1$ and $2\eta\le W_0$, every update before the crossing satisfies
\[
b_{t+1}=b_t-2\eta h(z_t)>b_t-W_0>0\qquad(t<T).
\]
Hence $0<b_T\le W_T$.
\begin{itemize}
\item If $b_T<W_T$, then $\theta_T\in\cC$; set $\tau:=T$.
\item If $b_T=W_T$, then by \eqref{eq:cone-entry-boundary}, $W_{T+1}-b_{T+1}>0$ and, as $A_T<1$ and $B_T>0$, $b_{T+1}>W_T-2\eta\ge0$. Hence $\theta_{T+1}\in\cC$; set $\tau:=T+1$.
\end{itemize}
In either case $W_\tau\ge W_0\ge2\eta$, so \ref{I1} holds at time $\tau$, and \Cref{cor:fixed-cone} (started at time $\tau$) keeps all later iterates in $\cC$.

\proofstep{Step 3b (the case $\eta\le1$ and $2g_0+b_0\le3W_0$).}
\emph{No overshoot.} For $t<T$, the conserved quantity and $b_t\le b_0$ give
\[
2g_t+b_t=2g_0+b_0+2(b_t-b_0)\le3W_0,
\qquad\text{hence}\qquad
z_t=\frac{2g_t+b_t}{2}+\frac32b_t\le\frac32(W_0+b_t).
\]
Since $z_t>\log2>0$, inequality \eqref{eq:sigma-linear} gives $h(z_t)<z_t/3$, and with $\eta\le1$,
\[
2\eta h(z_t)\le2h(z_t)<\frac23z_t\le W_0+b_t,
\qquad\text{so}\qquad b_{t+1}=b_t-2\eta h(z_t)>-W_0 .
\]
Thus $-W_T<b_T\le W_T$.

\emph{The quantity $W+b-2g$.} It is conserved by \eqref{eq:outside-gd}, since $W$ and $b-2g$ are. Using $2g_0\le3W_0-b_0$,
\[
W_T+b_T-2g_T=W_0+b_0-2g_0\ \ge\ 2(b_0-W_0)\ >\ 0,
\]
that is, $2g_T<W_T+b_T$.
\begin{itemize}
\item If $b_T<W_T$, then $\theta_T\in\cC$ and \ref{I2} holds at time $T$; set $\tau:=T$.
\item If $b_T=W_T$, then $2g_T<2W_T$, i.e., $g_T<W_T$. In \eqref{eq:cone-entry-boundary}, $A_T-2B_T=3\sigma(2W_T+g_T)-2=h(2W_T+g_T)$. Since $h$ is increasing and $2W_T+g_T<3W_T$, \eqref{eq:sigma-linear} gives
\[
A_T-2B_T=h(2W_T+g_T)<h(3W_T)<W_T .
\]
Therefore
\[
W_{T+1}+b_{T+1}=2W_T-2\eta\,h(2W_T+g_T)>2W_T-2\eta W_T\ge0,
\]
which together with $W_{T+1}-b_{T+1}>0$ gives $\theta_{T+1}\in\cC$. One checks from \eqref{eq:cone-entry-boundary} that $W+b-2g$ is also conserved by this step, so $2g_{T+1}<W_{T+1}+b_{T+1}$ and \ref{I2} holds at time $\tau:=T+1$.
\end{itemize}
In both subcases, \Cref{cor:fixed-cone} applied from the entry time $\tau$ keeps all subsequent iterates in $\cC$. Finally, $\theta_t\in\gE$ for $t<T$, and if $\tau=T+1$ then $\theta_T$ lies on the face $\{b=|w|\}\subset\overline{\gE}$.
\end{proof}

The update identities in the proof of \Cref{lem:cone-entry} also yield an obstruction to entry. Together with the invariance of $w=0$, they give the following corollary.

\begin{corollary}[Obstructions to cone entry]\label{cor:cone-entry-obstruction}
For any sequence of positive step sizes, every initialization with $b_0>|w_0|$ and $2b_0+g_0=\log2$ is stationary. Moreover, $w_0=0$ implies $w_t=0$ for all $t\ge0$, precluding entry into $\cC$. Consequently, no choice of step sizes guarantees entry into $\cC$ from every non-dead initialization.
\end{corollary}
\begin{proof}
\emph{Stationary points.} The identities \eqref{eq:outside-gd} hold with $\eta$ replaced by $\eta_t$. If $b_0>|w_0|$ and $z_0=2b_0+g_0=\log2$, then $h(z_0)=3\sigma(\log2)-2=0$, so all updates vanish and $\theta_t=\theta_0$ for all $t$.

\emph{The invariant set $\{w=0\}$.} If $w=0$, the two hidden neurons have identical preactivations, and their contributions to the $w$-derivative cancel:
\[
\left.\partial_w f_{w,b,g}(x)\right|_{w=0}
=x\,\mathrm{ReLU}'(b)-x\,\mathrm{ReLU}'(b)=0,
\]
also at $b=0$ under our derivative convention. Hence $w_t=0$ for all $t$ by induction, which is incompatible with $|w_t|>|b_t|$.
\end{proof}

These obstructions explain why \Cref{lem:cone-entry} requires restrictions on the initialization as well as on the step size. At a stationary point in the linear region, its boundary value satisfies
\[
z_\partial=g_0-\frac{b_0}{2}+\frac{5|w_0|}{2}
=\log2-\frac52(b_0-|w_0|)<\log2,
\]
so the lemma's strict condition $z_\partial>\log2$ excludes these points, while $|w_0|>0$ excludes the invariant set $w=0$. The corollary does not assert that the lemma's sufficient conditions are necessary for entry.

\begin{proof}[Proof of \Cref{thm:finite}]
We follow the comparator argument of \citet{WuCOLT24}: in $\cC$, GD is a subgradient method on the convex function $\Ltil$, and a comparator shifted along the common separating direction $\thh$ absorbs the second-order term of each step.

\proofstep{Step 1 (subgradient and a self-bounding inequality).}
Write $d:=\thh=(2,0,-1)$ and $s_t:=-2B_t\xe-A_tx_{\xi_t}$. By \eqref{eq:gd-vector}, $\vartheta_{t+1}=\vartheta_t-\eta s_t$, and by property~(b) in the proof of \Cref{lem:piecewise}, $s_t\in\partial\Ltil(\vartheta_t)$. Since every feature has margin one under $d$ by \eqref{eq:feature-gram},
\[
\ip{s_t}{d}=-(2B_t+A_t).
\]
The Gram entries in \eqref{eq:feature-gram} give
\begin{equation}\label{eq:symmetric-subgradient-norm}
\|s_t\|^2=
\begin{cases}
12B_t^2-4A_tB_t+A_t^2, & b_t\le0,\\
12B_t^2-12A_tB_t+5A_t^2, & b_t>0.
\end{cases}
\end{equation}
Dropping the nonpositive cross terms and using $A_t,B_t\in(0,1)$,
\begin{equation}\label{eq:self-bounding}
\|s_t\|^2\le12B_t^2+5A_t^2\le12B_t+6A_t=-6\ip{s_t}{d}.
\end{equation}

\proofstep{Step 2 (one-step comparator inequality).}
Fix $R>0$ and set $u_R:=Rd$ and $\tilde u:=(R+3\eta)d$. Expanding the square and splitting $\tilde u-\vartheta_t=(u_R-\vartheta_t)+3\eta d$,
\[
\|\vartheta_{t+1}-\tilde u\|^2
=\|\vartheta_t-\tilde u\|^2+2\eta\ip{s_t}{u_R-\vartheta_t}+\eta^2\bigl(6\ip{s_t}{d}+\|s_t\|^2\bigr).
\]
The last term is nonpositive by \eqref{eq:self-bounding}. By convexity, $\ip{s_t}{u_R-\vartheta_t}\le\Ltil(u_R)-\Ltil(\vartheta_t)$, and $\Ltil(\vartheta_t)=\cL(\theta_t)$ because $\theta_t\in\cC$. Hence
\[
\|\vartheta_{t+1}-\tilde u\|^2\le\|\vartheta_t-\tilde u\|^2+2\eta\bigl(\Ltil(u_R)-\cL(\theta_t)\bigr).
\]
The comparator $u_R=(2R,0,-R)$ lies in $\cC$ and has margin $R$ on every feature, so $\Ltil(u_R)=3\ell(R)\le3e^{-R}$.

\proofstep{Step 3 (telescoping).}
Sum over $t<T$ and drop $\|\vartheta_T-\tilde u\|^2\ge0$. Since $\|\vartheta_0\|=\|\theta_0\|$ and $\|d\|=\sqrt5$, we have $\|\vartheta_0-\tilde u\|\le\|\theta_0\|+\sqrt5(R+3\eta)$, and therefore
\[
\frac1T\sum_{t<T}\cL(\theta_t)\le3e^{-R}+\frac{\bigl(\|\theta_0\|+\sqrt5(R+3\eta)\bigr)^2}{2\eta T}.
\]
Choosing $R=\log(T+1)$ gives the explicit bound
\begin{equation}\label{eq:finite-explicit}
\min_{t<T}\cL(\theta_t)\le\frac1T\sum_{t<T}\cL(\theta_t)
\le\frac3{T+1}+\frac{\bigl(\|\theta_0\|+\sqrt5(\log(T+1)+3\eta)\bigr)^2}{2\eta T}.
\end{equation}

\proofstep{Step 4 (simplification).}
For $T\ge2$ we have $\log(T+1)\le2\log T$, and by the AM--GM inequality $(\log T+\eta)^2\ge4\eta\log T\ge4\eta\log2$, so $\frac{3}{T+1}\le\frac{3(\log T+\eta)^2}{4\log2\cdot\eta T}$. Hence both terms in \eqref{eq:finite-explicit} are $O\bigl((\log T+\eta+\|\theta_0\|)^2/(\eta T)\bigr)$ with a universal constant.
\end{proof}

\subsection{Progressive Sharpening}\label{app:sharpening}

\begin{theorem}[Progressive sharpening]\label{thm:sharpening}
Fix $\eta\ge9/10$ and an integer $N\ge1$. Suppose $b_0>0$,
\[
0<W_0-b_0<\frac14,\qquad
|W_0+b_0+g_0+6\eta N|<\frac18,
\]
and either $\eta\le1$ or $W_0>2\eta$. Then, the sharpness $S_t:=\lambda_{\max}(\nabla^2\cL(\theta_t))$ satisfies $S_{t-1} < S_t$ with $S_N > 2/\eta$ for $0 \leq t \leq N$.
\end{theorem}

The proof and quantitative bounds below show that the endpoint and middle margins move closer to zero through the first EoS crossing, increasing the logistic curvature and hence the sharpness while the activation pattern remains fixed. The choice of step size and initialization guarantee that the iterates lie in $\cC$ so \Cref{thm:finite} also applies.

Throughout this subsection, assume the hypotheses of \Cref{thm:sharpening}, and write
\[
M:=-(W_0+b_0+g_0),\qquad
\Delta_t:=W_t-b_t,\qquad \zeta_t:=e^{-M+6\eta t},\qquad
E_\eta:=\frac{9\eta e^{1/8}}{e^{6\eta}-1}.
\]
In this notation, the second initialization condition reads $|\log\zeta_N|<\frac18$. We first control the margins through the prescribed crossing time, then bound the classical Hessian.

\begin{lemma}[Activity and margin control]\label{lem:sharpening-margins}
For $0\le t\le N$, we have $W_t>b_t>0$ and
\[
u_t=\log\zeta_t-\delta_t,\qquad v_t=\Delta_t-u_t,\qquad
0\le\delta_t<E_\eta<\frac1{16},\qquad
0\le\Delta_t-\Delta_0<\frac1{64}.
\]
For $t<N$, $u_t<-5$ and $v_t>5$. At the terminal iterate,
\[
|u_N|<\frac3{16},\qquad |v_N|<\frac{29}{64}<\frac12.
\]
\end{lemma}

\begin{proof}
We first show by induction that the activation pattern $W>b>0$ persists and that the endpoint margin stays very negative before time $N$; then we sum the exact updates to obtain the margin expansions.

\proofstep{Step 1 (exact dynamics in the cell $W>b>0$).}
In this cell $\xi_t=+$, so $u_t=W_t+b_t+g_t$ and $v_t=-2b_t-g_t=\Delta_t-u_t$. \Cref{lem:gd-update-insideCone} gives
\begin{align}
u_{t+1}&=u_t+\eta(6B_t-3A_t),&
v_{t+1}&=v_t-\eta(6B_t-5A_t),\nonumber\\
b_{t+1}&=b_t+2\eta(B_t-A_t),&
\Delta_{t+1}&=\Delta_t+2\eta A_t.
\label{eq:sharpening-dynamics}
\end{align}

\proofstep{Step 2 (the activation pattern persists).}
Initially $u_0=-M$ and $0<\Delta_0<\frac14$, and $|\log\zeta_N|<\frac18$ means $-M<-6\eta N+\frac18$. Suppose that, for some $t<N$,
\[
W_t>b_t>0\qquad\text{and}\qquad u_t\le-M+6\eta t .
\]
Since $t\le N-1$ and $\eta\ge\frac9{10}$,
\[
u_t\le-M+6\eta(N-1)<\tfrac18-6\eta<-5 .
\]
As $\Delta_t>0$, also $v_t=\Delta_t-u_t>5$. Consequently $B_t=\sigma(-u_t)>\frac12>\sigma(-v_t)=A_t$, and \eqref{eq:sharpening-dynamics} gives
\[
b_{t+1}>b_t>0,\qquad \Delta_{t+1}>\Delta_t>0,\qquad u_{t+1}\le u_t+6\eta\le-M+6\eta(t+1).
\]
By induction, $W_t>b_t>0$ for $0\le t\le N$, and $u_t<-5$, $v_t>5$ for $t<N$.

\proofstep{Step 3 (the corrections $\delta_t$ and $\Delta_t-\Delta_0$).}
Define $\delta_t$ by $u_t=\log\zeta_t-\delta_t$, i.e., $\delta_t=-M+6\eta t-u_t$. Summing the $u$-update in \eqref{eq:sharpening-dynamics},
\[
\delta_t=\eta\sum_{j=0}^{t-1}\bigl(6(1-B_j)+3A_j\bigr)\ \ge\ 0 .
\]
For $j<N$, each summand is at most a multiple of $\zeta_j$: using $\delta_j\ge0$ and $\Delta_j\ge\Delta_0$,
\[
1-B_j=\sigma(u_j)\le e^{u_j}\le\zeta_j,\qquad
A_j=\sigma(-v_j)\le e^{-v_j}=e^{u_j-\Delta_j}\le e^{-\Delta_0}\zeta_j .
\]
Since $\zeta_j=\zeta_0e^{6\eta j}$ is geometric, $\sum_{j<t}\zeta_j=\frac{\zeta_t-\zeta_0}{e^{6\eta}-1}$; moreover $\zeta_t\le\zeta_N<e^{1/8}$. Therefore, for $0\le t\le N$,
\begin{align*}
\delta_t&\le\eta\,(6+3e^{-\Delta_0})\,\frac{\zeta_t-\zeta_0}{e^{6\eta}-1}<\frac{9\eta e^{1/8}}{e^{6\eta}-1}=E_\eta,\\
\Delta_t-\Delta_0&=2\eta\sum_{j=0}^{t-1}A_j\le2\eta e^{-\Delta_0}\,\frac{\zeta_t-\zeta_0}{e^{6\eta}-1}<\frac{2E_\eta}{9}.
\end{align*}

\proofstep{Step 4 (numerical values).}
The function $\eta\mapsto\eta/(e^{6\eta}-1)$ is decreasing on $(0,\infty)$, so for $\eta\ge\frac9{10}$ it is at most its value at $\frac9{10}$. Using $e^{1/8}<\frac87$ and $e^{27/5}-1>150$ (truncate the exponential series after the term of order six),
\[
E_\eta\le\frac{9\cdot\frac9{10}\cdot\frac87}{150}=\frac{54}{875}<\frac1{16},\qquad
\frac{2E_\eta}{9}<\frac{12}{875}<\frac1{64}.
\]

\proofstep{Step 5 (the terminal iterate).}
Since $|\log\zeta_N|<\frac18$,
\[
|u_N|\le|\log\zeta_N|+\delta_N<\frac18+\frac1{16}=\frac3{16},\qquad
|v_N|\le\Delta_N+|u_N|<\Bigl(\frac14+\frac1{64}\Bigr)+\frac3{16}=\frac{29}{64}.
\qedhere
\]
\end{proof}

\begin{lemma}[Sharpness bounds]\label{lem:sharpening-spectrum}
For $0\le t\le N$, the sharpness in \Cref{thm:sharpening} satisfies
\[
\frac{6e^{-E_\eta}\zeta_t}{(1+\zeta_t)^2}
\le S_t\le\min\left\{(6+5e^{-\Delta_0})\zeta_t,\frac{11+\sqrt{73}}8\right\}.
\]
\end{lemma}

\begin{proof}
Let $\psi(z):=\ell''(z)=\frac{e^z}{(1+e^z)^2}$. By \Cref{lem:sharpening-margins}, every iterate with $t\le N$ lies in the open cell $W>b>0$, where the loss is smooth with Hessian
\[
\nabla^2\Ltil(\vartheta_t)=2\psi(u_t)\,\xe\xe^\top+\psi(v_t)\,\xp\xp^\top .
\]

\proofstep{Spectral sandwich.} The Rayleigh quotient at $\xe/\|\xe\|$ equals $6\psi(u_t)+3\psi(v_t)\ge6\psi(u_t)$, and the trace $6\psi(u_t)+5\psi(v_t)$ bounds the top eigenvalue of this positive semidefinite matrix. Hence
\[
6\psi(u_t)\le S_t\le6\psi(u_t)+5\psi(v_t).
\]

\proofstep{Exponential bounds.} By \Cref{lem:sharpening-margins}, $e^{u_t}=\zeta_te^{-\delta_t}\in[e^{-E_\eta}\zeta_t,\zeta_t]$ and $v_t=\Delta_t-u_t$ with $\Delta_t\ge\Delta_0$. Hence
\[
\psi(u_t)\ge\frac{e^{-E_\eta}\zeta_t}{(1+\zeta_t)^2},\qquad
\psi(u_t)\le e^{u_t}\le\zeta_t,\qquad
\psi(v_t)\le e^{-v_t}=e^{u_t-\Delta_t}\le e^{-\Delta_0}\zeta_t .
\]
Substituting into the spectral sandwich gives the exponential lower bound and the first upper bound.

\proofstep{Uniform bound.} Since $\psi\le\frac14$, we have $\nabla^2\Ltil(\vartheta_t)\preceq\frac12\xe\xe^\top+\frac14\xp\xp^\top$. Writing the latter matrix as $XDX^\top$ with $X=[\xe\ \xp]$ and $D=\mathrm{diag}(\frac12,\frac14)$, its nonzero eigenvalues are those of $D^{1/2}X^\top XD^{1/2}$, which by \eqref{eq:feature-gram} is
\[
\begin{pmatrix}\frac12\|\xe\|^2 & \frac{1}{2\sqrt2}\ip{\xe}{\xp}\\[2pt] \frac{1}{2\sqrt2}\ip{\xe}{\xp} & \frac14\|\xp\|^2\end{pmatrix}
=\begin{pmatrix}\frac32 & -\frac{3}{2\sqrt2}\\[2pt] -\frac{3}{2\sqrt2} & \frac54\end{pmatrix},
\]
which have sum $\frac{11}4$ and product $\frac34$. The larger one is $\frac{11+\sqrt{73}}8$.
\end{proof}

\begin{proof}[Proof of \Cref{thm:sharpening}]
\proofstep{Step 1 (the iterates stay in $\cC$).}
Since $M>6\eta N-\frac18>0$, we have $g_0=-M-W_0-b_0<0<\frac{W_0+b_0}{2}$. If $\eta\le1$, condition \ref{I2} holds at time $0$; if $W_0>2\eta$, condition \ref{I1} holds. By \Cref{cor:fixed-cone}, $\theta_t\in\cC$ for every $t\ge0$. By \Cref{lem:sharpening-margins}, $W_t>b_t>0$ for $t\le N$, so the classical Hessian exists at these iterates.

\proofstep{Step 2 (strict sharpening).}
We first show that both $|u_t|$ and $|v_t|$ strictly decrease for $t=0,1,\dots,N$.
\begin{itemize}
\item For $t\le N-2$: $u_t,u_{t+1}<-5$, $v_t,v_{t+1}>5$, and $B_t>\frac12>A_t$. By \eqref{eq:sharpening-dynamics}, $u_{t+1}-u_t=\eta(6B_t-3A_t)>0$ and $v_{t+1}-v_t=-\eta(6B_t-5A_t)<0$, so $|u_{t+1}|<|u_t|$ and $|v_{t+1}|<|v_t|$.
\item For $t=N-1$: $|u_{N-1}|,|v_{N-1}|>5$, while $|u_N|,|v_N|<\frac12$ by \Cref{lem:sharpening-margins}.
\end{itemize}
Since $\psi(z)=\frac{1}{4\cosh^2(z/2)}$ is strictly decreasing in $|z|$, both coefficients of
\[
\nabla^2\Ltil(\vartheta_t)=2\psi(u_t)\,\xe\xe^\top+\psi(v_t)\,\xp\xp^\top
\]
strictly increase with $t$. Thus $\nabla^2\Ltil(\vartheta_{t+1})-\nabla^2\Ltil(\vartheta_t)$ is positive definite on $\mathrm{span}\{\xe,\xp\}$, which contains the top eigenvector of $\nabla^2\Ltil(\vartheta_t)$. Hence $S_{t+1}>S_t$, i.e., $S_0<S_1<\cdots<S_N$.

\proofstep{Step 3 (below the threshold before time $N$).}
For $t<N$, $\zeta_t\le\zeta_{N-1}=\zeta_Ne^{-6\eta}<e^{1/8-6\eta}$. By the upper bound in \Cref{lem:sharpening-spectrum} and $6+5e^{-\Delta_0}<11$,
\[
S_t<11e^{1/8-6\eta}<\frac{11\cdot\frac87}{18\eta^2}=\frac{44}{63\eta^2}<\frac2\eta,
\]
using $e^{1/8}<\frac87$, $e^{6\eta}>\frac{(6\eta)^2}{2}=18\eta^2$, and $\eta\ge\frac9{10}>\frac{22}{63}$.

\proofstep{Step 4 (above the threshold at time $N$).}
Both $|u_N|$ and $|v_N|$ are below $\frac12$, so $\psi(u_N),\psi(v_N)>\frac{15}{64}$ by \eqref{eq:ellpp-lower}. With the unit vector $r=(1,2,1)/\sqrt6$ from the proof of \Cref{rem:eos-threshold},
\[
S_N\ \ge\ r^\top\nabla^2\Ltil(\vartheta_N)\,r=\frac{32\psi(u_N)+25\psi(v_N)}{6}\ >\ \frac{285}{128}\ >\ \frac{20}{9}\ \ge\ \frac2\eta .
\]

\proofstep{Step 5 (the set of initializations is open and nonempty).}
Choose $\Delta_0\in(0,\frac14)$, $W_0>\Delta_0$ (with $W_0>2\eta$ for the \ref{I1} alternative), $b_0=W_0-\Delta_0$, and $g_0=-W_0-b_0-6\eta N$. All hypotheses hold, and since they are strict inequalities, they persist under small perturbations.
\end{proof}

\subsection{Regularization and momentum}
\label{app:regularization}

We use the symmetric network \eqref{eq:model}, the summed logistic
loss, and the convention $\mathrm{ReLU}'(0)=0$. For $\lambda\ge0$,
let
\begin{equation}
\label{eq:reg-objective}
\cL_\lambda(w,b,g)
=\cL(w,b,g)+\frac\lambda2(w^2+b^2+g^2).
\end{equation}
The shared weight $w$ is penalized once. Write
$\theta=(w,b,g)$, $W=|w|$, and $\vartheta=(W,b,g)$.
For $\theta\in\cC$, this objective equals
$\Ltil_\lambda(\vartheta)=\Ltil(\vartheta)+\lambda\|\vartheta\|^2/2$,
which is $\lambda$-strongly convex when $\lambda>0$.

\paragraph{Regularization can cause an exit.}

\begin{proposition}[Regularization can break cone preservation]
\label{prop:reg-exit}
Let $(w_0,b_0,g_0)=(2\log2,\log2,0)$ and $\eta=3/4$.
For $\lambda=0$, GD stays in $\cC$ for all iterations.
For $\lambda=1$, the first iterate satisfies
\[
w_1=\frac{\log2}{2}+\frac16>0,\qquad
b_1=\frac{\log2}{4}-\frac{31}{30}<-w_1,
\]
and therefore leaves $\cC$.
\end{proposition}

\begin{proof}
The unregularized initialization satisfies
$W_0>|b_0|$ and $2g_0<W_0+b_0$, so the unit-step condition
of \Cref{lem:invariance} applies.
For $\lambda=1$, the residuals are
$B_0=\sigma(-3\log2)=1/9$ and $A_0=\sigma(2\log2)=4/5$.
Substitution into $w_1=(1-\eta)w_0+2\eta B_0$ and
$b_1=(1-\eta)b_0+2\eta(B_0-A_0)$ gives the displayed values.
Their sum is
$\frac34\log2-\frac{13}{15}<\frac34-\frac{13}{15}<0$.
Since $w_1>0$, this gives $b_1<-|w_1|$.
\end{proof}

Although $\eta\lambda<1$, the combined loss and penalty update exits.
With zero initial velocity, momentum has the same first update.

\paragraph{Conditions for cone preservation.}

\begin{theorem}[Cone preservation under regularization]
\label{thm:reg-cone}
Let $\lambda\ge0$, $\theta_0\in\cC$, and $2g_0\le W_0+b_0$.
Run GD on \eqref{eq:reg-objective} with
\begin{equation}
\label{eq:reg-gd-cap}
0<\eta_t\le\frac1{1+\lambda}\qquad(t\ge0).
\end{equation}
Then $\theta_t\in\cC$ and $2g_t\le W_t+b_t$ for every $t\ge0$,
and the sign of $w_t$ never changes.
\end{theorem}

For momentum, write $\theta_t=(w_t,b_t,g_t)$ and use
\begin{equation}
\label{eq:reg-heavy-ball}
\begin{aligned}
\theta_{t+1}
&=\theta_t+\beta(\theta_t-\theta_{t-1})
-\eta_t\nabla\cL_\lambda(\theta_t),\\
\theta_{-1}&=\theta_0,\qquad 0\le\beta<1.
\end{aligned}
\end{equation}
Thus the initial velocity is zero and $\beta$ is fixed.

\begin{corollary}[Cone preservation with momentum]
\label{cor:reg-momentum-cone}
Under the initialization assumptions of \Cref{thm:reg-cone},
the same conclusions hold for \eqref{eq:reg-heavy-ball} if
\begin{equation}
\label{eq:reg-momentum-cap}
0<\eta_t\le\frac{(1-\sqrt\beta)^2}{1+\lambda}
\qquad(t\ge0).
\end{equation}
\end{corollary}

\begin{lemma}[A logistic inequality]
\label{lem:reg-logistic}
For every $y\ge0$, $3\sigma(y)-2<y/3$.
\end{lemma}

\begin{proof}
Let $\phi(y)=y/3-3\sigma(y)+2$.
Its derivative $1/3-3\sigma(y)(1-\sigma(y))$ changes sign
once on $[0,\infty)$, from negative to positive, at
$y_*=\log((3+\sqrt5)/(3-\sqrt5))$. Hence
\[
\min_{y\ge0}\phi(y)
=\frac13\log\frac{3+\sqrt5}{3-\sqrt5}
+\frac{1-\sqrt5}{2}.
\]
For $z=\sqrt5/3$, the logarithm series gives
\[
\log\frac{1+z}{1-z}
>2\left(z+\frac{z^3}{3}+\frac{z^5}{5}\right)
=\frac{202\sqrt5}{243}
>\frac32(\sqrt5-1).
\]
The last inequality is equivalent to $729>325\sqrt5$,
verified by $729^2>5\cdot325^2$. Thus $\min\phi>0$.
\end{proof}

\paragraph{Updates of the cone gaps.}
Each endpoint contributes $-B(1,1,1)$ to the loss
gradient in $\vartheta$, while the middle point contributes
$A(0,2I,1)$, where $I=\mathbf1_{\{b>0\}}$.
In particular, $I=0$ at $b=0$ under our ReLU convention.
Adding the regularization gradient and changing the
sign gives the update direction
\begin{equation}
\label{eq:reg-force}
F_\lambda(\vartheta)
=(2B-\lambda W,\;2B-2IA-\lambda b,\;2B-A-\lambda g).
\end{equation}
To prove cone preservation, we track the two inequalities
$W-b>0$ and $W+b>0$, together with the additional
condition $(W+b)/2-g\ge0$. Define
\begin{equation}
\label{eq:reg-gaps}
\delta_1=W-b,\qquad
\delta_2=W+b,\qquad
\delta_3=\frac{W+b}{2}-g.
\end{equation}
We need to show that the first two quantities
remain positive and the third remains nonnegative.
We need $\delta_1,\delta_2>0$ and $\delta_3\ge0$.
Applying these linear coordinates to the update direction gives
\begin{equation}
\label{eq:reg-gap-forces}
\begin{aligned}
F_{\lambda,1}^{\delta}(\vartheta)&=2IA-\lambda\delta_1,\\
F_{\lambda,2}^{\delta}(\vartheta)&=4B-2IA-\lambda\delta_2,\\
F_{\lambda,3}^{\delta}(\vartheta)&=(1-I)A-\lambda\delta_3.
\end{aligned}
\end{equation}
We next bound how negative each gap's update direction
can be. Let $L=1+\lambda$. We claim that
\begin{equation}
\label{eq:reg-sector}
F_{\lambda,i}^{\delta}(\vartheta)>-L\delta_i\quad(i=1,2),
\qquad
F_{\lambda,3}^{\delta}(\vartheta)\ge-L\delta_3.
\end{equation}

For the first and third coordinates,
\[
F_{\lambda,1}^{\delta}(\vartheta)+L\delta_1
=2IA+\delta_1>0,
\qquad
F_{\lambda,3}^{\delta}(\vartheta)+L\delta_3
=(1-I)A+\delta_3\ge0.
\]
Here we used $\delta_1>0$, $\delta_3\ge0$, and
$I\in\{0,1\}$.

For the second coordinate, $\delta_2=W+b>0$.
Since $\delta_3\ge0$, we have $g\le\delta_2/2$.
Also,
\[
B=1-\sigma(\delta_2+g).
\]
If $b>0$, then $2b<W+b=\delta_2$, so
$IA=\sigma(2b+g)\le\sigma(\delta_2+g)$.
If $b\le0$, the same bound holds because $I=0$.
Therefore
\[
4B-2IA
\ge4-6\sigma(\delta_2+g)
\ge4-6\sigma(3\delta_2/2).
\]
Applying \Cref{lem:reg-logistic} at $y=3\delta_2/2$
gives $4-6\sigma(3\delta_2/2)>-\delta_2$. Hence
\[
F_{\lambda,2}^{\delta}(\vartheta)
=4B-2IA-\lambda\delta_2
>-(1+\lambda)\delta_2=-L\delta_2.
\]
This proves \eqref{eq:reg-sector}.
The next lemma controls the effect of the previous
displacement. We will apply it to each cone gap.

\begin{lemma}[Positivity with momentum]
\label{lem:reg-momentum-positive}
Let
\[
r_{t+1}=r_t+\beta(r_t-r_{t-1})+\eta_t f_t,
\]
where $0\le\beta<1$ and $\eta_t\ge0$, and define
$e_t=r_t-\sqrt\beta\,r_{t-1}$.
Suppose, for some $L>0$,
\[
r_t,e_t\ge0,\qquad f_t\ge-Lr_t,\qquad
\eta_tL\le(1-\sqrt\beta)^2.
\]
Then $r_{t+1},e_{t+1}\ge0$.
Both are strictly positive if
$\eta_t>0$ and $f_t>-Lr_t$.
\end{lemma}

\begin{proof}
Write $s=\sqrt\beta$. Substituting the recurrence gives
\begin{equation}
\label{eq:reg-positive-identity}
e_{t+1}
=\big((1-s)^2-\eta_tL\big)r_t
+s e_t+\eta_t(f_t+Lr_t).
\end{equation}
All three terms are nonnegative under the assumptions.
If $\eta_t>0$ and $f_t>-Lr_t$, the last term is positive.
Finally, $r_{t+1}=e_{t+1}+sr_t$, which gives the
same conclusions for $r_{t+1}$.
\end{proof}

\begin{proof}[Proof of \Cref{thm:reg-cone,cor:reg-momentum-cone}]
GD is the case $\beta=0$, so it suffices to prove the momentum
assertion. Suppose $w_0>0$ and define the gaps using $W_t=w_t$
until sign preservation is established. Their linearity gives
\[
\delta_{i,t+1}=\delta_{i,t}
+\beta(\delta_{i,t}-\delta_{i,t-1})
+\eta_tF_{\lambda,i}^{\delta}(\vartheta_t).
\]
Let $s=\sqrt\beta$, $L=1+\lambda$, and
$e_{i,t}=\delta_{i,t}-s\delta_{i,t-1}$.
Zero initial velocity gives
$e_{i,0}=(1-s)\delta_{i,0}\ge0$.
Assume inductively that
$\delta_{1,t},\delta_{2,t}>0$, $\delta_{3,t}\ge0$,
and $e_{i,t}\ge0$ for all three coordinates.
Apply \Cref{lem:reg-momentum-positive} separately to
each gap, with
\[
r_t=\delta_{i,t},\qquad
f_t=F_{\lambda,i}^{\delta}(\vartheta_t),
\qquad i=1,2,3.
\]
Its assumptions follow from \eqref{eq:reg-sector},
the induction hypothesis, and
$\eta_tL\le(1-s)^2$.
Thus the first two next gaps are positive, the third is nonnegative,
and all next $e_{i,t+1}$ are nonnegative. Moreover,
\[
w_{t+1}=\frac{\delta_{1,t+1}+\delta_{2,t+1}}2>0.
\]
Thus $W_{t+1}=|w_{t+1}|$, the next state is in $\cC$,
and $2g_{t+1}\le W_{t+1}+b_{t+1}$. This closes the induction.

If $w_0<0$, define $\widetilde w_t=-w_t$.
Changing the sign of $w$ swaps the two hidden neurons
and leaves both the loss and the penalty unchanged.
Consequently, $(\widetilde w_t,b_t,g_t)$ satisfies the
same update rule, with $\widetilde w_0>0$ and zero
initial velocity. The positive-weight argument above
therefore gives $\widetilde w_t>0$, so $w_t<0$ for
every $t$.
Finally, $w_0=0$ is impossible because the initialization
satisfies $|w_0|>|b_0|$.
\end{proof}

\section{PROOFS FOR THE IMPLICIT BIAS in SECTION~\ref{sec:implicit}}\label{app:implicit}

\subsection{Max-Margin Vector, Rotated Coordinates, and the Potential}

\begin{proof}[Proof of \Cref{lem:maxmargin}]
The problem is convex with a strictly convex objective, so it has at most one solution, and any point satisfying the KKT conditions is that solution. For $\thh=(2,0,-1)$:
\begin{itemize}
\item \emph{feasibility, with all constraints tight:} $\ip{\thh}{\xe}=\ip{\thh}{\xm}=\ip{\thh}{\xp}=1$;
\item \emph{stationarity with nonnegative multipliers:} $\thh=2\xe+2\xm+\xp$, with multipliers $(2,2,1)$.
\end{itemize}
Complementary slackness holds because all constraints are tight. Hence $\thh$ is the unique solution.
\end{proof}

\paragraph{Rotated coordinates.}
Let
\[
e_1:=\frac{\thh}{\sqrt5}=\frac{(2,0,-1)}{\sqrt5},\qquad e_2:=\frac{(1,0,2)}{\sqrt5},\qquad e_3:=(0,1,0).
\]
This is an orthonormal basis of $\R^3$, and the coordinates of $\vartheta=(W,b,g)$ in it are
\[
p=\ip{\vartheta}{e_1}=\frac{2W-g}{\sqrt5},\qquad q=\ip{\vartheta}{e_2}=\frac{W+2g}{\sqrt5},\qquad b=\ip{\vartheta}{e_3},\qquad \thb:=(q,b).
\]
Conversely, $W=\frac{2p+q}{\sqrt5}$ and $g=\frac{-p+2q}{\sqrt5}$. Every feature has the same component $\ip{x_s}{e_1}=\gamma=\frac1{\sqrt5}$ along $e_1$, and its orthogonal part $h_s:=(\ip{x_s}{e_2},\ip{x_s}{e_3})$ is listed in \Cref{tab:notation}. Thus every margin splits as
\[
\ip{\vartheta}{x_s}=\gamma p+z_s,\qquad z_s:=\ip{\thb}{h_s}.
\]
Projecting \eqref{eq:gd-vector} onto these coordinates gives, while $\theta_t\in\cC$,
\begin{equation}\label{eq:rot}
p_{t+1}=p_t+\eta_t\gamma\,(2B_t+A_t),\qquad \thb_{t+1}=\thb_t+\eta_t\bigl(2B_t\he+A_th_{\xi_t}\bigr).
\end{equation}
In particular, $p_t$ is non-decreasing. Projecting $\thh=2\xe+2\xm+\xp$ onto $(e_2,e_3)$, and using $\ip{\thh}{e_2}=\ip{\thh}{e_3}=0$, gives the identity
\begin{equation}\label{eq:h-relation}
2\he+2\hmid+\hp=0 .
\end{equation}

\begin{lemma}[Non-separability; the minimal feature is active]\label{lem:active-nonsep}
\leavevmode
\begin{enumerate}[label=\textup{(\alph*)},leftmargin=*,itemsep=0pt,topsep=2pt]
\item For every $\thb\in\R^2$, $\min_{s\in\{\mathrm e,-,+\}}\ip{\thb}{h_s}\le-m\|\thb\|$, where $m=\min_{\|u\|=1}\max_s(-\ip{u}{h_s})=2/\sqrt{30}$.
\item For $\thb=(q,b)$, $\min_s\ip{\thb}{h_s}=\min\{\ip{\thb}{\he},\ip{\thb}{h_{\xi(b)}}\}$.
\end{enumerate}
\end{lemma}

\begin{proof}
\proofstep{(a)} Let $\mu(u):=\max_s\bigl(-\ip{u}{h_s}\bigr)$ for $u$ on the unit circle.

\emph{$\mu$ is positive.} If $\ip{u}{h_s}\ge0$ for all $s$, then \eqref{eq:h-relation} gives $2\ip{u}{\he}+2\ip{u}{\hmid}+\ip{u}{\hp}=0$, which forces all three inner products to vanish. Then $u\perp\mathrm{span}\{\he,\hmid\}=\R^2$, so $u=0$, which is impossible on the circle. Since $\mu$ is continuous and the circle is compact, $m:=\min_{\|u\|=1}\mu(u)>0$, and the claim follows by homogeneity:
\[
\min_s\ip{\thb}{h_s}=-\|\thb\|\,\mu\bigl(\thb/\|\thb\|\bigr)\le-m\|\thb\|\qquad(\thb\neq0).
\]

\emph{Computing $m$.} Each function $u\mapsto-\ip{u}{h_s}$ has a unique local minimum on the circle, at $u=h_s/\|h_s\|$, with negative value $-\|h_s\|$. Hence, at a minimizer of $\mu>0$, the maximum defining $\mu$ is attained by at least two of the three functions (otherwise the minimizer would be a local minimizer of a single one, and $\mu$ would be negative there). It therefore suffices to evaluate $\mu$ at the six points where two of the functions coincide:
\begin{center}
\renewcommand{\arraystretch}{1.2}
\begin{tabular}{lll}
\toprule
coinciding pair & points $u$ & value of $\mu$ at $+u$,\ at $-u$\\
\midrule
$\mathrm e,\,-$ & $\pm(1,-\sqrt5)/\sqrt6$ & $2/\sqrt{30}$,\ \ $8/\sqrt{30}$\\
$\mathrm e,\,+$ & $\pm(3,-\sqrt5)/\sqrt{14}$ & $6/\sqrt{70}$,\ \ $4/\sqrt{70}$\\
$-,\,+$ & $\pm(1,0)$ & $2/\sqrt5$,\ \ $3/\sqrt5$\\
\bottomrule
\end{tabular}
\end{center}
The smallest value is $m=2/\sqrt{30}\approx0.365$, attained at $u=(1,-\sqrt5)/\sqrt6$.

\proofstep{(b)} Since $\hp=\hmid+(0,-2)$, we have $\ip{\thb}{\hp}=\ip{\thb}{\hmid}-2b$. If $b>0$, then $\hp$ has the smaller inner product, and $\xp$ is the active middle feature; if $b\le0$, then $\hmid$ has the smaller (or equal) inner product, and $\xm$ is active. Thus the smaller of the two middle-feature inner products always belongs to the active middle feature.
\end{proof}

By \Cref{lem:active-nonsep}, at every iterate the active feature $j$ with the smallest $z_j=\ip{\thb}{h_j}$ satisfies $z_j\le-m\|\thb\|\le0$; we call it the \emph{minimal active feature} and denote the other active feature by $k$.

\begin{proof}[Proof of \Cref{lem:G}]
\proofstep{Convexity and the two branches.} Since $e^{2\relu b}=\max\{1,e^{2b}\}$, we have $G=\max\{G_-,G_+\}$ with
\[
G_-(q,b):=2e^{-3q/\sqrt5-b}+e^{2q/\sqrt5},\qquad
G_+(q,b):=2e^{-3q/\sqrt5-b}+e^{2q/\sqrt5+2b}.
\]
Each branch is a sum of exponentials of linear functions, hence convex, and so is their maximum $G$. In terms of the orthogonal features, $G_\pm(\thb)=2e^{-\ip{\thb}{\he}}+e^{-\ip{\thb}{h_\pm}}$, and $G_+\ge G_-$ exactly when $b\ge0$. Hence
\begin{equation}\label{eq:G-active}
G(\thb)=\sum_{i\text{ active}}n_i\,e^{-\ip{\thb}{h_i}},
\end{equation}
where the active features are $\mathrm e$ (with $n_{\mathrm e}=2$) and $\xi(b)$ (with $n_\xi=1$).

\proofstep{Nondifferentiability.} At $b=0$, the right and left derivatives of $G$ in $b$ are $\partial_bG_+$ and $\partial_bG_-$, which differ by $2e^{2q/\sqrt5}>0$.

\proofstep{Coercivity.} Let $j$ be the minimal active feature. By \eqref{eq:G-active} and \Cref{lem:active-nonsep}(a), $G(\thb)\ge e^{-z_j}\ge e^{m\|\thb\|}$.

\proofstep{Not strongly convex.} On $\{b>0\}$, $G=G_+$ is smooth with
\[
\nabla^2G_+(\thb)=2e^{-\ip{\thb}{\he}}\,\he\he^\top+e^{-\ip{\thb}{\hp}}\,\hp\hp^\top .
\]
Let $d$ be a unit vector orthogonal to $\he$. Then
\[
d^\top\nabla^2G_+(\thb)\,d=e^{-\ip{\thb}{\hp}}\ip{\hp}{d}^2=e^{2q/\sqrt5+2b}\ip{\hp}{d}^2 .
\]
At $\thb_Q:=(-Q,1)$ this equals $e^{-2Q/\sqrt5+2}\ip{\hp}{d}^2\to0$ as $Q\to\infty$. If $G$ were $\mu$-strongly convex for some $\mu>0$, its Hessian would satisfy $\nabla^2G\succeq\mu I$ wherever $G$ is twice differentiable, in particular at every $\thb_Q$, a contradiction.

\proofstep{Minimizer.} Put $U:=e^{2q/\sqrt5+2\relu b}$ and $V:=e^{-3q/\sqrt5-b}$, so that $G=U+2V$ and $U^3V^2=e^{6\relu b-2b}$. By the AM--GM inequality applied to the five numbers $\frac U3,\frac U3,\frac U3,V,V$,
\[
G=\tfrac U3+\tfrac U3+\tfrac U3+V+V\ \ge\ 5\Bigl(\frac{U^3V^2}{27}\Bigr)^{1/5}=5\cdot3^{-3/5}\,e^{(6\relu b-2b)/5}\ \ge\ 5\cdot3^{-3/5}.
\]
The exponent $6\relu b-2b$ equals $4b$ for $b\ge0$ and $-2b$ for $b<0$, so it is nonnegative and vanishes only at $b=0$. Hence equality holds throughout if and only if $b=0$ and $\frac U3=V$, i.e., $e^{5q/\sqrt5}=3$, i.e., $q=\log3/\sqrt5$. Thus $\ths=(\log3/\sqrt5,0)$ is the unique minimizer and $\Gs=5\cdot3^{-3/5}$.
\end{proof}

\subsection{\texorpdfstring{Growth of $p$, Boundedness, and Expansions}{Growth of p, Boundedness, and Expansions}}

\begin{lemma}[$p$ becomes nonnegative]\label{lem:pnonneg}
If $\theta_t\in\cC$ and $p_t<0$, then $p_{t+1}\ge p_t+\eta_t\gamma/2$.
\end{lemma}

\begin{proof}
Let $j$ be the minimal active feature. Since $p_t<0$ and $z_j\le0$, its margin $\gamma p_t+z_j$ is negative, so its residual $\sigma(-\gamma p_t-z_j)$ exceeds $\frac12$. This residual is one of the summands of $2B_t+A_t$, hence $2B_t+A_t>\frac12$, and \eqref{eq:rot} gives $p_{t+1}-p_t=\eta_t\gamma(2B_t+A_t)\ge\eta_t\gamma/2$.
\end{proof}

\begin{lemma}[Boundedness of the orthogonal component]\label{lem:bounded}
Let $\bar\eta>0$ and $\rho(\bar\eta)=(4/e+9\bar\eta H^2)/m$. If $\theta_s\in\cC$, $p_s\ge0$ and $\eta_s\le\bar\eta$ for all $s\in[t_1,t)$, then $\|\thb_t\|\le\max\{\|\thb_{t_1}\|,\rho(\bar\eta)+3\bar\eta H\}$.
\end{lemma}

\begin{proof}
The update of $\thb$ always contains a component along the minimal active feature $h_j$, which points inward since $\ip{\thb}{h_j}=z_j\le-m\|\thb\|$. Far from the origin, this inward pull dominates both the possible outward pull of the other active feature and the second-order term of the step.

Recall that $H=\max_s\|h_s\|$. Fix $s\in[t_1,t)$ and suppress the index $s$ in $p$ and $z_i$. Let $j$ be the minimal active feature and $k$ the other active feature, and write
\[
\omega_i:=n_i\,\sigma(-\gamma p-z_i)\qquad(i\in\{j,k\}),
\]
so that \eqref{eq:rot} reads $\thb_{s+1}=\thb_s+\eta_s(\omega_jh_j+\omega_kh_k)$, and $\omega_j+\omega_k=2B_s+A_s\le3$. In particular, every step is short:
\begin{equation}\label{eq:bounded-step}
\|\thb_{s+1}-\thb_s\|\le\eta_sH(\omega_j+\omega_k)\le3\bar\eta H .
\end{equation}
We use three facts.
\begin{enumerate}[label=(\roman*),leftmargin=*,itemsep=2pt,topsep=2pt]
\item \emph{$\omega_k\le2\omega_j$.} Since $\sigma$ is increasing and $z_j\le z_k$, we have $\omega_k\le n_k\sigma(-\gamma p-z_j)=\frac{n_k}{n_j}\,\omega_j\le2\omega_j$.
\item \emph{$\omega_kz_k\le\frac4e\,\omega_j$.} This is trivial if $z_k\le0$. If $z_k>0$, then by \eqref{eq:logistic-facts} and $\max_{z>0}ze^{-z}=e^{-1}$,
\[
\omega_kz_k\le2z_ke^{-\gamma p-z_k}\le\frac2e\,e^{-\gamma p},
\qquad
\omega_j\ge\sigma(-\gamma p-z_j)\ge\sigma(-\gamma p)\ge\frac12e^{-\gamma p},
\]
where the last two inequalities use $z_j\le0$ and $\gamma p\ge0$. Combining the two bounds gives the claim.
\item \emph{$\|\omega_jh_j+\omega_kh_k\|^2\le18H^2\omega_j$.} By (i) and $\omega_j\le n_j\le2$, we get $\|\omega_jh_j+\omega_kh_k\|^2\le H^2(\omega_j+\omega_k)^2\le9H^2\omega_j^2\le18H^2\omega_j$.
\end{enumerate}
Now expand the squared norm and use $z_j\le-m\|\thb_s\|$ (\Cref{lem:active-nonsep}(a)), (ii), (iii) and $\eta_s\le\bar\eta$:
\begin{align*}
\|\thb_{s+1}\|^2-\|\thb_s\|^2
&=2\eta_s(\omega_jz_j+\omega_kz_k)+\eta_s^2\|\omega_jh_j+\omega_kh_k\|^2\\
&\le2\eta_s\omega_j\Bigl(-m\|\thb_s\|+\frac4e+9\bar\eta H^2\Bigr)
=2\eta_s\omega_j\,m\bigl(\rho(\bar\eta)-\|\thb_s\|\bigr).
\end{align*}
Hence $\|\thb_{s+1}\|\le\|\thb_s\|$ whenever $\|\thb_s\|\ge\rho(\bar\eta)$.

\emph{Induction.} Let $K:=\max\{\|\thb_{t_1}\|,\rho(\bar\eta)+3\bar\eta H\}$ and suppose $\|\thb_s\|\le K$. If $\|\thb_s\|\ge\rho(\bar\eta)$, then $\|\thb_{s+1}\|\le\|\thb_s\|\le K$; otherwise \eqref{eq:bounded-step} gives $\|\thb_{s+1}\|<\rho(\bar\eta)+3\bar\eta H\le K$. Hence $\|\thb_s\|\le K$ for all $s\in[t_1,t]$.
\end{proof}

\begin{lemma}[Expansions]\label{lem:expansions}
Let $\theta\in\cC$ with $p\ge0$ and $\|\thb\|\le M$, and let $E=e^{\gamma p}$. Then
\begin{enumerate}[label=\textup{(\roman*)},leftmargin=*,itemsep=0pt,topsep=2pt]
\item $1\le G(\thb)\le3e^{HM}$;
\item $\frac12\le E(2B+A)\le G(\thb)$ and $|E(2B+A)-G(\thb)|\le3e^{2HM}/E$;
\item $|E\,\cL(\theta)-G(\thb)|\le3e^{2HM}/(2E)$.
\end{enumerate}
\end{lemma}

\begin{proof}
Write $x_i:=\gamma p+z_i$ for the margins of the active features (with multiplicities $n_i$, $\sum_in_i=3$). By \eqref{eq:G-active},
\[
2B+A=\sum_in_i\sigma(-x_i),\qquad \cL(\theta)=\sum_in_i\ell(x_i),\qquad G(\thb)=\sum_in_ie^{-z_i}=E\sum_in_ie^{-x_i}.
\]
Since $\|\thb\|\le M$, we have $|z_i|\le HM$ for every $i$.

\proofstep{(i)} For the minimal active feature $j$, $z_j\le0$, so $G\ge e^{-z_j}\ge1$. Also $G\le\bigl(\sum_in_i\bigr)e^{HM}=3e^{HM}$.

\proofstep{(ii)} Lower bound: $E(2B+A)\ge E\,\sigma(-x_j)\ge E\cdot\frac12e^{-\gamma p}=\frac12$, exactly as in fact~(ii) of the proof of \Cref{lem:bounded}. Upper bound: $\sigma(-x)\le e^{-x}$ gives $E(2B+A)\le G$. Error: for every $x$,
\[
0\le e^{-x}-\sigma(-x)=\frac{e^{-2x}}{1+e^{-x}}\le e^{-2x},
\]
so $|E(2B+A)-G|\le E\sum_in_ie^{-2x_i}=E^{-1}\sum_in_ie^{-2z_i}\le3e^{2HM}/E$.

\proofstep{(iii)} Similarly, $0\le e^{-x}-\ell(x)\le\frac12e^{-2x}$ (from $\log(1+y)\ge y-\frac{y^2}2$ for $y\ge0$), so $|E\,\cL(\theta)-G|\le\frac E2\sum_in_ie^{-2x_i}\le3e^{2HM}/(2E)$.
\end{proof}

\subsection{A Perturbed Subgradient Method}

\begin{lemma}[Perturbed subgradient method]\label{lem:subgrad}
Let $G:\R^d\to\R$ be convex with a unique minimizer $\ths$. Let $(\thb_t)$ be bounded with $\thb_{t+1}=\thb_t-\alpha_t(s_t-\varepsilon_t)$, where $s_t\in\partial G(\thb_t)$, $\sup_t\|s_t\|<\infty$, $\alpha_t\ge0$, $\sum_t\alpha_t=\infty$, $\sum_t\alpha_t^2<\infty$ and $\sum_t\alpha_t\|\varepsilon_t\|<\infty$. Then $\thb_t\to\ths$.
\end{lemma}

\begin{proof}
Let $D_t:=\|\thb_t-\ths\|^2$ and $\Gs:=G(\ths)$, and let $R:=\sup_t\|\thb_t-\ths\|$ and $C:=\sup_t\|s_t\|$, both finite by assumption.

\proofstep{Step 1 (one-step inequality).} Expanding the update,
\[
D_{t+1}=D_t-2\alpha_t\ip{s_t}{\thb_t-\ths}+2\alpha_t\ip{\varepsilon_t}{\thb_t-\ths}+\alpha_t^2\|s_t-\varepsilon_t\|^2 .
\]
By convexity, $\ip{s_t}{\thb_t-\ths}\ge G(\thb_t)-\Gs$; by Cauchy--Schwarz, the third term is at most $2R\alpha_t\|\varepsilon_t\|$; and $\|s_t-\varepsilon_t\|^2\le2C^2+2\|\varepsilon_t\|^2$. Hence
\begin{equation}\label{eq:subgrad-step}
D_{t+1}\le D_t-2\alpha_t\bigl(G(\thb_t)-\Gs\bigr)+\beta_t,\qquad
\beta_t:=2R\alpha_t\|\varepsilon_t\|+2C^2\alpha_t^2+2\alpha_t^2\|\varepsilon_t\|^2 .
\end{equation}
The errors are summable, $\sum_t\beta_t<\infty$, because $\sum_t\alpha_t\|\varepsilon_t\|<\infty$, $\sum_t\alpha_t^2<\infty$, and $\sum_t(\alpha_t\|\varepsilon_t\|)^2\le(\sum_t\alpha_t\|\varepsilon_t\|)^2$.

\proofstep{Step 2 ($D_t$ converges, and $G(\thb_t)$ approaches $\Gs$ along a subsequence).} Since $G(\thb_t)\ge\Gs$, \eqref{eq:subgrad-step} shows that $D_t+\sum_{r\ge t}\beta_r$ is non-increasing. It is bounded below by $0$, so it converges, and since $\sum_{r\ge t}\beta_r\to0$, $D_t$ converges. Summing \eqref{eq:subgrad-step} also gives
\[
2\sum_t\alpha_t\bigl(G(\thb_t)-\Gs\bigr)\le D_0+\sum_t\beta_t<\infty,
\]
which, together with $\sum_t\alpha_t=\infty$, forces $\liminf_tG(\thb_t)=\Gs$.

\proofstep{Step 3 (the limit is $\ths$).} Choose a subsequence with $G(\thb_{t_k})\to\Gs$. Since $(\thb_t)$ is bounded, a further subsequence converges to some $\thb'$, and $G(\thb')=\Gs$ by continuity of the finite convex function $G$. By uniqueness of the minimizer, $\thb'=\ths$. Hence $\liminf_tD_t=0$, and since $D_t$ converges, $D_t\to0$.
\end{proof}

Uniqueness is essential: \Cref{lem:subgrad} replaces strong convexity. Equivalently, the unique minimizer of a finite convex function on $\R^d$ is a \emph{strong} minimizer, i.e., $G(\thb_t)\to\Gs$ implies $\thb_t\to\ths$ \citep{PlanidenAndWang2015SIAM-StrongMinimizers}.

\subsection{\texorpdfstring{Proof of \Cref{thm:implicit}}{Proof of the implicit-bias theorem}}

\begin{proof}[Proof of \Cref{thm:implicit}]
Throughout, the step size is the constant $\eta$ and $E_t:=e^{\gamma p_t}$. The proof has four steps: (1) after a finite time, $p_t\ge0$ and $\thb_t$ stays bounded; (2) consequently $E_t$ grows linearly in $t$; (3) the orthogonal component $\thb_t$ runs a perturbed subgradient method on $G$ with step sizes of order $1/t$, so it converges to $\ths$; (4) this pins down the growth rate of $E_t$, from which parts (a)--(c) follow.

\proofstep{Step 1 (a trapping region).} By \Cref{lem:pnonneg}, while $p_t<0$ it increases by at least $\eta\gamma/2$ per step, and $p_t$ is non-decreasing. Hence $p_t\ge0$ for all $t\ge t_1:=\lceil2\relu{-p_0}/(\eta\gamma)\rceil$. Applying \Cref{lem:bounded} with $\bar\eta=\eta$ from time $t_1$ gives
\[
M:=\sup_{t\ge t_1}\|\thb_t\|\le\max\bigl\{\|\thb_{t_1}\|,\ \rho(\eta)+3\eta H\bigr\}<\infty .
\]
From now on $t\ge t_1$, so \Cref{lem:expansions} applies with this $M$.

\proofstep{Step 2 ($E_t$ grows linearly).} Let $\chi_t:=\gamma(p_{t+1}-p_t)=\log(E_{t+1}/E_t)$. By \eqref{eq:rot} and \Cref{lem:expansions}(i)--(ii),
\[
\chi_t=\eta\gamma^2(2B_t+A_t)\in\Bigl[\frac{\eta\gamma^2}{2E_t},\ \frac{3\eta\gamma^2e^{HM}}{E_t}\Bigr].
\]
Since $E_{t+1}-E_t=E_t(e^{\chi_t}-1)$, the bounds $x\le e^x-1\le xe^x$ and $E_t\ge1$ give
\[
E_t+\tfrac12\eta\gamma^2\ \le\ E_{t+1}\ \le\ E_t+K,\qquad K:=3\eta\gamma^2e^{HM}\exp\bigl(3\eta\gamma^2e^{HM}\bigr).
\]
Hence $E_t\asymp t$, that is, $c_1t\le E_t\le c_2t$ for some constants $c_1,c_2>0$ and all large $t$.

\proofstep{Step 3 ($\thb_t$ is a perturbed subgradient method on $G$).} Define
\[
\alpha_t:=\frac{\eta}{E_t},\qquad
s_t:=-\sum_{i\text{ active}}n_ie^{-z_{i,t}}h_i,\qquad
\varepsilon_t:=\sum_{i\text{ active}}n_i\bigl(E_t\,\sigma(-\gamma p_t-z_{i,t})-e^{-z_{i,t}}\bigr)h_i .
\]
Multiplying and dividing by $E_t$ in \eqref{eq:rot},
\[
\thb_{t+1}=\thb_t+\frac{\eta}{E_t}\sum_{i\text{ active}}n_i\,E_t\sigma(-\gamma p_t-z_{i,t})\,h_i=\thb_t-\alpha_t(s_t-\varepsilon_t).
\]
We check the hypotheses of \Cref{lem:subgrad}.
\begin{itemize}
\item \emph{$s_t$ is a subgradient of $G$.} By \eqref{eq:G-active}, $s_t=\nabla G_{\xi_t}(\thb_t)$. If $b_t\ne0$, then $G=G_{\xi_t}$ near $\thb_t$, so $s_t=\nabla G(\thb_t)$. If $b_t=0$, then $G_-(\thb_t)=G_+(\thb_t)$, so $\partial G(\thb_t)=\mathrm{conv}\{\nabla G_-(\thb_t),\nabla G_+(\thb_t)\}$, which contains $\nabla G_-(\thb_t)=s_t$ (recall $\xi_t=-$ when $b_t=0$).
\item \emph{Bounded iterates and subgradients.} $\|\thb_t\|\le M$ by Step~1, and $\|s_t\|\le H\sum_in_ie^{-z_{i,t}}\le3He^{HM}$.
\item \emph{Step sizes and errors.} As in the proof of \Cref{lem:expansions}(ii), $\|\varepsilon_t\|\le3He^{2HM}/E_t$. Since $E_t\asymp t$, we get $\alpha_t\asymp1/t$ and $\alpha_t\|\varepsilon_t\|=O(1/t^2)$, so $\sum_t\alpha_t=\infty$, $\sum_t\alpha_t^2<\infty$, and $\sum_t\alpha_t\|\varepsilon_t\|<\infty$.
\end{itemize}
Since $G$ has the unique minimizer $\ths$ (\Cref{lem:G}), \Cref{lem:subgrad} gives $\thb_t\to\ths$.

\proofstep{Step 4 (the exact growth rate of $E_t$).} By continuity of $G$, $G(\thb_t)\to G(\ths)=\Gs$, and therefore $E_t(2B_t+A_t)\to\Gs$ by \Cref{lem:expansions}(ii). In particular $\chi_t\to0$, so $e^{\chi_t}-1=\chi_t(1+o(1))$ and
\[
E_{t+1}-E_t=E_t\bigl(e^{\chi_t}-1\bigr)=\eta\gamma^2E_t(2B_t+A_t)\bigl(1+o(1)\bigr)\ \to\ \eta\gamma^2\Gs=\eta\,3^{-3/5}.
\]
By the Stolz--Ces\`aro theorem, $E_t/t\to\eta\,3^{-3/5}$, that is,
\begin{equation}\label{eq:implicit-p}
\gamma p_t=\log t+\log\bigl(\eta\,3^{-3/5}\bigr)+o(1).
\end{equation}

\proofstep{Part (a): parameter asymptotics.} Since $e_1=\gamma\thh$, the rotated decomposition reads $\vartheta_t=\gamma p_t\,\thh+q_te_2+b_te_3$. Using \eqref{eq:implicit-p}, $q_t\to q^\star=\log3/\sqrt5$, and $b_t\to0$,
\[
\vartheta_t-\thh\log t\ \to\ \log\bigl(\eta\,3^{-3/5}\bigr)(2,0,-1)+\frac{\log3}{5}(1,0,2)=\Bigl(\log\frac{\eta^2}{3},\ 0,\ \log\frac3\eta\Bigr).
\]
Since $\vartheta_t=(|w_t|,b_t,g_t)$ and $\|\theta_t\|=\|\vartheta_t\|=\sqrt5\log t\,(1+o(1))$, this gives both statements in (a).

\proofstep{Margins and part (b).} The endpoint margin is $u_t=\gamma p_t+z_{\mathrm e,t}$, where $z_{\mathrm e,t}\to\ip{\ths}{\he}=\frac{3}{\sqrt5}q^\star=\frac35\log3$. By \eqref{eq:implicit-p},
\[
u_t=\log(\eta t)+o(1).
\]
For the middle point, both orthogonal margins have the same limit because $b^\star=0$: $\ip{\ths}{\hmid}=\ip{\ths}{\hp}=-\frac{2}{\sqrt5}q^\star=-\frac25\log3$. Hence, regardless of the active branch,
\[
v_t=\log(\eta t)-\log3+o(1).
\]
Finally, \Cref{lem:expansions}(iii) gives $E_t\cL(\theta_t)\to\Gs$, so
\[
t\,\cL(\theta_t)=\frac{t}{E_t}\cdot E_t\cL(\theta_t)\ \to\ \frac{\Gs}{\eta\,3^{-3/5}}=\frac5\eta .
\]

\proofstep{Part (c): the learned function.} For $|x|\le1$, put $a:=W|x|\ge0$, so that $f_\theta(x)=\relu{a+b}+\relu{-a+b}+g$. Since $\mathrm{ReLU}$ is $1$-Lipschitz and $\relu{a}+\relu{-a}=a$,
\begin{equation}\label{eq:f-linearization}
\bigl|f_\theta(x)-(W|x|+g)\bigr|\le2|b|\qquad\text{for all }|x|\le1 .
\end{equation}
Substituting $W_t=2\log t+\log\frac{\eta^2}{3}+o(1)$, $g_t=-\log t+\log\frac3\eta+o(1)$ and $b_t\to0$ from part (a),
\[
f_{\theta_t}(x)=(2|x|-1)\log t+|x|\log\frac{\eta^2}{3}+\log\frac3\eta+o(1)=(2|x|-1)\log(\eta t)+(1-|x|)\log3+o(1),
\]
uniformly over $|x|\le1$. The endpoint and middle margins in (c) were computed above.
\end{proof}

\section{LOSS SPIKES AND RECOVERY WITH LARGE STEP SIZES}
\label{app:init-phase}
\label{app:init-catapult}

We prove the two-step catapult theorem and a five-step
recovery result on an expanded initialization set.
Throughout, the initialization is fixed as $\eta\to\infty$.
We use the unregularized tied model, the summed logistic
loss, and $\mathrm{ReLU}'(0)=0$. In the reflected coordinates
$\vartheta=(W,b,g)$, with $W=|w|$, recall
\[
u=W+b+g,\qquad v=-2[b]_+-g,\qquad
B=\sigma(-u),\qquad A=\sigma(-v).
\]
Inside $\cC=\{W>|b|\}$, differentiation gives
\begin{equation}
\label{eq:catapult-update}
\begin{aligned}
W'&=W+2\eta B,\\
b'&=b+2\eta(B-A\mathbf1_{\{b>0\}}),\\
g'&=g+\eta(2B-A).
\end{aligned}
\end{equation}
The two cone gaps therefore satisfy
\begin{equation}
\label{eq:catapult-gaps}
\begin{aligned}
W'-b'&=W-b+2\eta A\mathbf1_{\{b>0\}},\\
W'+b'&=W+b+4\eta B-2\eta A\mathbf1_{\{b>0\}}.
\end{aligned}
\end{equation}
In particular, $W-b$ never decreases while the iterates
are active.

\begin{proposition}[Recovery in two or five updates]
\label{prop:finite-step-recovery}
Fix $\theta_0$ independently of $\eta$,
with
\[
0<b_0<W_0,\qquad B_0<A_0<2B_0.
\]
Let $\nu=2B_0-A_0$ and suppose $\nu\ne1/3$.
For every sufficiently large constant step size $\eta$,
\[
\frac{\cL(\theta_1)}{\eta}\longrightarrow\nu>0.
\]
Moreover:
\begin{enumerate}
\item If $\nu>1/3$, then
$\theta_0,\theta_1,\theta_2\in\cC$ and
\[
\frac{(u_2,v_2)}{\eta}\longrightarrow(3\nu-1,1-\nu),
\qquad
-\frac{\log\cL(\theta_2)}{\eta}\longrightarrow3\nu-1.
\]
\item If $0<\nu<1/3$, then
$\theta_0,\ldots,\theta_5\in\cC$ and
\[
\frac{(u_5,v_5)}{\eta}\longrightarrow(3\nu+1,1-\nu),
\qquad
-\frac{\log\cL(\theta_5)}{\eta}\longrightarrow1-\nu.
\]
\end{enumerate}
All limits are taken as $\eta\to\infty$.
\end{proposition}

\begin{proof}
Since $u_0+v_0=W_0-b_0>0$, we have $A_0+B_0<1$.
Together with $B_0<A_0<2B_0$, this gives
\[
0<\nu<B_0<1/2.
\]
We calculate the first two updates for both cases.
If the endpoint margin is still negative, we calculate
three more. 
\\
By \eqref{eq:catapult-update},
\[
W_1=W_0+2\eta B_0,\qquad
b_1=b_0-2\eta(A_0-B_0),\qquad
g_1=g_0+\eta\nu.
\]
For sufficiently large $\eta$, $b_1<0$. Also,
\[
W_1-b_1=W_0-b_0+2\eta A_0>0,\qquad
W_1+b_1=W_0+b_0+2\eta\nu>0,
\]
so $\theta_1\in\cC$. The endpoint and middle margins are
\[
u_1=u_0+3\eta\nu,\qquad v_1=-g_0-\eta\nu.
\]
The endpoint margin becomes large and positive, while
the middle margin becomes large and negative.
Consequently, $B_1\to0$, $A_1\to1$, and
\[
\frac{\cL(\theta_1)}{\eta}
=\frac{2\ell(u_1)+\ell(v_1)}{\eta}
\longrightarrow\nu.
\]

\paragraph{Second update.}
Since $b_1<0$, neither neuron activates at the middle
point. The update is
\[
W_2=W_1+2\eta B_1,\qquad
b_2=b_1+2\eta B_1,\qquad
g_2=g_1+\eta(2B_1-A_1).
\]
Thus $g$ decreases by $\eta+o(\eta)$, while the changes
in $W$ and $b$ are $o(\eta)$.

Write $d=2(B_0-A_0)$. Then $-2<d<0$, and
\[
\frac{\vartheta_2}{\eta}\longrightarrow(2B_0,d,\nu-1).
\]
Both cone gaps remain positive:
\[
W_2-b_2=W_1-b_1>0,\qquad
W_2+b_2=W_1+b_1+4\eta B_1>0.
\]
The bias remains negative for sufficiently large $\eta$,
so
\[
\frac{u_2}{\eta}\longrightarrow3\nu-1,\qquad
\frac{v_2}{\eta}\longrightarrow1-\nu.
\]
If $\nu>1/3$, both limits are positive. This completes
the two-step case.

\paragraph{When $\nu<1/3$, three more updates are needed.}
Now the correction at time two has made the endpoint
margin negative, while the middle margin is positive.
Thus $B_2\to1$ and $A_2\to0$. The third update gives
\[
\frac{\vartheta_3}{\eta}
\longrightarrow(2B_0+2,d+2,\nu+1).
\]
Its bias is positive because $d+2>0$. Its lower cone
gap satisfies
\[
\frac{W_3+b_3}{\eta}\longrightarrow2\nu+4>0,
\]
and its upper gap remains positive by
\eqref{eq:catapult-gaps}. Hence $\theta_3\in\cC$.
Its margin limits are
\[
\frac{u_3}{\eta}\longrightarrow3\nu+5>0,\qquad
\frac{v_3}{\eta}\longrightarrow2A_0-3\nu-5<0.
\]
The last inequality follows from $A_0<1$ and $\nu>0$.

The middle point is therefore misclassified again:
$B_3\to0$ and $A_3\to1$. Because $b_3>0$, the fourth
update decreases $b$ by $2\eta+o(\eta)$ and $g$ by
$\eta+o(\eta)$:
\[
\frac{\vartheta_4}{\eta}
\longrightarrow(2B_0+2,d,\nu).
\]
The bias is now negative. The lower cone gap tends to
$2\nu+2>0$ after division by $\eta$, and the upper
gap remains positive. Thus $\theta_4\in\cC$, with
\[
\frac{u_4}{\eta}\longrightarrow3\nu+2>0,\qquad
\frac{v_4}{\eta}\longrightarrow-\nu<0.
\]

Hence $B_4\to0$ and $A_4\to1$. Since $b_4<0$, the
fifth update decreases only $g$ at leading order:
\[
\frac{\vartheta_5}{\eta}
\longrightarrow(2B_0+2,d,\nu-1).
\]
Both cone gaps remain positive under this
negative-bias update. The bias remains negative, and
\[
\frac{u_5}{\eta}\longrightarrow3\nu+1>0,\qquad
\frac{v_5}{\eta}\longrightarrow1-\nu>0.
\]
This completes the five-step calculation.

\paragraph{The loss at the final step.}
The calculations above show that, for sufficiently large
$\eta$, every iterate through time $k$ lies in $\cC$,
where $k=2$ or $k=5$. Each bias and margin used to
determine an update has a nonzero limit after division
by $\eta$, so its sign holds for all sufficiently large
$\eta$. Taking the largest of the finitely many required
thresholds gives a single threshold for the entire
calculation.

At the final step, both margins are positive and grow
linearly with $\eta$. If
$u_k/\eta\to U>0$ and $v_k/\eta\to V>0$, then
\[
\ell(u_k)=e^{-\eta U+o(\eta)},\qquad
\ell(v_k)=e^{-\eta V+o(\eta)}.
\]
The smaller margin therefore determines the exponential
rate of the total loss:
\[
-\frac1\eta\log\cL(\theta_k)\to\min\{U,V\}.
\]
For the two-step case, $3\nu-1<1-\nu$ because $\nu<1/2$,
so the limit is $3\nu-1$. For the five-step case,
$1-\nu<3\nu+1$ because $\nu>0$, so the limit is $1-\nu$.
\end{proof}

\begin{proof}[Proof of \Cref{thm:catapult}]
With $B_0=\sigma(-u_0)$ and $A_0=\sigma(-v_0)$,
the assumptions of \Cref{thm:catapult} imply those of
\Cref{prop:finite-step-recovery}, with $\nu>1/3$.
Its first case gives the stated conclusions.
\end{proof}

\begin{figure}[t]
\centering
\includegraphics[width=\columnwidth]{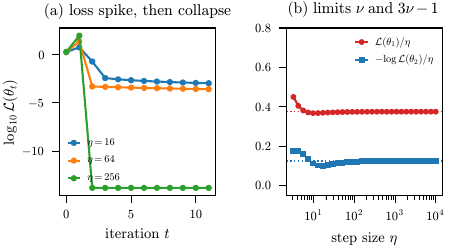}
\caption{\textbf{Two-step catapult} from
$\theta_0=(0.2,0.1,-0.1)$, for which $\nu\approx0.375$.
(a) The loss increases after the first update and drops
after the second.
(b) The normalized losses approach the limits in
\Cref{thm:catapult}.}
\label{fig:catapult-asymptotics}
\end{figure}

\section{PROOFS FOR THE ADAPTIVE SCHEDULES}\label{app:adaptive}

\begin{lemma}[Divergence under the adaptive schedule]\label{lem:adaptive-divergence}
Let $\theta_0\in\cC$, $0<\kappa\le1/2$, and $\eta_t=\kappa W_t$. Then $W_t\to\infty$ and $\eta_t\to\infty$.
\end{lemma}

\begin{proof}
Since $\eta_t=\kappa W_t\le W_t/2$, condition \ref{I1} holds at every step, so all iterates stay in $\cC$ by \Cref{lem:invariance}. By \Cref{lem:gd-update-insideCone},
\[
W_{t+1}-W_t=2\eta_tB_t\ge0,\qquad (W_{t+1}-g_{t+1})-(W_t-g_t)=\eta_tA_t\ge0,
\]
so $W_t$ is non-decreasing and $g_t\le g_0+W_t-W_0$.

Suppose, for contradiction, that $W_t\to\bar W<\infty$. Since $W_{t+1}-W_t=2\kappa W_tB_t\ge2\kappa W_0B_t$, summing over $t$ gives $\sum_tB_t\le(\bar W-W_0)/(2\kappa W_0)<\infty$. Hence $B_t=\sigma(-u_t)\to0$, i.e., $u_t\to\infty$. But $u_t$ is bounded above: using $|b_t|<W_t$,
\[
u_t=W_t+b_t+g_t<2W_t+g_0+W_t-W_0\le3\bar W+g_0-W_0 .
\]
This contradiction shows $W_t\to\infty$, and therefore $\eta_t=\kappa W_t\to\infty$.
\end{proof}

\subsection{Finite-Time Convergence with Increasing Step Sizes}\label{app:increasing}

The split-comparator proof of \citet[Lemma~7]{WuCOLT24}, also used in \Cref{thm:finite}, writes a fixed comparator as $u=u_1+u_2$: $u_1$ has small loss, while $u_2$ is a separating-direction shift proportional to the fixed step size and absorbs the quadratic update term. Substituting $\eta_t=\kappa W_t$ into this shift would introduce a moving comparator and additional drift terms. Here we separate the argument into a bound on cumulative loss in terms of $W_T$ (\Cref{lem:increasing-loss}) and an independent bound on the growth of $W_T$ (\Cref{lem:increasing-growth}).

We first record two scalar inequalities for the logistic loss and its residual.

\begin{samepage}
\begin{lemma}[Logistic loss bound]\label{lem:logistic-loss-bound}
For every $z\in\R$,
\[
\ell(z)\le2(1+\relu{-z})\sigma(-z).
\]
\end{lemma}

\begin{proof}
If $z\ge0$, then $\relu{-z}=0$ and $e^{-z}\le1$. Using $\log(1+x)\le x$ for $x\ge0$ gives
\[
\ell(z)\le e^{-z}\le\frac{2e^{-z}}{1+e^{-z}}=2\sigma(-z).
\]
If $z<0$, then $\relu{-z}=-z$. Since $e^z<1$,
\[
\ell(z)=-z+\log(1+e^z)\le-z+\log2\le1-z.
\]
Also $\sigma(-z)\ge1/2$, so $\ell(z)\le1-z\le2(1-z)\sigma(-z)$, proving the claim.
\end{proof}
\end{samepage}

\begin{lemma}[Logistic residual drift bound]\label{lem:logistic-drift-bound}
For every $z\in\R$ and $h\ge0$, let $r=\sigma(-z)$. Then
\[
r\bigl(z-h(1-r)\bigr)\le e^{-h/2-1}.
\]
\end{lemma}

\begin{proof}
If $z\le0$, then $z-h(1-r)\le z\le0$, so the left-hand side is nonpositive. If $0<z\le h/2$, then $1-r=\sigma(z)\ge1/2$, giving $z-h(1-r)\le z-h/2\le0$ and the same conclusion.

If $z>h/2$, then $1-r\ge1/2$ and $r\le e^{-z}$. Since $z-h/2>0$, we obtain
\[
r\bigl(z-h(1-r)\bigr)
\le r(z-h/2)
\le e^{-z}(z-h/2).
\]
Setting $x=z-h/2>0$ gives
\[
e^{-z}(z-h/2)=e^{-h/2}xe^{-x}\le e^{-h/2-1},
\]
because $\frac{d}{dx}(xe^{-x})=e^{-x}(1-x)$, so $xe^{-x}$ attains its maximum $1/e$ at $x=1$.
\end{proof}

The next two lemmas prove \Cref{thm:increasing}. For the remaining results in this subsection, let $\theta_0\in\cC$, $0<\kappa\le1/4$, and $\eta_t=\kappa W_t$. Define
\[
\Sigma_T:=\sum_{t=0}^{T-1}\cL(\theta_t),\qquad \Sigma_0:=0,\qquad
M:=\max\left\{3,1+\frac{|g_0|}{W_0}\right\}.
\]

\begin{lemma}[Cumulative loss is controlled by the endpoint weight]\label{lem:increasing-loss}
Every iterate remains in $\cC$, $W_t\ge W_0>0$, and $|g_t|\le MW_t$. For every integer $T\ge1$,
\[
\Sigma_T\le K_{\mathrm{loss}}W_T,\qquad
K_{\mathrm{loss}}:=\frac{4(M+2)}{\kappa}\left(\frac1{W_0}+M+2\right).
\]
\end{lemma}

\begin{proof}
We first bound the coordinates in terms of $W_t$, then control the loss by $W_t(2B_t+A_t)$, and finally sum an exact update identity.

\proofstep{Step 1 (coordinate bounds).}
Since $\eta_t=\kappa W_t\le W_t/2$, \Cref{lem:invariance} keeps all iterates in $\cC$, so $|b_t|<W_t$. It remains to establish $W_t\ge W_0$ and $|g_t|\le MW_t$. The updates in \Cref{lem:gd-update-insideCone} give
\begin{align}
W_{t+1}&=W_t+2\eta_t B_t = W_t+2\kappa W_tB_t=W_t(1+2\kappa B_t)>W_t,\label{eq:increasing-w-update}\\
g_{t+1} - g_t &=\eta_t(2B_t-A_t)=\kappa W_t(2B_t-A_t) = 2\kappa W_tB_t - \kappa W_tA_t = W_{t+1}-W_t - \kappa W_tA_t.\label{eq:increasing-g-update}
\end{align}
Therefore, $W_t$ is non-decreasing. To get the upper bound $g_t \leq MW_t$ we sum \eqref{eq:increasing-g-update} over $s=0,\ldots,t-1$ and telescoping gives
\begin{align*}
g_t-g_0
&=\sum_{s=0}^{t-1}(g_{s+1}-g_s)\\
&=\sum_{s=0}^{t-1}(W_{s+1}-W_s)-\kappa\sum_{s=0}^{t-1}W_sA_s\\
&=W_t-W_0-\kappa\sum_{s=0}^{t-1}W_sA_s\le W_t-W_0,
\end{align*}
since $\kappa W_sA_s\ge0$. Using $W_t\ge W_0>0$ and $M\ge1+|g_0|/W_0$, we obtain
\[
g_t\le g_0+W_t-W_0
\le W_t+|g_0|
\le\left(1+\frac{|g_0|}{W_0}\right)W_t
\le MW_t.
\]
To prove the lower bound $g_t \geq -MW_t$, we use induction. Since $W_0>0$ and the choice of $M$ gives $M\ge1+|g_0|/W_0$, we have $MW_0\ge|g_0|$. Thus the base case follows from $g_0\ge-|g_0|\ge-MW_0$. Assume $g_t\ge-MW_t$; we show that $g_{t+1}\ge-MW_{t+1}$. Consider two cases: $g_t\ge-2W_t$ or $g_t<-2W_t$.

If $g_t\ge-2W_t$, then
\begin{equation*}
  g_{t+1} = g_t +  W_{t+1}-W_t - \kappa W_tA_t
  \geq g_t-\kappa W_t\ge-(2+\kappa)W_t\ge-3W_{t+1}\ge-MW_{t+1}
\end{equation*}

If $g_t<-2W_t$, then $\theta_t\in\cC$ gives $|b_t|<W_t$, hence $\relu{b_t}<W_t$. Using the definitions of the margins,
\begin{align*}
u_t&=W_t+b_t+g_t<2W_t+g_t<0,\\
-v_t&=2\relu{b_t}+g_t<2W_t+g_t<0,
\end{align*}
where the last inequality in each line uses $g_t<-2W_t$. Consequently, $B_t=\sigma(-u_t)>1/2$ and $A_t=\sigma(-v_t)<1/2$. Thus $2B_t-A_t>0$, so the update increases $g_t$. Using the induction hypothesis $g_t\ge-MW_t$ and $W_{t+1}\ge W_t$, we obtain
\[
g_{t+1}=g_t+\kappa W_t(2B_t-A_t)>g_t\ge-MW_t\ge-MW_{t+1}.
\]
This completes the induction and proves $|g_t|\le MW_t$.

\proofstep{Step 2 (a loss bound in terms of the residuals).}
In the active cone, the two endpoints $x=\pm1$ have label $+1$ and share the signed margin $u_t=W_t+b_t+g_t$. The middle point $x=0$ has label $-1$, so its signed margin is $v_t=-2\relu{b_t}-g_t$. Thus the loss decomposes as
\begin{align*}
\cL(\theta_t)
&=\ell(f_{\theta_t}(-1))+\ell(-f_{\theta_t}(0))+\ell(f_{\theta_t}(1))\\
&=2\ell(u_t)+\ell(v_t).
\end{align*}
Using the definitions of the margins, the triangle inequality, and $\relu{b_t}\le|b_t|$, we have
\begin{align*}
|u_t|&=|W_t+b_t+g_t|\le W_t+|b_t|+|g_t|\le(M+2)W_t,\\
|v_t|&=|-2\relu{b_t}-g_t|\le2\relu{b_t}+|g_t|\le2|b_t|+|g_t|\le(M+2)W_t,
\end{align*}
where the final inequalities use $|b_t|<W_t$ and $|g_t|\le MW_t$. In particular, $\relu{-u_t}$ and $\relu{-v_t}$ are also at most $(M+2)W_t$. Applying \Cref{lem:logistic-loss-bound} to each margin and recalling $B_t=\sigma(-u_t)$ and $A_t=\sigma(-v_t)$ gives
\begin{align}
\cL(\theta_t)
&=2\ell(u_t)+\ell(v_t)\nonumber\\
&\le4(1+\relu{-u_t})B_t+2(1+\relu{-v_t})A_t\nonumber\\
&\le2\bigl(1+(M+2)W_t\bigr)(2B_t+A_t)\nonumber\\
&\le2\left(\frac1{W_0}+M+2\right)W_t(2B_t+A_t),\label{eq:increasing-loss-bound}
\end{align}
where the last step uses $W_t\ge W_0>0$, so $1\le W_t/W_0$.

\proofstep{Step 3 (telescoping the cumulative loss bound).}
To express $W_t(2B_t+A_t)$ in terms of parameter increments, multiply the $W_t$ increment in \eqref{eq:increasing-w-update} by two and subtract the $g_t$ increment in \eqref{eq:increasing-g-update}. This gives
\begin{align*}
(2W_{t+1}-g_{t+1})-(2W_t-g_t)
&=2(W_{t+1}-W_t)-(g_{t+1}-g_t)\\
&=4\kappa W_tB_t-\kappa W_t(2B_t-A_t)\\
&=\kappa W_t(2B_t+A_t).
\end{align*}
This is progress along the common separating direction $\thh=(2,0,-1)$. The choice $\eta_t=\kappa W_t$ matches the factor $W_t$ in the loss bound, so this linear coordinate directly controls the unweighted cumulative loss, without a squared-distance comparator estimate.
Summing and using the coordinate bounds yields
\begin{equation}\label{eq:increasing-sum-bound}
\kappa\sum_{t=0}^{T-1}W_t(2B_t+A_t)
=2(W_T-W_0)-g_T+g_0\le2(M+2)W_T.
\end{equation}
Combining \eqref{eq:increasing-loss-bound} and \eqref{eq:increasing-sum-bound} proves the claim.
\end{proof}

\begin{lemma}[Logarithmic growth of the endpoint weight]\label{lem:increasing-growth}
Under the same assumptions, there is $C_{\mathrm{growth}}>0$, depending only on the initialization, such that
\[
W_T\le C_{\mathrm{growth}}\log(T+1),\qquad T\ge1.
\]
More explicitly, define
\[
V(\theta):=\frac{\|\theta\|^2}{W}-\frac{3W-g-2b}{4},\qquad
K_V:=(M+2)^2,\qquad \lambda:=\frac1{16K_V}.
\]
Then
\[
W_T\le\frac8\lambda\log\left(e^{\lambda V(\theta_0)}+\lambda e^\lambda T\right).
\]
\end{lemma}

\begin{proof}
We use the potential $V$, which is comparable to $W$ and whose one-step increase is exponentially small in $W_t$; exponentiating then turns this into logarithmic growth. Write $V_t:=V(\theta_t)$.

\proofstep{Step 1 ($V$ is comparable to $W$).} Completing the squares,
\[
V(\theta)=\frac{W^2+b^2+g^2}{W}-\frac{3W}{4}+\frac g4+\frac b2=\frac{11W}{64}+\frac{(b+W/4)^2+(g+W/8)^2}{W}.
\]
Since $W_t>0$, the squared terms are nonnegative, giving the lower bound
\[
V_t\ge\frac{11W_t}{64}\ge\frac{W_t}{8}.
\]
For the upper bound, use $\theta = (w, b, g)$ and expand $V_t$. Then apply $|b_t|<W_t$ and $|g_t|\le MW_t$ from \Cref{lem:increasing-loss}:
\begin{align*}
V_t
&=\frac{W_t}{4}+\frac{b_t^2}{W_t}+\frac{g_t^2}{W_t}+\frac{b_t}{2}+\frac{g_t}{4}\\
&\le\frac{W_t}{4}+W_t+M^2W_t+\frac{W_t}{2}+\frac{MW_t}{4}\\
&=\left(M^2+\frac M4+\frac74\right)W_t.
\end{align*}
Finally, $K_V=(M+2)^2$ and
\[
(M+2)^2-\left(M^2+\frac M4+\frac74\right)=\frac{15M+9}{4}>0.
\]
Combining these bounds gives
\begin{equation}\label{eq:increasing-coercivity}
\frac{W_t}{8}\le V_t
\le\left(M^2+\frac M4+\frac74\right)W_t\le K_VW_t.
\end{equation}

\proofstep{Step 2 (an exponentially small upper bound on the drift).}
We next bound the upward drift of $V_t$. Let $s_t=-2B_t\xe-A_tx_{\xi_t}\in\partial\Ltil(\vartheta_t)$ be the selected subgradient of the convex extension in the reflected coordinates $\vartheta_t=(W_t,b_t,g_t)$, under the convention $\mathrm{ReLU}'(0)=0$. For $b_t\ne0$, $s_t=\nabla\Ltil(\vartheta_t)$; at $b_t=0$, it is the subgradient selected by this convention. The reflected update is therefore
\[
\vartheta_{t+1}=\vartheta_t-\kappa W_ts_t.
\]
Set $k_t=1+4\mathbf{1}\{b_t>0\}$. The two-case norm calculation in \eqref{eq:symmetric-subgradient-norm}, from the proof of \Cref{thm:finite}, gives
\[
\|s_t\|^2\le12B_t^2+k_tA_t^2,
\]
since $A_t,B_t>0$ and the cross terms in both cases are nonpositive. The coordinate updates in \Cref{lem:gd-update-insideCone} also give
\[
(3W_{t+1}-g_{t+1}-2b_{t+1})-(3W_t-g_t-2b_t)=\kappa W_tk_tA_t.
\]
We compute the change in $V_t$ directly from the update, using $\|\vartheta_t\|=\|\theta_t\|$. First, expanding the squared norm gives
\begin{align*}
\|\vartheta_{t+1}\|^2-\|\vartheta_t\|^2
&=-2\kappa W_t\ip{\vartheta_t}{s_t}+\kappa^2W_t^2\|s_t\|^2\\
&=2\kappa W_t(2B_tu_t+A_tv_t)+\kappa^2W_t^2\|s_t\|^2,
\end{align*}
where $\ip{\vartheta_t}{s_t}=-2B_tu_t-A_tv_t$ follows from the definitions of $s_t$ and the margins. To account for the changing denominator, use $W_{t+1}-W_t=2\kappa W_tB_t$ from \eqref{eq:increasing-w-update}:
\begin{align*}
\frac{\|\vartheta_{t+1}\|^2}{W_{t+1}}-\frac{\|\vartheta_t\|^2}{W_t}
&=\frac{\|\vartheta_{t+1}\|^2-\|\vartheta_t\|^2}{W_t}
-\frac{W_{t+1}-W_t}{W_tW_{t+1}}\|\vartheta_{t+1}\|^2\\
&=2\kappa(2B_tu_t+A_tv_t)+\kappa^2W_t\|s_t\|^2
-\frac{2\kappa B_t\|\vartheta_{t+1}\|^2}{W_{t+1}}.
\end{align*}
The linear correction $-(3W_t-g_t-2b_t)/4$ changes by $-\kappa W_tk_tA_t/4$, as shown above. Combining this change with the preceding identity, dividing by $2\kappa$, and applying the norm bound yields
\begin{align*}
\frac{V_{t+1}-V_t}{2\kappa}
&\le2B_tu_t+A_tv_t-W_tB_t(1-4\kappa B_t)-\frac{W_tk_t}{8}A_t(1-4\kappa A_t)\\
&\le2B_t\Bigl(u_t-\frac{W_t}{2}(1-B_t)\Bigr)+A_t\Bigl(v_t-\frac{W_t}{8}(1-A_t)\Bigr),
\end{align*}
where the second line uses $4\kappa\le1$ and $k_t\ge1$. The negative term with $B_t$ comes from the growing denominator $W_t$ in $V$, and the one with $A_t$ from the linear correction in $V$; for $\kappa\le\frac14$ they absorb the quadratic update term, playing the role of the comparator shift in \citet{WuCOLT24}. The remaining drift can be positive only when a margin is large compared with $W_t$, in which case its residual is exponentially small, as quantified by \Cref{lem:logistic-drift-bound}.

Applying \Cref{lem:logistic-drift-bound} with $(z,h)=(u_t,W_t/2)$ and $(z,h)=(v_t,W_t/8)$ to the two terms above yields
\begin{equation}\label{eq:increasing-drift}
V_{t+1}-V_t\le\frac{2\kappa}{e}\left(2e^{-W_t/4}+e^{-W_t/16}\right)\le\frac{6\kappa}{e}\,e^{-W_t/16}\le e^{-W_t/16},
\end{equation}
since $6\kappa/e<1$ for $\kappa\le\frac14$.

\proofstep{Step 3 (logarithmic growth by exponentiation and telescoping).}
Thus the potential need not decrease at every step; its upward drift becomes exponentially small as the parameter scale grows. Exponentiation converts this drift bound into a uniform bound on increments.
Combining \eqref{eq:increasing-coercivity} and \eqref{eq:increasing-drift} gives $\Delta_t:=V_{t+1}-V_t\le e^{-\lambda V_t}\le1$. We consider the sign of this increment separately at each step.

If $\Delta_t>0$, the mean value theorem gives some $c_t\in(V_t,V_{t+1})$ such that
\[
e^{\lambda V_{t+1}}-e^{\lambda V_t}
=\lambda e^{\lambda c_t}\Delta_t
\le\lambda e^\lambda e^{\lambda V_t}\Delta_t,
\]
where the inequality uses $c_t\le V_{t+1}=V_t+\Delta_t\le V_t+1$.
If $\Delta_t\le0$, then $V_{t+1}\le V_t$, so monotonicity of the exponential and $\lambda>0$ give $e^{\lambda V_{t+1}}-e^{\lambda V_t}\le0$. Also $\relu{\Delta_t}=0$ in this case. Thus, in both cases,
\[
e^{\lambda V_{t+1}}-e^{\lambda V_t}
\le\lambda e^\lambda e^{\lambda V_t}\relu{\Delta_t}
\le\lambda e^\lambda.
\]
The last inequality uses $\relu{\Delta_t}\le e^{-\lambda V_t}$.
Summing over $t=0,\ldots,T-1$ telescopes to
\[
e^{\lambda V_T}-e^{\lambda V_0}
=\sum_{t=0}^{T-1}\left(e^{\lambda V_{t+1}}-e^{\lambda V_t}\right)
\le\lambda e^\lambda T.
\]
Rearranging, taking logarithms, and using $W_T\le8V_T$ gives
\[
W_T\le8V_T\le\frac8\lambda\log\left(e^{\lambda V_0}+\lambda e^\lambda T\right).
\]
To obtain the stated order, note that
\[
\log\left(e^{\lambda V_0}+\lambda e^\lambda T\right)
\le\log\left(e^{\lambda V_0}+\lambda e^\lambda\right)+\log(T+1).
\]
For $T\ge1$, $\log(T+1)\ge\log2$, so the fixed first term is also bounded by a constant multiple of $\log(T+1)$. Since $V_0$ and $\lambda$ depend only on $\theta_0$, this proves $W_T=O(\log(T+1))$ uniformly over $0<\kappa\le1/4$.
\end{proof}

\begin{proof}[Proof of \Cref{thm:increasing}]
Combining \Cref{lem:increasing-loss,lem:increasing-growth},
\[
\frac1T\sum_{t=0}^{T-1}\cL(\theta_t)
\le\frac{K_{\mathrm{loss}}W_T}{T}
\le\frac{K_{\mathrm{loss}}C_{\mathrm{growth}}\log(T+1)}{T}.
\]
With $C:=\max\{1,K_{\mathrm{loss}}\}\,C_{\mathrm{growth}}$, this gives both the loss bound and $W_T\le C\log(T+1)$.

\emph{The averaged iterate.} The same bound holds for $\theta_T^{\mathrm{av}}:=T^{-1}\sum_{t<T}\theta_t$. Since the sign of $w_t$ is fixed, the reflection of $\theta_T^{\mathrm{av}}$ is $\vartheta_T^{\mathrm{av}}:=T^{-1}\sum_{t<T}\vartheta_t$, and its first coordinate satisfies
\[
\frac1T\sum_{t=0}^{T-1}W_t>\frac1T\sum_{t=0}^{T-1}|b_t|\ge\Bigl|\frac1T\sum_{t=0}^{T-1}b_t\Bigr|,
\]
so $\theta_T^{\mathrm{av}}\in\cC$. Jensen's inequality for the convex function $\Ltil$ then gives
\[
\cL(\theta_T^{\mathrm{av}})=\Ltil(\vartheta_T^{\mathrm{av}})
\le\frac1T\sum_{t<T}\Ltil(\vartheta_t)
=\frac1T\sum_{t<T}\cL(\theta_t).
\qedhere
\]
\end{proof}

\subsection{Implicit Bias with Increasing Step Sizes}



\begin{lemma}[Bounded orthogonal component for increasing steps]\label{lem:adaptive-bounded}
Let $\theta_0\in\cC$, $0<\kappa\le1/4$, and $\eta_t=\kappa W_t$. Then $p_t\to\infty$ and $\sup_t\|\thb_t\|<\infty$.
\end{lemma}

\begin{proof}
Write $L_t:=\cL(\theta_t)$, $E_t:=e^{\gamma p_t}$ and $\alpha_t:=\eta_t/E_t$; as in Step~3 of the proof of \Cref{thm:implicit}, $\alpha_t$ is the effective step size of the orthogonal component. Compared with \Cref{lem:bounded}, the difficulty is that $\eta_t\to\infty$, so the second-order term of the $\thb$-update is not uniformly small. Instead, we show that $\alpha_t\to0$ (Step~1), introduce a smooth potential $F$ comparable to $G$ whose gradient is positively aligned with the update (Step~2), show that a sublevel set of $F$ is invariant once $\alpha_tF$ is small (Step~3), and find a time at which this happens (Step~4).

\proofstep{Step 1 ($p_t\to\infty$, $\alpha_t\to0$, and a time with small $\eta_tL_t$).}
By \Cref{lem:invariance,lem:adaptive-divergence}, the iterates remain in $\cC$ and $W_t\to\infty$. As in the proof of \Cref{lem:adaptive-divergence}, $g_t\le g_0+W_t-W_0$, hence
\[
\gamma p_t=\frac{2W_t-g_t}{5}\ge\frac{W_t+W_0-g_0}{5}\ \to\ \infty .
\]
In particular,
\[
0<\alpha_t=\kappa W_te^{-\gamma p_t}\le\kappa e^{(g_0-W_0)/5}\,W_te^{-W_t/5}\ \to\ 0 .
\]
Moreover, $\alpha_t$ is eventually non-increasing: by \eqref{eq:rot} and $W_{t+1}=W_t(1+2\kappa B_t)$, once $W_t\ge5$,
\[
\log\frac{\alpha_{t+1}}{\alpha_t}=\log(1+2\kappa B_t)-\frac{\kappa W_t(2B_t+A_t)}{5}\le2\kappa B_t\Bigl(1-\frac{W_t}{5}\Bigr)\le0 .
\]
Finally, by \Cref{lem:increasing-loss,lem:increasing-growth}, $W_T=O(\log(T+1))$ and $\sum_{t<2T}L_t=O(\log(T+1))$. Let $s_T\in\{T,\dots,2T-1\}$ be an index at which $L_t$ is smallest in that window. Then
\[
L_{s_T}\le\frac1T\sum_{t<2T}L_t=O\Bigl(\frac{\log(T+1)}{T}\Bigr),\qquad
\eta_{s_T}L_{s_T}=\kappa W_{s_T}L_{s_T}=O\Bigl(\frac{\log^2(T+1)}{T}\Bigr)\ \to\ 0 .
\]

\proofstep{Step 2 (a smooth potential comparable to $G$).}
For $\thb=(q,b)$, put
\[
X:=2e^{-3q/\sqrt5-b},\qquad Y:=e^{2q/\sqrt5},\qquad Z:=e^{2q/\sqrt5+2b},\qquad F(\thb):=X+Y+Z .
\]
Then $X^2Y^2Z=4$, $G_-=X+Y$, $G_+=X+Z$, and $G=X+\max\{Y,Z\}$, so $G\le F\le2G$. In particular, $F$ is smooth and, by \Cref{lem:G}, coercive. We claim that on the branch $\xi(b)$ selected by GD,
\begin{equation}\label{eq:adaptive-common-drift}
\ip{\nabla F(\thb)}{\nabla G_{\xi(b)}(\thb)}
\ge\frac{F(\thb)^2}{50}-32 .
\end{equation}
In the coordinates $(q,b)$, $\nabla X=X\,a$, $\nabla Y=Y\,c$ and $\nabla Z=Z\,d$, where
\[
a=\Bigl(-\frac3{\sqrt5},-1\Bigr),\qquad c=\Bigl(\frac2{\sqrt5},0\Bigr),\qquad d=\Bigl(\frac2{\sqrt5},2\Bigr),
\]
with
\[
\|a\|^2=\frac{14}5,\quad \|c\|^2=\frac45,\quad \|d\|^2=\frac{24}5,\quad \ip ac=-\frac65,\quad \ip ad=-\frac{16}5,\quad \ip cd=\frac45 .
\]
Expanding $\ip{\nabla F}{\nabla G_\pm}$ with these values gives the two quadratic forms below.

\emph{Case $b\le0$ (branch $-$, including $b=0$).} Here $Z\le Y$, and
\[
5\ip{\nabla F}{\nabla G_-}=14X^2+4Y^2-12XY-16XZ+4YZ .
\]
We use three estimates. (1) $14X^2+4Y^2-12XY\ge X^2+Y^2$, since the difference $13X^2-12XY+3Y^2$ has the positive definite matrix $\left(\begin{smallmatrix}13&-6\\-6&3\end{smallmatrix}\right)$. (2) $XZ\le X+2$: if $Z\le1$, then $XZ\le X$; if $Z>1$, then $X=2/(Y\sqrt Z)$ gives $XZ=2\sqrt Z/Y\le2/\sqrt Z<2$, using $Z\le Y$. (3) $YZ\ge0$. Hence
\[
\ip{\nabla F}{\nabla G_-}\ge\frac{X^2+Y^2-16X-32}{5}\ge\frac{X^2+Y^2}{10}-32\ge\frac{F^2}{50}-32,
\]
where the middle step is equivalent to $\frac{(X-16)^2+Y^2}{10}\ge0$, and the last uses $F\le X+2Y$ and $(X+2Y)^2\le5(X^2+Y^2)$.

\emph{Case $b>0$ (branch $+$).} Here $Y<Z$, and
\[
5\ip{\nabla F}{\nabla G_+}=14X^2+24Z^2-32XZ-6XY+4YZ .
\]
Similarly: (1) $14X^2+24Z^2-32XZ\ge X^2+Z^2$, since $13X^2-32XZ+23Z^2$ has the positive definite matrix $\left(\begin{smallmatrix}13&-16\\-16&23\end{smallmatrix}\right)$. (2) $XY\le X+2$: if $Z<1$, then $Y<Z<1$ and $XY\le X$; if $Z\ge1$, then $XY=2/\sqrt Z\le2$. (3) $YZ\ge0$. Hence
\[
\ip{\nabla F}{\nabla G_+}\ge\frac{X^2+Z^2-6X-12}{5}\ge\frac{X^2+Z^2}{10}-6\ge\frac{F^2}{50}-32,
\]
where the middle step is equivalent to $\frac{(X-6)^2+Z^2}{10}\ge0$, and the last uses $F\le X+2Z$ and $(X+2Z)^2\le5(X^2+Z^2)$. This proves \eqref{eq:adaptive-common-drift}. In particular,
\begin{equation}\label{eq:adaptive-drift-large}
F(\thb)\ge64\quad\Longrightarrow\quad\ip{\nabla F(\thb)}{\nabla G_{\xi(b)}(\thb)}\ge\frac{F(\thb)^2}{100}.
\end{equation}

\proofstep{Step 3 (one step of $F$).}
Suppress the time index, write $\thb'$ for the next orthogonal iterate, and let $C:=e^{-\ip{\thb}{h_{\xi(b)}}}$, i.e., $C=Y$ if $b\le0$ and $C=Z$ if $b>0$. By \eqref{eq:rot}, the exact orthogonal update is
\[
\thb'=\thb-\alpha s_E,\qquad s_E:=-2EB\,\he-EA\,h_{\xi(b)} .
\]
Since $e^{-u}=X/(2E)$ and $e^{-v}=C/E$, the coefficients in $s_E$ are
\[
2EB=\frac{X}{1+X/(2E)},\qquad EA=\frac{C}{1+C/E},
\]
which are slightly smaller than the corresponding coefficients in $\nabla G_{\xi(b)}=-X\he-C\,h_{\xi(b)}$. With $H=\sqrt{24/5}$, this gives
\[
\|s_E\|\le H(X+C)\le HF,\qquad
\|s_E-\nabla G_{\xi(b)}\|\le\frac HE\Bigl(\frac{X^2}{2}+C^2\Bigr)\le\frac{HF^2}{E}.
\]
Since $F$ is a sum of three exponentials of linear functions whose gradients have norm at most $H$, it also satisfies $\|\nabla F\|\le HF$, $\|\nabla^2F\|\le H^2F$, and $F(\thb+\delta)\le F(\thb)e^{H\|\delta\|}$ for every $\delta\in\R^2$. Taylor's theorem with these bounds gives
\begin{equation}\label{eq:adaptive-taylor}
F(\thb')\le F-\alpha\ip{\nabla F}{\nabla G_{\xi(b)}}+\alpha\,\frac{H^2F^3}{E}+\frac12H^4e^{H^2\alpha F}\alpha^2F^3 ,
\end{equation}
where $F$ and its gradient are evaluated at $\thb$: the third term bounds $\alpha|\ip{\nabla F}{s_E-\nabla G_{\xi(b)}}|$, and the last term bounds the second-order remainder.

Set $\delta_0:=1/(1000H^4)$ and suppose $\eta\ge1$ and $\alpha F\le\delta_0$. Then $F/E=\alpha F/\eta\le\delta_0$, and the last two terms of \eqref{eq:adaptive-taylor} are at most
\[
\alpha F^2\Bigl(H^2\delta_0+\tfrac12H^4e^{H^2\delta_0}\delta_0\Bigr)<\frac{\alpha F^2}{200}.
\]
Consequently:
\begin{itemize}
\item if $F(\thb)\ge64$, then \eqref{eq:adaptive-drift-large} gives $F(\thb')\le F-\frac{\alpha F^2}{100}+\frac{\alpha F^2}{200}\le F(\thb)$;
\item if $F(\thb)<64$, then directly $F(\thb')\le F(\thb)e^{H\alpha\|s_E\|}\le F(\thb)e^{H^2\alpha F}<64e^{H^2\delta_0}<128$.
\end{itemize}

\proofstep{Step 4 (entering the trapping set).}
Along the times $s_T$ of Step~1, eventually $L_{s_T}\le1$. Since $e^{-z}=e^{\ell(z)}-1$ and $e^\ell-1\le2\ell$ for $0\le\ell\le1$, and since $\ell(u),\ell(v)\le L\le1$,
\[
\frac{G(\thb_{s_T})}{E_{s_T}}=2e^{-u_{s_T}}+e^{-v_{s_T}}\le2L_{s_T},
\qquad\text{hence}\qquad
\alpha_{s_T}F(\thb_{s_T})\le2\alpha_{s_T}G(\thb_{s_T})\le4\eta_{s_T}L_{s_T}\to0 .
\]
Together with $\alpha_t\to0$ and $W_t\to\infty$, this yields a finite time $t_0$ (one of the $s_T$) with
\[
W_{t_0}\ge\max\{5,1/\kappa\},\qquad
\alpha_{t_0}K\le\delta_0,\qquad\text{where}\quad K:=\max\{F(\thb_{t_0}),128\}.
\]
For $t\ge t_0$, we have $\alpha_t\le\alpha_{t_0}$ (Step~1, since $W_t\ge5$) and $\eta_t=\kappa W_t\ge1$. We show by induction that $F(\thb_t)\le K$ for all $t\ge t_0$. If $F(\thb_t)\le K$, then $\alpha_tF(\thb_t)\le\alpha_{t_0}K\le\delta_0$, so Step~3 applies: either $F(\thb_t)\ge64$ and $F(\thb_{t+1})\le F(\thb_t)\le K$, or $F(\thb_t)<64$ and $F(\thb_{t+1})<128\le K$. Coercivity of $F$ now bounds $\thb_t$ for $t\ge t_0$, and the finitely many earlier iterates are trivially bounded. Hence $\sup_t\|\thb_t\|<\infty$.
\end{proof}

\begin{lemma}[A growth recursion]\label{lem:recursion}
Let $E_t\to\infty$ satisfy $E_{t+1}-E_t=c\log E_t\,(1+o(1))$ for some $c>0$. Then $E_t/(t\log t)\to c$.
\end{lemma}

\begin{proof}
We first obtain crude bounds on $E_t$, then sharpen them to $\log E_t=\log t\,(1+o(1))$, and conclude with the Stolz--Ces\`aro theorem.

\proofstep{Crude bounds.} Since $\log E_t\to\infty$, for large $t$ we have $E_{t+1}-E_t\ge1$, so $E_t\ge t/2$ eventually and $\log E_t\ge\log t-1$. Also, eventually $E_{t+1}-E_t\le2c\log E_t\le\sqrt{E_t}$, which gives $\sqrt{E_{t+1}}\le\sqrt{E_t+\sqrt{E_t}}\le\sqrt{E_t}+\frac12$. Hence $\sqrt{E_t}=O(t)$, i.e., $E_t=O(t^2)$ and $\log E_t\le2\log t+O(1)$.

\proofstep{Sharper bounds.} Summing $E_{t+1}-E_t\le2c\log E_t\le4c\log t+O(1)$ gives $E_t\le C\,t\log t$ for some constant $C$, hence $\log E_t\le\log t+\log\log t+O(1)$. Together with $\log E_t\ge\log t-1$, this gives $\log E_t=\log t\,(1+o(1))$, and therefore $E_{t+1}-E_t=c\log t\,(1+o(1))$.

\proofstep{Conclusion.} Let $D_t:=t\log t$, so $D_{t+1}-D_t=\log(t+1)+t\log(1+1/t)=\log t\,(1+o(1))$. By the Stolz--Ces\`aro theorem, $E_t/D_t\to\lim_t(E_{t+1}-E_t)/(D_{t+1}-D_t)=c$.
\end{proof}

\begin{proof}[Proof of \Cref{thm:adaptive}]
The proof follows that of \Cref{thm:implicit}, with the constant $\eta$ replaced by $\eta_t=\kappa W_t\asymp\log t$; this extra logarithmic factor in the step size produces the $\log\log t$ term in $R_t$. By \Cref{lem:invariance,lem:adaptive-bounded}, all iterates lie in $\cC$, $p_t\to\infty$, and $\|\thb_t\|\le M$ for some $M<\infty$. Let $E_t:=e^{\gamma p_t}$; since $p_t\to\infty$, \Cref{lem:expansions} applies for all large $t$.

\proofstep{Step 1 ($E_t\asymp t\log t$).}
Since $W=\frac{2p+q}{\sqrt5}=2\gamma p+\gamma q$ and $q_t$ is bounded,
\[
\eta_t=\kappa W_t=\kappa\bigl(2\log E_t+\gamma q_t\bigr)\asymp\log E_t .
\]
As in Step~2 of the proof of \Cref{thm:implicit}, \Cref{lem:expansions}(i)--(ii) give, for $\chi_t:=\gamma(p_{t+1}-p_t)$,
\[
\chi_t=\eta_t\gamma^2(2B_t+A_t)\in\Bigl[\frac{\eta_t\gamma^2}{2E_t},\ \frac{3e^{HM}\eta_t\gamma^2}{E_t}\Bigr].
\]
Thus $\chi_t\to0$ (because $\eta_t/E_t\to0$), and $E_{t+1}-E_t=E_t(e^{\chi_t}-1)\asymp E_t\chi_t\asymp\eta_t\asymp\log E_t$. The comparison argument in the proof of \Cref{lem:recursion}, run with these two-sided bounds, first gives $E_t\gtrsim t$ and $E_t=O(t^2)$, then $\log E_t=\Theta(\log t)$; summing the increments yields $E_t\asymp t\log t$.

\proofstep{Step 2 ($\thb_t\to\ths$).}
Exactly as in Step~3 of the proof of \Cref{thm:implicit}, now with $\alpha_t:=\eta_t/E_t$,
\[
\thb_{t+1}=\thb_t-\alpha_t(s_t-\varepsilon_t),\qquad s_t\in\partial G(\thb_t),\qquad
\|s_t\|\le3He^{HM},\qquad \|\varepsilon_t\|\le\frac{3He^{2HM}}{E_t}.
\]
Since $\eta_t\asymp\log t$ and $E_t\asymp t\log t$, we have $\alpha_t\asymp1/t$ and $\alpha_t\|\varepsilon_t\|=O\bigl(1/(t^2\log t)\bigr)$. Hence $\sum_t\alpha_t=\infty$, $\sum_t\alpha_t^2<\infty$ and $\sum_t\alpha_t\|\varepsilon_t\|<\infty$, and \Cref{lem:G,lem:subgrad} give $\thb_t\to\ths$.

\proofstep{Step 3 (exact growth of $E_t$).}
As in Step~4 of the proof of \Cref{thm:implicit}, $E_t(2B_t+A_t)\to\Gs$. Moreover, $\eta_t=2\kappa\log E_t\,(1+o(1))$ because $q_t$ is bounded. Therefore
\[
E_{t+1}-E_t=\eta_t\gamma^2E_t(2B_t+A_t)\bigl(1+o(1)\bigr)=2\kappa\gamma^2\Gs\log E_t\,\bigl(1+o(1)\bigr),
\]
and \Cref{lem:recursion} with $c=2\kappa\gamma^2\Gs=2\kappa\,3^{-3/5}$ gives
\begin{equation}\label{eq:adaptive-E}
\frac{E_t}{t\log t}\to2\kappa\,3^{-3/5},\qquad\text{i.e.,}\qquad \gamma p_t=R_t-\frac35\log3+o(1),
\end{equation}
recalling $R_t=\log(2\kappa t\log t)$.

\proofstep{Part (a).}
We use $W=2\gamma p+\gamma q$, $g=-\gamma p+2\gamma q$, $\gamma q_t\to\gamma q^\star=\frac15\log3$, $b_t\to0$, and \eqref{eq:adaptive-E}:
\[
W_t=2R_t-\tfrac65\log3+\tfrac15\log3+o(1)=2R_t-\log3+o(1),\qquad
g_t=-R_t+\tfrac35\log3+\tfrac25\log3+o(1)=-R_t+\log3+o(1).
\]
This is the formula in (a). Since $\vartheta_t=(|w_t|,b_t,g_t)$, $\|\vartheta_t\|=\|\theta_t\|$, and $R_t\to\infty$, the normalized direction converges to $(2,0,-1)/\sqrt5$.

\proofstep{Part (b).}
By \Cref{lem:expansions}(iii), $E_t\cL(\theta_t)\to\Gs$, so by \eqref{eq:adaptive-E}
\[
t\log t\;\cL(\theta_t)=\frac{t\log t}{E_t}\cdot E_t\cL(\theta_t)\ \to\ \frac{\Gs}{2\kappa\,3^{-3/5}}=\frac5{2\kappa}.
\]
For reference, part (a) also gives the margins $u_t=W_t+b_t+g_t=R_t+o(1)$ and $v_t=-2\relu{b_t}-g_t=R_t-\log3+o(1)$, and the step size $\eta_t=2\kappa R_t-\kappa\log3+o(1)$.

\proofstep{Part (c).}
By \eqref{eq:f-linearization}, $\sup_{|x|\le1}\bigl|f_{\theta_t}(x)-(W_t|x|+g_t)\bigr|\le2|b_t|\to0$. Substituting part (a),
\[
f_{\theta_t}(x)=(2R_t-\log3)|x|-R_t+\log3+o(1)=(2|x|-1)R_t+(1-|x|)\log3+o(1)
\]
uniformly over $x\in[-1,1]$.
\end{proof}





\section{UNSYMMETRIC HIDDEN WEIGHTS: CONVERGENCE AND FAILURE}\label{app:unsymmetric}

\subsection{NTK Separability and Necessary Alignment}\label{app:ntk}

For a model $g_\theta$, its tangent feature at initialization is $\Phi_0(x)=\nabla_\theta g_{\theta_0}(x)$, with $\mathrm{ReLU}'(0)=0$, and its empirical NTK is
\[
K_0(x,x')=\ip{\Phi_0(x)}{\Phi_0(x')}.
\]
We call the data \emph{NTK separable at initialization} if there exist a unit vector $v$ and $\gamma>0$ such that
\begin{equation}\label{eq:ntk-separability}
\min_{(x,y)\in\cS}y\,\ip{v}{\Phi_0(x)}\ge\gamma.
\end{equation}
This condition asks for a common positive margin on the \emph{signed} features $y\Phi_0(x)$. For the unsymmetric model $g_\theta=f^a_\theta$, we have $g_{\theta_0}(x)=\ip{\theta_0}{\Phi_0(x)}$, including at zero preactivations under our convention. Thus its frozen tangent model is
\[
g^{\mathrm{lin}}_{\theta_0,\theta}(x)
:=g_{\theta_0}(x)+\ip{\theta-\theta_0}{\Phi_0(x)}
=\ip{\theta}{\Phi_0(x)}.
\]

\begin{lemma}[Same-sign initialization and necessary alignment]\label{lem:unsymmetric-ntk}
For $f^a$, let $\theta_0=(w_{1,0},w_{2,0},b_0,g_0)$ satisfy $w_{1,0}w_{2,0}>0$ and $b_0\le0$. Then the data are not NTK separable at initialization, and the frozen tangent model has logistic loss at least $\log4$ for every parameter vector. Moreover, every parameter vector giving strictly positive margins on all three examples must, after possibly exchanging the neurons, satisfy
\[
w_1>|b|,\qquad -w_2>|b|.
\]
Consequently, any GD trajectory from such an initialization whose loss tends to zero must change its activation patterns and at least one weight's sign, and eventually remain in one of these two cones.
\end{lemma}

\begin{proof}
\proofstep{Non-separability.} Let $s:=\operatorname{sgn}(w_{1,0})=\operatorname{sgn}(w_{2,0})$. At $x=-s$, both preactivations $w_{i,0}x+b_0=-|w_{i,0}|+b_0$ are strictly negative; at $x=0$, both equal $b_0\le0$. With $\mathrm{ReLU}'(0)=0$, only the output bias contributes to the gradient at these two inputs:
\[
\Phi_0(-s)=\Phi_0(0)=(0,0,0,1).
\]
Their labels are $+1$ and $-1$, so their signed tangent features are negatives of each other, and no direction has positive inner product with both. Likewise, the frozen tangent model predicts the same value $z$ at both points, and their contribution to the loss alone is
\[
\ell(z)+\ell(-z)=\log\bigl(2+e^z+e^{-z}\bigr)\ge\log4 .
\]

\proofstep{Necessary alignment.} Strictly positive margins on all three examples mean
\[
f^a(-1)>0,\qquad f^a(0)<0,\qquad f^a(1)>0,\qquad\text{and hence}\qquad f^a(\pm1)>f^a(0).
\]
\begin{itemize}
\item \emph{Opposite signs are necessary.} If $w_1w_2\ge0$, both terms $\relu{w_ix+b}$ are monotone in $x$ in the same direction (also when one weight is zero), so $f^a$ is monotone, and $f^a(-1)>f^a(0)<f^a(1)$ is impossible. After possibly exchanging the neurons, write $w_1=r_1>0$ and $w_2=-r_2<0$.
\item \emph{The case $b\le0$.} Then $f^a(0)=g$ and
\[
f^a(1)-f^a(0)=\relu{r_1+b},\qquad f^a(-1)-f^a(0)=\relu{r_2+b}.
\]
Both are positive if and only if $r_1>-b=|b|$ and $r_2>|b|$.
\item \emph{The case $b>0$.} Suppose $r_1\le b$. Then
\[
f^a(1)-f^a(0)=r_1-b+\relu{b-r_2},
\]
and positivity forces $\relu{b-r_2}>b-r_1\ge0$, i.e., $r_2<r_1\le b$. But then $f^a(-1)-f^a(0)=(b-r_1)+(r_2+b)-2b=r_2-r_1<0$, a contradiction. Hence $r_1>b$, and the symmetric argument gives $r_2>b$.
\end{itemize}
In all cases $w_1>|b|$ and $-w_2>|b|$.

\proofstep{Consequences for GD.} In the cone $\{w_1>|b|,\,-w_2>|b|\}$, the unit direction $(2,-2,0,-1)/3$ has inner product $\frac13$ with every signed tangent feature (see the features in \Cref{app:unsymmetric-invariant}), so the data are NTK separable there. If the loss tends to zero, all three margins are strictly positive from some time on, so the iterates eventually lie in one of the two cones. Since the initial weights have the same sign, at least one weight must change sign, and the initially inactive endpoint $x=-s$ must acquire an active neuron. Finally, inside such a cone the update of \Cref{lem:unsymmetric-gd} moves the positive weight up and the negative weight down, so once all subsequent iterates classify the data perfectly, their orientation cannot switch.
\end{proof}

The assumption $b_0\le0$ is essential to the nonseparability claim. For example, if $w_{1,0},w_{2,0}>b_0>0$, then
\[
\Phi_0(-1)=(0,0,0,1),\quad
\Phi_0(0)=(0,0,2,1),\quad
\Phi_0(1)=(1,1,2,1).
\]
The unit vector $(2,2,-1,1)/\sqrt{10}$ separates the signed features with margin $1/\sqrt{10}$. Thus same-sign weights alone do not imply failure of NTK separability, even though opposite orientations are necessary for the nonlinear network to classify all three points.

\subsection{Finite-Time Exit under Gradient Flow}\label{app:unsymmetric-gf-exit}

\begin{lemma}[Gradient flow can leave the active region]\label{lem:unsymmetric-gf-exit}
Consider gradient flow for $\cL^a$ from $(w_{1,0},w_{2,0},b_0,g_0)=(r_0,-r_0,b_0,g_0)$, where $r_0>b_0>0$. Set $\delta=r_0-b_0$ and assume
\[
r_0+b_0+g_0+10\delta\le-\log5.
\]
Then there exists a time $\tau$ such that $b(t)>|w_1(t)|=|w_2(t)|$ for all $t>\tau$.
\end{lemma}

\begin{proof}[Proof of \Cref{lem:unsymmetric-gf-exit}]
By the symmetry of the initialization and of the dataset, $w_1(t)=-w_2(t)=:r(t)$ for all $t$. We show that $b$ catches up with $r$ within time $2\delta$ (Phase~1), and that the trajectory is then trapped in the linear region $\{b>r\}$ (Phase~2).

\proofstep{Phase 1 (inside the cone).} While $0<b<r$, write
\[
u:=r+b+g,\qquad z:=2b+g,\qquad B:=\sigma(-u),\qquad A:=\sigma(z)
\]
for the endpoint margin, the middle output, and the corresponding residuals. The gradient-flow equations (cf.\ \Cref{lem:unsymmetric-gd}) are
\[
\dot r=B,\qquad \dot b=2B-2A,\qquad \dot g=2B-A,
\]
and consequently, for the gap $d:=r-b$,
\[
\dot u=5B-3A\le5,\qquad \dot d=2A-B .
\]
As long as $0<b<r$ and $t\le2\delta$, the initialization condition gives
\[
u(t)\le u(0)+5t\le r_0+b_0+g_0+10\delta\le-\log5 .
\]
Since $z=u-d<u$, this yields $B\ge\sigma(\log5)=\frac56$ and $A\le\sigma(-\log5)=\frac16$, hence
\[
\dot b\ge\tfrac43>0,\qquad \dot d\le-\tfrac12 .
\]
Thus $b$ increases (so it never reaches $0$), and $d(t)\le\delta-t/2$, so the gap closes at some time $\tau\le2\delta$, with $b(\tau)=r(\tau)$ and $z(\tau)=u(\tau)\le-\log5$.

\proofstep{Phase 2 (the linear region).} If $b>r>0$, both neurons are active on all three inputs, so $f^a(x)=z$ on the training set and $\cL^a=2\ell(z)+\ell(-z)$. The endpoint contributions to each weight gradient cancel, and
\[
\dot r=0,\qquad \dot b=2\bigl(2-3\sigma(z)\bigr),\qquad \dot g=2-3\sigma(z),\qquad \dot z=5\bigl(2-3\sigma(z)\bigr).
\]
At time $\tau$, the one-sided derivatives of $b-r$ are positive on both sides of the boundary: $\dot b-\dot r=B-2A\ge\frac12$ on the cone side, and $2(2-3\sigma(z(\tau)))\ge3$ on the linear side. Hence the trajectory crosses into $\{b>r\}$. There, $z$ increases monotonically from $z(\tau)<\log2$ towards the equilibrium $\log2$, so $\dot b>0$ while $\dot r=0$: the trajectory never returns to the active region, and
\[
\cL^a(\theta(t))=2\ell(z(t))+\ell(-z(t))\ \longrightarrow\ \log\tfrac{27}{4}.
\]
The gradient-flow equation holds almost everywhere; its value at the isolated switching time does not affect the argument.
\end{proof}

\subsection{An Invariant Region}\label{app:unsymmetric-invariant}

\paragraph{Setup.}
We train
\[
f^a_{w_1,w_2,b,g}(x)=\relu{w_1x+b}+\relu{w_2x+b}+g
\]
on the dataset $\cS$ with the logistic loss, by GD with a constant step size $\eta>0$ and the convention $\mathrm{ReLU}'(0)=0$. We assume $w_{1,0}>0>w_{2,0}$, write $\theta=(w_1,w_2,b,g)\in\R^4$ and $a_1:=w_1$, $a_2:=-w_2$, and work in the active cone
\[
\cCu=\{\theta:\ a_1>|b|,\ a_2>|b|\},
\]
on which the endpoint $x=1$ activates only the first neuron and $x=-1$ only the second. As discussed in \Cref{sec:unsymmetric} (see \Cref{lem:unsymmetric-gf-exit} for the gradient-flow version), $\cCu$ is not invariant under GD for general initializations, even for arbitrarily small step sizes. The goal of this subsection is a smaller region, around the max-margin ray, that GD never leaves.

\paragraph{Features.}
Let $\xi(b)=+$ if $b>0$ and $\xi(b)=-$ otherwise, and define the four signed features
\[
x_1=(1,0,1,1),\qquad x_2=(0,-1,1,1),\qquad \xm=(0,0,0,-1),\qquad \xp=(0,0,-2,-1).
\]
For $\theta\in\cCu$ let $B_i=\sigma(-\ip{\theta}{x_i})$ for $i=1,2$ and $A=\sigma(-\ip{\theta}{x_{\xi(b)}})$.

\begin{lemma}[GD on the unsymmetric cone]\label{lem:unsymmetric-gd}
For $\theta\in\cCu$, $\cL^a(\theta)=\ell(\ip{\theta}{x_1})+\ell(\ip{\theta}{x_2})+\ell(\ip{\theta}{x_{\xi(b)}})$, and one GD step maps $\theta$ to
\begin{equation}\label{eq:unsymmetric-gd}
\theta'=\theta+\eta\big(B_1x_1+B_2x_2+A\,x_{\xi(b)}\big).
\end{equation}
Moreover, $\cL^a=\Ltil^a$ on $\cCu$, where $\Ltil^a(\theta)=\ell(a_1+b+g)+\ell(a_2+b+g)+\log(1+e^{2\relu{b}+g})$ is convex on $\R^4$ and $-(B_1x_1+B_2x_2+Ax_{\xi(b)})\in\partial\Ltil^a(\theta)$.
\end{lemma}

\begin{proof}
\proofstep{Outputs on the data.} Let $\theta\in\cCu$. At $x=1$ the preactivations are $a_1+b>0$ and $-a_2+b<0$; at $x=-1$ they are $-a_1+b<0$ and $a_2+b>0$; at $x=0$ both equal $b$. Hence
\[
f^a(1)=\ip{\theta}{x_1},\qquad f^a(-1)=\ip{\theta}{x_2},\qquad -f^a(0)=-2\relu{b}-g=\ip{\theta}{x_{\xi(b)}},
\]
which gives the formula for $\cL^a$.

\proofstep{Gradient.} The derivative of the loss term at $(x,y)$ with respect to the network output is $-y\,\sigma(-yf^a(x))$, i.e., $-B_1$ at $x=1$, $-B_2$ at $x=-1$, and $+A$ at $x=0$. Neuron~$1$ is active only at $x=1$ (input $+1$), neuron~$2$ only at $x=-1$ (input $-1$), and at $x=0$ both neurons are active if and only if $b>0$. Hence
\[
\partial_{w_1}\cL^a=-B_1,\qquad \partial_{w_2}\cL^a=B_2,\qquad \partial_b\cL^a=-B_1-B_2+2A\,\mathbf 1\{b>0\},\qquad \partial_g\cL^a=-B_1-B_2+A,
\]
that is, $-\nabla\cL^a(\theta)=B_1x_1+B_2x_2+A\,x_{\xi(b)}$, which gives \eqref{eq:unsymmetric-gd}. In particular, $w_1=a_1$ increases by $\eta B_1$ and $w_2=-a_2$ decreases by $\eta B_2$. Convexity of $\Ltil^a$ and the subgradient claim follow exactly as in the proof of \Cref{lem:piecewise}.
\end{proof}

\paragraph{Rotated coordinates.}
The vector $\thu=(2,-2,0,-1)=2x_1+2x_2+3\xm+2\xp$ satisfies $\ip{\thu}{x}=1$ for all four features and $\|\thu\|=3$. It is the max-margin vector of the four features, expressed in the original parameters, and represents the same classifier as in the symmetric model. Let
\[
e_0=\tfrac13\thu,\qquad f_1=\tfrac{1}{\sqrt2}(1,1,0,0),\qquad f_2=(0,0,1,0),\qquad f_3=\tfrac{1}{\sqrt{18}}(1,-1,0,4),
\]
an orthonormal basis of $\R^4$, and write $p=\ip{\theta}{e_0}$ and $\thbu=(\ip{\theta}{f_1},\ip{\theta}{f_2},\ip{\theta}{f_3})\in\R^3$. In the original parameters,
\[
p=\frac{2w_1-2w_2-g}{3},\qquad \thbu=\Big(\frac{w_1+w_2}{\sqrt2},\;b,\;\frac{w_1-w_2+4g}{\sqrt{18}}\Big).
\]
Every feature has $\ip{x_s}{e_0}=\frac13$, so $\ip{\theta}{x_s}=\frac p3+\ip{\thbu}{h_s}$ with orthogonal parts
\[
h_1=\Big(\tfrac{1}{\sqrt2},1,\tfrac{5}{\sqrt{18}}\Big),\quad h_2=\Big(-\tfrac{1}{\sqrt2},1,\tfrac{5}{\sqrt{18}}\Big),\quad h_-=\Big(0,0,-\tfrac{4}{\sqrt{18}}\Big),\quad h_+=\Big(0,-2,-\tfrac{4}{\sqrt{18}}\Big),
\]
which satisfy $2h_1+2h_2+3h_-+2h_+=0$ (the orthogonal part of $\thu=2x_1+2x_2+3\xm+2\xp$). By \eqref{eq:unsymmetric-gd}, a GD step from $\theta\in\cCu$ gives
\begin{equation}\label{eq:unsymmetric-rot}
p'=p+\tfrac{\eta}{3}\,(B_1+B_2+A),\qquad \thbu'=\thbu+\eta\big(B_1h_1+B_2h_2+A\,h_{\xi(b)}\big).
\end{equation}
In particular, $p$ is non-decreasing as long as the iterates are in $\cCu$. For $M,P>0$, let
\[
\mathcal R(M,P):=\{\theta\in\R^4:\ \|\thbu\|\le M,\ p\ge P\},
\]
a cylinder of radius $M$ around the max-margin ray.

\begin{theorem}[Invariant region for unsymmetric weights]\label{thm:unsymmetric}
Let $\eta>0$, $\rho(\eta)=\frac{3\sqrt5}{2}\big(\frac4e+\frac{43}{9}\eta\big)$, $M\ge\rho(\eta)+\eta\sqrt{86}/3$ and $P>\frac{\sqrt{14}}{2}M$. If $\theta_0\in\mathcal R(M,P)$, then $\theta_t\in\mathcal R(M,P)\subset\cCu$ for all $t\ge0$. In particular, the single region $\mathcal R(24.1,45.1)$ is invariant for every $\eta\in(0,1]$.
\end{theorem}

\begin{corollary}[Loss convergence]\label{cor:unsymmetric}
If $\eta\in(0,1]$ and $\theta_0\in\mathcal R(24.1,45.1)$, then for all $T\ge1$,
\[
\frac1T\sum_{t<T}\cL^a(\theta_t)\le\frac{3}{T+1}+\frac{\big(\|\theta_0\|+3\,(\log(T+1)+3\eta)\big)^2}{2\eta T}.
\]
\end{corollary}

\Cref{thm:local-convergence-unsymmetric} in \Cref{sec:unsymmetric} is \Cref{cor:unsymmetric} with $(M,P)=(24.1,45.1)$: its two conditions on $\theta_0$ state exactly that $p_0=\ip{\theta_0}{\thu}/\|\thu\|\ge P$ and that the component of $\theta_0$ orthogonal to $\thu$, whose norm is $\|\thbu_0\|$, is at most $M$. The proofs of \Cref{thm:unsymmetric,cor:unsymmetric} follow the symmetric case: we show that the most negative orthogonal feature is always active (\Cref{lem:unsymmetric-active}), which prevents $\|\thbu\|$ from growing once it is large (\Cref{lem:unsymmetric-step}).

\begin{lemma}[Non-separability; the minimal feature is active]\label{lem:unsymmetric-active}
Let $m=\frac{2}{3\sqrt5}$.
\begin{enumerate}[label=\textup{(\alph*)},leftmargin=*,itemsep=0pt,topsep=2pt]
\item For every $\thbu\in\R^3$, $\min_{s\in\{1,2,-,+\}}\ip{\thbu}{h_s}\le-m\|\thbu\|$.
\item For $\thbu=(\bar\theta_1,b,\bar\theta_3)$, the minimum in (a) is attained at an active feature, i.e., $\min_s\ip{\thbu}{h_s}=\min\{\ip{\thbu}{h_1},\ip{\thbu}{h_2},\ip{\thbu}{h_{\xi(b)}}\}$.
\end{enumerate}
\end{lemma}

\begin{proof}
\proofstep{(a)} The function $\mu(u):=\max_s\bigl(-\ip{u}{h_s}\bigr)$ is the support function of the tetrahedron $Q:=\mathrm{conv}\{-h_1,-h_2,-h_-,-h_+\}$. Since $2h_1+2h_2+3h_-+2h_+=0$ with positive coefficients and the $h_s$ span $\R^3$, the origin lies in the interior of $Q$. Hence $\min_{\|u\|=1}\mu(u)$ equals the distance from the origin to the boundary of $Q$, i.e., to the nearest facet plane of $Q$, or equivalently of $-Q=\mathrm{conv}\{h_s\}$. The four facet planes of $\mathrm{conv}\{h_s\}$ and their distances to the origin are
\begin{center}
\renewcommand{\arraystretch}{1.3}
\begin{tabular}{lll}
\toprule
facet opposite & plane & distance\\
\midrule
$h_1$ \ (resp.\ $h_2$) & $\bigl\{z:\ \ip{\tfrac{1}{\sqrt{10}}(\pm3,0,1)}{z}=-\tfrac{2}{3\sqrt5}\bigr\}$ & $\tfrac{2}{3\sqrt5}\approx0.298$\\
$h_-$ & $\bigl\{z:\ \ip{\tfrac{1}{\sqrt3}(0,-1,\sqrt2)}{z}=\tfrac{2}{3\sqrt3}\bigr\}$ & $\tfrac{2}{3\sqrt3}\approx0.385$\\
$h_+$ & $\bigl\{z:\ \ip{\tfrac{1}{\sqrt{11}}(0,-3,\sqrt2)}{z}=-\tfrac{4}{3\sqrt{11}}\bigr\}$ & $\tfrac{4}{3\sqrt{11}}\approx0.402$\\
\bottomrule
\end{tabular}
\end{center}
as one checks by evaluating the inner products at the three vertices of each facet. The smallest distance is $\frac{2}{3\sqrt5}=m$, and homogeneity gives the claim.

\proofstep{(b)} Since $h_+=h_-+(0,-2,0)$, we have $\ip{\thbu}{h_+}=\ip{\thbu}{h_-}-2b$, so $h_+$ has the smaller inner product if and only if $b>0$, i.e., if and only if $\xp$ is the active middle feature.
\end{proof}

\begin{lemma}[One step of the orthogonal component]\label{lem:unsymmetric-step}
Let $\theta\in\cCu$ with $p\ge0$, let $\theta'$ be its GD image, and let $\rho(\eta)=\big(\frac4e+\frac{43}{9}\eta\big)/m$. Then
(i) $\|\thbu'-\thbu\|\le\eta\sqrt{86}/3$, and (ii) if $\|\thbu\|\ge\rho(\eta)$, then $\|\thbu'\|\le\|\thbu\|$.
\end{lemma}

\begin{proof}
The argument parallels \Cref{lem:bounded}, now with three active features of unit multiplicity. Write $\omega_1:=B_1$, $\omega_2:=B_2$, $\omega_3:=A$ for the residuals of the active features $h_1,h_2,h_3:=h_{\xi(b)}$, and $z_i:=\ip{\thbu}{h_i}$. Then $\omega_i=\sigma(-\frac p3-z_i)\le1$, and \eqref{eq:unsymmetric-rot} reads $\thbu'=\thbu+\eta\sum_i\omega_ih_i$. Let $j$ be an active index with $z_j=\min_iz_i$; by \Cref{lem:unsymmetric-active}, $z_j\le-m\|\thbu\|\le0$.

\proofstep{Step 1 ($\omega_i\le\omega_j$ for all $i$).} This holds because $\sigma$ is increasing and $z_j\le z_i$.

\proofstep{Step 2 ($\|\sum_i\omega_ih_i\|^2\le\frac{86}{9}\,\omega_j^2$).} The left-hand side is a convex quadratic function of $\omega=(\omega_1,\omega_2,\omega_3)$, which by Step~1 lies in the box $[0,\omega_j]^3$. A convex function on a box is maximized at a vertex, i.e., at $\omega=\omega_j\mathbf 1_I$ for some $I\subseteq\{1,2,3\}$, where it equals $\omega_j^2\|\sum_{i\in I}h_i\|^2$. The values of $\|\sum_{i\in I}h_i\|^2$ are
\begin{center}
\renewcommand{\arraystretch}{1.3}
\begin{tabular}{lccccccc}
\toprule
$I$ & $\{1\}$ & $\{2\}$ & $\{3\}$ & $\{1,2\}$ & $\{1,3\}$ & $\{2,3\}$ & $\{1,2,3\}$\\
\midrule
$\xi=+$ & $\frac{26}9$ & $\frac{26}9$ & $\frac{44}9$ & $\frac{86}9$ & $\frac{14}9$ & $\frac{14}9$ & $2$\\
$\xi=-$ & $\frac{26}9$ & $\frac{26}9$ & $\frac{8}9$ & $\frac{86}9$ & $\frac{14}9$ & $\frac{14}9$ & $6$\\
\bottomrule
\end{tabular}
\end{center}
and the maximum is $\|h_1+h_2\|^2=\frac{86}9$ in both cases. Since $\omega_j\le1$, this proves (i): $\|\thbu'-\thbu\|=\eta\|\sum_i\omega_ih_i\|\le\eta\sqrt{86}/3$.

\proofstep{Step 3 ($\omega_iz_i\le\frac2e\,\omega_j$ for $i\ne j$).} This is trivial if $z_i\le0$. If $z_i>0$, then $\omega_iz_i\le z_ie^{-p/3-z_i}\le e^{-1}e^{-p/3}$, while $\omega_j\ge\sigma(-p/3)\ge\frac12e^{-p/3}$ because $z_j\le0\le p$.

\proofstep{Step 4 (conclusion).} Expanding the squared norm, and using $z_j\le-m\|\thbu\|$, Step~3 for the two indices $i\ne j$, Step~2, and $\omega_j\le1$,
\[
\|\thbu'\|^2-\|\thbu\|^2=2\eta\sum_i\omega_iz_i+\eta^2\Big\|\sum_i\omega_ih_i\Big\|^2\le2\eta\,\omega_j\Big(-m\|\thbu\|+\frac4e+\frac{43}{9}\eta\Big)=2\eta\,\omega_j\,m\bigl(\rho(\eta)-\|\thbu\|\bigr),
\]
which is nonpositive when $\|\thbu\|\ge\rho(\eta)$. This proves (ii).
\end{proof}

\begin{proof}[Proof of \Cref{thm:unsymmetric}]
Note that $1/m=\frac{3\sqrt5}{2}$, so $\rho(\eta)$ in the theorem coincides with that of \Cref{lem:unsymmetric-step}. Set $M(\eta):=\rho(\eta)+\eta\sqrt{86}/3$.

\proofstep{Step 1 ($\mathcal R(M,P)\subset\cCu$).} Expanding the coordinate vectors of $a_1=w_1$, $a_2=-w_2$ and $b$ in the basis $(e_0,f_1,f_2,f_3)$,
\[
(1,0,0,0)=\tfrac23e_0+\tfrac{1}{\sqrt2}f_1+\tfrac{1}{\sqrt{18}}f_3,\qquad
(0,-1,0,0)=\tfrac23e_0-\tfrac{1}{\sqrt2}f_1+\tfrac{1}{\sqrt{18}}f_3,\qquad
(0,0,1,0)=f_2 .
\]
Hence, with $\pi_{1,2}:=\bigl(\pm\frac1{\sqrt2},0,\frac1{\sqrt{18}}\bigr)$ and $e:=(0,1,0)$,
\[
a_i\pm b=\tfrac23p+\ip{\thbu}{\pi_i\pm e},\qquad \|\pi_i\pm e\|^2=\|\pi_i\|^2+1=\tfrac59+1=\tfrac{14}9 .
\]
By Cauchy--Schwarz, for $\theta\in\mathcal R(M,P)$,
\[
a_i\pm b\ \ge\ \tfrac23p-\tfrac{\sqrt{14}}{3}\|\thbu\|\ \ge\ \tfrac23P-\tfrac{\sqrt{14}}{3}M\ >\ 0,
\]
using $P>\frac{\sqrt{14}}2M$. Thus $a_i>|b|$ for $i=1,2$, i.e., $\theta\in\cCu$.

\proofstep{Step 2 (invariance).} Let $\theta_t\in\mathcal R(M,P)$. By Step~1, $\theta_t\in\cCu$, so \eqref{eq:unsymmetric-rot} gives $p_{t+1}\ge p_t\ge P>0$, and \Cref{lem:unsymmetric-step} applies:
\begin{itemize}
\item if $\|\thbu_t\|\ge\rho(\eta)$, then $\|\thbu_{t+1}\|\le\|\thbu_t\|\le M$;
\item otherwise, $\|\thbu_{t+1}\|<\rho(\eta)+\eta\sqrt{86}/3=M(\eta)\le M$.
\end{itemize}
Hence $\theta_{t+1}\in\mathcal R(M,P)$, and the claim follows by induction.

\proofstep{Step 3 (one region for all $\eta\le1$).} Numerically,
\[
M(1)=\frac{3\sqrt5}{2}\Bigl(\frac4e+\frac{43}{9}\Bigr)+\frac{\sqrt{86}}{3}=24.052\ldots\le24.1,\qquad \frac{\sqrt{14}}{2}\cdot24.1=45.087\ldots<45.1 .
\]
Since $M(\eta)$ is increasing in $\eta$, the hypotheses hold with $(M,P)=(24.1,45.1)$ for every $\eta\in(0,1]$.
\end{proof}

\begin{proof}[Proof of \Cref{cor:unsymmetric}]
By \Cref{thm:unsymmetric}, $\theta_t\in\cCu$ for all $t$, so by \Cref{lem:unsymmetric-gd}, GD is a subgradient method on the convex function $\Ltil^a$, which agrees with $\cL^a$ along the trajectory. We verify the two ingredients of the proof of \Cref{thm:finite}, with $\thu$ in place of $\thh$.

\proofstep{Self-bounding inequality.} Let $s_t:=-(B_{1,t}x_1+B_{2,t}x_2+A_tx_{\xi_t})$. Since every feature has margin one under $\thu$, $\ip{s_t}{\thu}=-(B_{1,t}+B_{2,t}+A_t)$. The Gram entries of the features are
\[
\|x_1\|^2=\|x_2\|^2=3,\quad \ip{x_1}{x_2}=2,\quad \ip{x_i}{\xm}=-1,\quad \ip{x_i}{\xp}=-3,\quad \|\xm\|^2=1,\quad \|\xp\|^2=5 .
\]
Dropping the nonpositive cross terms and using $B_{1,t}+B_{2,t}\le2$ and $A_t\le1$,
\[
\|s_t\|^2\le3B_{1,t}^2+3B_{2,t}^2+4B_{1,t}B_{2,t}+5A_t^2\le3(B_{1,t}+B_{2,t})^2+5A_t^2\le6(B_{1,t}+B_{2,t}+A_t)=-6\ip{s_t}{\thu}.
\]

\proofstep{Comparator.} For $R>0$, the comparator $R\thu=(2R,-2R,0,-R)$ lies in $\cCu$ and has margin $R$ on every feature, so $\Ltil^a(R\thu)=3\ell(R)\le3e^{-R}$.

With these two facts, Steps~2 and~3 of the proof of \Cref{thm:finite} apply verbatim to the original iterates $\theta_t$, with $d=\thu$, $\|d\|=3$ in place of $\sqrt5$, and $R=\log(T+1)$.
\end{proof}

\section{Experimental Details}\label{app:experiments}

\subsection{MNIST experimental details}
\label{app:mnist-experiments}

\paragraph{Active cones on MNIST.}
On MNIST $4$ vs.\ $7$ ($32$ images per class, raw pixels, $16$ hidden units, $3\times10^4$ steps), we ran $15$ runs with output weights fixed to $\pm1$ and $15$ with a trained output layer, over three initializations and five step sizes each. Every initial cone is active. The $23$ runs that fit the data do so by step $32$ and remain in active cones afterwards; $9$ of them pass through non-active cones, but only at steps $4$--$14$, during the initial loss spike. The other $7$ runs, all with a trained output layer and $\eta\ge0.02$, end with every hidden unit inactive on every image. Thus no run ends with a misclassified image and a live unit, and the dichotomy of \Cref{lem:dichotomy} holds at the end of training, though not at every step. Since these inputs are linearly separable, the cone in which every unit is active on every image is itself active, so the region above $\cC$, where the loss is at least $\log\frac{27}{4}$, has no counterpart here.

\paragraph{Data and preprocessing.}
We considered binary classification of MNIST digits 3 and 5.
The training set contained 128 images sampled without replacement
from each class, giving $n=256$ examples.
Sampling and subsequent shuffling used a PyTorch random generator
with seed 0.
Each $28\times28$ image was flattened into
$x_i\in\mathbb{R}^{784}$, with pixel values transformed as
\[
x_{ik}=\frac{p_{ik}/255-0.1307}{0.3081}.
\]
We assigned labels $y_i=-1$ to digit 3 and $y_i=+1$ to digit 5.
No data augmentation was used.
Test accuracy was evaluated on all 1,902 examples of these
two digits in the standard MNIST test set.

\paragraph{Architecture and initialization.}
Both experiments used a two-layer ReLU network with $m=16$
hidden units,
\[
f_\theta(x)
=
\sum_{j=1}^{m} a_j\,\relu{w_j^\top x+b_j},
\]
without an output bias.
Hidden weights were initialized using Kaiming normal initialization,
$w_{jk}\sim\mathcal{N}(0,2/784)$, and hidden biases were initialized
to zero.
The output weights were initialized to a random permutation of
eight $+1/4$ and eight $-1/4$ values.
Initialization used seed 0.
Both experiments used exactly the same training examples and
initial parameters.

In the fully trainable experiment, both weight layers and the
hidden biases were optimized.
This model is positively homogeneous of degree $L=2$ in its
trainable parameters.
In the fixed-output experiment, the output weights remained fixed
at their initial values, and only the hidden weights and biases
were optimized.
The resulting model is positively homogeneous of degree $L=1$
in its trainable parameters.
In particular, the fixed output weights were $\pm1/4$,
rather than $\pm1$.

\paragraph{Optimization.}
Both networks were trained for $T=10^4$ full-batch
gradient-descent updates with constant learning rate $\eta=0.1$.
The objective was the mean logistic loss
\[
\mathcal{L}(\theta)
=
\frac{1}{n}\sum_{i=1}^{n}
\log\!\left(1+\exp\!\left[-y_i f_\theta(x_i)\right]\right).
\]
All trainable parameters were updated simultaneously using
gradients evaluated at the current iterate.
No momentum, regularization, weight decay, or learning-rate
schedule was used.
We adopted the convention $\mathrm{ReLU}'(0)=0$.
Computations used float64 arithmetic on CPU, one PyTorch thread,
and deterministic algorithms.
Each configuration was run once, with initialization seed 0.

\paragraph{Loss reference curves.}
Panels (a) and (c) show the training loss together with reference
curves of the form $C/q_L(t)$, where
\[
q_L(t)=
\begin{cases}
t\log t, & L=2,\\
t,       & L=1.
\end{cases}
\]
Only the multiplicative constant was fitted, using
\[
\log C
=
\frac{1}{5001}
\sum_{t=5000}^{10000}
\log\!\left(q_L(t)\mathcal{L}(\theta_t)\right).
\]
The fitted constants were approximately $4.5678$ for the fully
trainable network and $5.5892$ for the fixed-output network.
These curves provide finite-time comparisons with the proposed
rates; they do not establish asymptotic convergence laws.

\paragraph{Activation patterns and example selection.}
For every training image and every iterate, we recorded the
binary gate vector
\[
g_i(t)
=
\left(
\mathbf{1}\!\left\{
w_j(t)^\top x_i+b_j(t)>0
\right\}
\right)_{j=1}^{m}.
\]
A pattern switch was counted whenever
$g_i(t)\neq g_i(t-1)$, regardless of how many individual gates
changed.
Separately for each class and each experiment, the displayed
example was selected to maximize
\[
\sum_{t=2502}^{5000}
\mathbf{1}\!\left\{g_i(t)\neq g_i(t-1)\right\},
\]
with ties resolved by the first index in the shuffled training
subset.
Thus, example selection used an earlier window than the displayed
measurements.
This rule selected the same digit-3 image in both experiments,
but different digit-5 images.

Panels (b) and (d) display the gate vectors for
$t=9851,\ldots,10000$.
Each row corresponds to one hidden neuron, numbered 0--15,
and each column corresponds to one iterate.
Black denotes an active gate, while white denotes an inactive gate.
The horizontal coordinate is $t-T$.
The reported pattern counts are the numbers of distinct complete
$m$-dimensional gate vectors in the displayed window.

\paragraph{Recurrence and periodicity.}
To measure recurrence beyond the displayed interval, we formed
a dictionary of patterns observed at $t=5001,\ldots,9000$,
then counted previously unseen patterns during
$t=9001,\ldots,10000$.
We also tested whether the final 150-vector sequence had an
exact period $p\in\{1,\ldots,75\}$, requiring equality at every
pair of positions separated by $p$.

\begin{table}[t]
\centering
\small
\begin{tabular}{llrrr}
\hline
Configuration & Digit & MNIST index & Patterns & New patterns \\
\hline
Fully trainable & 3 & 46247 & 49 & 10 \\
Fully trainable & 5 & 38658 & 16 & 0 \\
Fixed output   & 3 & 46247 & 56 & 0 \\
Fixed output   & 5 & 20171 & 25 & 0 \\
\hline
\end{tabular}
\caption{
Activation-pattern measurements for the displayed examples.
MNIST indices are zero-based indices in the original training
split.
``Patterns'' counts distinct gate vectors in the final 150
iterates.
``New patterns'' counts distinct vectors in iterates
9001--10000 that were absent from iterates 5001--9000.
}
\label{tab:mnist-gate-patterns}
\end{table}

All four displayed examples changed pattern on all 149
transitions in the displayed window.
None had an exact period of at most 75 iterations.
These observations demonstrate finite-time recurrence and
switching, rather than infinite-time periodicity.

\paragraph{Interpretation as parameter-space features.}
For the fixed-output model, write
$\widetilde{x}_i=(x_i^\top,1)^\top$ and concatenate the trainable
hidden parameters as
\[
\theta=(v_1^\top,\ldots,v_m^\top)^\top,
\qquad
v_j=(w_j^\top,b_j)^\top.
\]
For each gate pattern $g\in\{0,1\}^{m}$, define the signed feature
vector
\[
\psi_{i,g}
=
y_i
\begin{pmatrix}
a_1g_1\widetilde{x}_i\\
\vdots\\
a_mg_m\widetilde{x}_i
\end{pmatrix}.
\]
The signed prediction then satisfies
\[
y_i f_\theta(x_i)
=
\left\langle
\theta,\psi_{i,g_i(\theta)}
\right\rangle.
\]
Since the image and output weights are fixed, each gate pattern
determines a fixed vector in parameter space.
Recurrent gate patterns therefore correspond to recurrent
fixed parameter-space feature vectors.
This does not imply that the real-valued hidden activations
are constant.
For the fully trainable model, repeated gates alone do not
specify fixed parameter-space features, because the output
weights also evolve.

\paragraph{Numerical results and validation.}
Table~\ref{tab:mnist-results} reports the final losses and
classification accuracies.
Analytical gradients were checked against PyTorch autograd at
initialization and termination, with maximum absolute
discrepancies below $10^{-12}$.
Saved gate codes were verified against recorded preactivations
over the final 1,000 iterates.
The fixed output weights were confirmed to remain exactly
unchanged, and the two experiments were verified to use identical
initial parameters and training examples.
The implementation used PyTorch 2.9.0 and NumPy 2.2.6.
These results describe one matched pair of runs and do not
quantify variability across initializations.

\begin{table}[t]
\centering
\small
\begin{tabular}{lrrr}
\hline
Configuration & Final loss & Train accuracy & Test accuracy \\
\hline
Fully trainable & $4.89253\times10^{-5}$ & $100\%$ & $94.53\%$ \\
Fixed output   & $5.59540\times10^{-4}$ & $100\%$ & $95.22\%$ \\
\hline
\end{tabular}
\caption{
Final performance after $10^4$ full-batch gradient-descent
updates on the MNIST 3-versus-5 training subset.
}
\label{tab:mnist-results}
\end{table}

\subsection{Implicit bias}

\paragraph{Implementation.}
All experiments use full-batch GD on \eqref{eq:model} with the exact gradient under $\mathrm{ReLU}'(0)=0$. The loss uses a numerically stable softplus. Double precision is used except for the catapult with $\eta\ge16$ and all runs with $\eta_t=\eta_0/\cL(\theta_t)$, which use 80 significant digits (\texttt{mpmath}). The sharpness in \Cref{fig:overview}(c) is the largest eigenvalue of the Hessian of the branch selected by the ReLU convention (see the proof of \Cref{rem:eos-threshold}). The script \texttt{code/experiments.py} regenerates all figures (Python~3, NumPy, SciPy, Matplotlib, mpmath) in a few minutes on a laptop CPU.

\begin{figure*}[t]
\centering
\includegraphics[width=\textwidth]{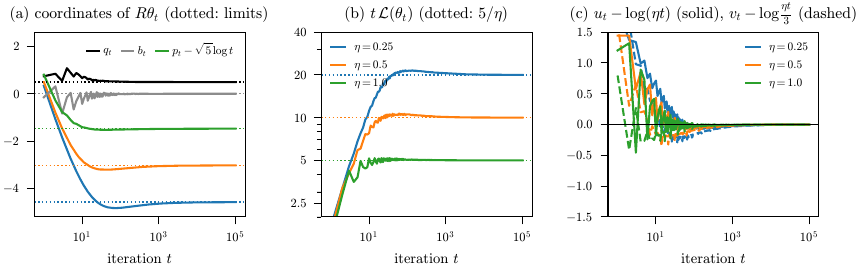}
\caption{\textbf{Implicit bias with constant steps} ($\theta_0=(0.2,0.1,0)$, $\eta\in\{0.25,0.5,1\}$, $10^5$ iterations). (a) Coordinates of $\vartheta_t$ in the rotated basis: $q_t$ (black) and $b_t$ (gray) for $\eta=1$, and $p_t-\sqrt5\log t$ for each $\eta$ (colors as in (b)); dotted lines are the limits of \Cref{thm:implicit}(a). (b) $t\,\cL(\theta_t)\to5/\eta$. (c) Endpoint and middle margins minus their predicted growth tend to zero.}
\label{fig:implicit}
\end{figure*}


\paragraph{After the catapult.}
From $(0.2,0.1,-0.1)$ with $\eta\in\{16,64,256,1024\}$ and $T=3000$ steps, the iterates stay in $\cC$ for all $t\le T$. For $\eta\ge64$, $b_t\le0$ for all $2\le t\le T$ (so the exit condition \eqref{eq:exit} can never be met), and $\log_{10}\cL(\theta_2)=-3.29,-13.80,-55.85$ for $\eta=64,256,1024$; for $\eta=16$ the sign of $b_t$ keeps switching but the iterates remain in $\cC$ and $\log_{10}\cL(\theta_{3000})=-4.06$.

\paragraph{Random initializations.}
We draw $20$ initializations with $|w_0|=10^{U}$, $U\sim\mathrm{Unif}[-1.3,0.7]$, random sign, $b_0\sim\mathrm{Unif}(-|w_0|,|w_0|)$ and $g_0\sim\mathrm{Unif}[-3,3]$ (seed $0$), and report in \Cref{tab:random} whether GD stays in $\cC$ and the final loss.

\begin{table}[h]
\centering
\caption{Twenty random initializations in $\cC$.}
\label{tab:random}
\begin{tabular}{lcccc}
\toprule
schedule & $T$ & stays in $\cC$ & median $\log_{10}\cL(\theta_T)$ & max $\log_{10}\cL(\theta_T)$\\
\midrule
$\eta_t=|w_t|/2$ & $2\cdot10^4$ & $20/20$ & $-4.63$ & $-4.63$\\
$\eta_t=1/\cL(\theta_t)$ & $300$ & $18/20$ & $-25.16$ & $\phantom{-}0.28$\\
$\eta_t=|w_t|^2/2$ & $2000$ & $0/20$ & $\phantom{-}1.05$ & $\phantom{-}2.21$\\
\bottomrule
\end{tabular}
\end{table}

The two failures of $\eta_t=1/\cL(\theta_t)$ start at $(-0.089,0.030,0.883)$ and $(-0.209,-0.006,2.337)$: both have small $|w_0|$ and $2g_0>W_0+b_0$, so neither \ref{I1} nor \ref{I2} holds at $t=0$, and GD enters $\cD$ within two steps. With $\eta_0=0.25$ both runs stay in $\cC$. For $\eta_t=|w_t|/2$, the median and the largest final loss coincide to two decimals, consistent with the empirical scaling $5/(t\log t)$. This value of $\kappa$ lies outside the range of \Cref{thm:adaptive}.

\end{document}